\documentclass[authoryear,3p,11pt]{elsarticle}

\usepackage{titlesec}

\usepackage{geometry}

\usepackage{setspace}

\newcommand{\w}{\omega}
\newcommand{\qedblack}{\hfill$\blacksquare$}
\newcommand*{\red}{\textcolor{black}}
\newtheorem{theorem}{Theorem}
\newtheorem{condition}{Condition}
\newtheorem{proposition}{Proposition}

\usepackage{bm}
\usepackage{dsfont}
\usepackage{amssymb}
\usepackage{amsmath}
\usepackage{url}

\usepackage{algorithm}
\usepackage{algpseudocode}
\usepackage{setspace}

\usepackage{enumitem}
\usepackage{xcolor}

\newtheorem{claim}{Claim}
\journal{}

\begin{document}

\makeatletter
\def\ps@pprintTitle{%
     \let\@oddhead\@empty
     \let\@evenhead\@empty
     \let\@oddfoot\@empty
     \let\@evenfoot\@empty
}
\makeatother

\begin{frontmatter}



\title{Soft regression trees: a model variant and a decomposition training algorithm} 

\author[mymainaddress]{Antonio Consolo\corref{mycorrespondingauthor}}
\author[mymainaddress]{Edoardo Amaldi}
\author[mysecondaryaddress]{Andrea Manno}

\cortext[mycorrespondingauthor]{Corresponding author}

\address[mymainaddress]{DEIB, Politecnico di Milano, Milano, Italy}
\address[mysecondaryaddress]{Centro di Eccellenza DEWS, DISIM, Università degli Studi dell’Aquila, L'Aquila, Italy}

\begin{abstract}
\textcolor{black}{
Decision trees are widely used for classification and regression tasks in a variety of application fields due to their interpretability and good accuracy. During the past decade, growing attention has been devoted to globally optimized decision trees with deterministic or soft splitting rules at branch nodes, which are trained by optimizing the error function over all the tree parameters. In this work, we propose a new variant of soft multivariate regression trees (SRTs) where, for every input vector, the prediction is defined as the linear regression associated to a single leaf node, namely, the leaf node obtained by routing the input vector from the root along the branches with higher probability. SRTs exhibit the conditional computational property, i.e., each prediction depends on a small number of nodes (parameters), and our nonlinear optimization formulation for training them is amenable to decomposition. After showing a universal approximation result for SRTs, we present a decomposition training algorithm including a clustering-based initialization procedure and a heuristic for reassigning the input vectors along the tree. Under mild assumptions, we establish asymptotic convergence guarantees. Experiments on 15 well-known datasets indicate that our SRTs and decomposition algorithm yield higher accuracy and robustness compared with traditional soft regression trees trained using the nonlinear optimization formulation of Blanquero et al., and a significant reduction in training times as well as a slightly better average accuracy compared with the mixed-integer optimization approach of Bertsimas and Dunn. We also report a comparison with the Random Forest ensemble method.
}
\end{abstract}





\end{frontmatter}




\section{Introduction}\label{sec:introduction}
Decision trees are popular supervised learning methods for classification and regression in Machine Learning (ML) and Statistics. 
They are widely used in a number of fields ranging from Business Analytics (see e.g. \cite{ghodselahi2011application,ouahilal2016comparative,oztekin2018creating}) to Medicine and Biology (see e.g. \cite{ozcan2023classification,johns2021distance,yu2020improving}).
The success of decision trees lies mainly in their interpretability and good accuracy. Unlike most black-box ML models, they reveal the 
feature-based decisions leading to the tree response for any input vector.  
Interpretability is of particular importance in applications where the ML models complement human decision making and justifiable predictions are required, such as for instance in medical diagnosis and criminal sentencing (see e.g. \cite{rudin2019stop}). 

A decision tree is a directed binary tree with a set of branch nodes, including the root, a set of leaf nodes, and two outgoing arcs (branches) for each branch node associated to a splitting rule. 
Any input vector is routed from the root along the tree according to the splitting rules at the branch nodes, eventually falling into a leaf node where the output (the linear prediction or the class number) is determined. Hard or soft \textcolor{black}{splitting rules} can be considered at branch nodes. In hard (deterministic) splits, the left branch is followed if a single feature (univariate case) or a linear combination of the features (multivariate case) exceeds a given threshold value. In soft splits, both left and right branches are followed with complementary probabilities given by a continuous sigmoid function of a linear combination of the features.


\textcolor{black}{Since training decision trees is known to be NP-hard \citep{laurent1976constructing} and is very challenging in practice, early methods like CART \citep{breimanclassification} build classification and regression trees via a
greedy approach, where at each branch node the split is determined by minimizing a local error function. The later variants C4:5 \citep{quinlan2014c4} and ID3 \citep{quinlan1986induction} also include a pruning phase} to decrease the tree size and reduce overfitting. \textcolor{black}{Different types of tree models and approaches to train them have been proposed and studied during the past forty years, see for instance the recent survey \citep{costa2022} and the references therein.} 


Due to the remarkable progresses in Mixed Integer Linear Optimization (MILO) and Non-Linear Optimization (NLO) methods and solvers, a growing attention has been devoted during the last decade to revisit decision trees under a modern optimization lens. 
\textcolor{black}{The ultimate goal is to develop 
algorithms that globally optimize decision trees, i.e., that simultaneously tune the values of all the tree parameters, with local or global optimality guarantees.}
See, for instance, \citep{bertsimas2019machine,aghaei2020learning} for MILO approaches to deterministic classification and regression trees and \citep{blanquero2020sparsity,blanquero2022sparse} for NLO approaches to soft classification and regression trees. 



Globally optimized decision trees are of interest not only because they may lead to improved accuracy but also because they allow to impose \textcolor{black}{additional constraints such as fairness constraints} which prevent discrimination of sensitive groups (see e.g. \cite{nanfack2022constraint}).


In this work, we first propose and investigate a
\textcolor{black}{new variant of soft multivariate regression trees where, for any input vector, the prediction is defined as the linear regression associated to a single leaf node. Such soft trees satisfy the conditional computational property, i.e., each prediction depends on a small number of nodes (parameters), and lead to computational and statistical benefits. Since the proposed nonlinear optimization formulation for training them is amenable to decomposition, we present a convergent decomposition training algorithm.}
The results obtained for a collection of well-known datasets are compared with those provided by \textcolor{black}{four} 
alternative methods for training regression trees or closely related models.  

The remainder of the paper is organized as follows. In Section \ref{sec:previous-work} we mention some previous and related work on globally optimized deterministic and soft regression trees. In Section \ref{sec:mrrt}, we present the new soft regression tree variant, the associated formulation for training, and a universal approximation theorem. In Section \ref{sec:dec_reg}, we describe the general node-based decomposition scheme and discuss asymptotic convergence guarantees. Section \ref{subsec:implementation} is devoted to the detailed description of the implemented decomposition algorithm, including \textcolor{black}{an ad hoc initialization procedure} and a heuristic for reassigning the input vectors along the tree branch nodes. 
In Section \ref{sec:experiments_regression} we assess the performance of the proposed model variant and decomposition algorithm on 15 datasets from the literature. 
The testing accuracy of the SRTs is compared with that of traditional soft regression trees  
\textcolor{black}{trained via the NLO formulation in \citep{blanquero2022sparse} and the deterministic regression trees built using the MILO approach in \citep{bertsimas2019machine}.} 
Finally, Section \ref{sec:conclusions} contains some concluding remarks. 
\textcolor{black}{The proofs of the main results, some additional computational results and a comparison with 
the Random Forest ensemble model (\cite{breiman2001random}) are included in the Appendices.}

\section{Previous 
work} \label{sec:previous-work}


\textcolor{black}{As previously mentioned, the notable improvements in optimization solvers and computer performance have recently been fostering research on globally optimized decision trees for classification and regression tasks. In this section, we mention previous work on MILO and NLO approaches to design deterministic regression trees and, respectively, train soft regression trees.} 

\textcolor{black}{In \cite{yang2017regression}, the authors present a recursive partitioning algorithm for building univariate deterministic regression trees. A MILO formulation is solved to determine for each branch node the 
input feature and the associated optimal threshold value, as well as the coefficients of the linear regressions for the two child leaf nodes. 
}

In \citep{bertsimas2019machine,dunn2018optimal} the authors propose MILO formulations and a local search method to build deterministic optimal regression trees considering both the mean square error and the tree complexity. They introduce an $\ell_1$ norm regularization term and consider two splitting rules, namely, orthogonal (univariate) and oblique (multivariate) splits. \textcolor{black}{The method for deterministic regression trees with univariate splits is referred to as ORT-L, while that with multivariate splits is referred to as ORT-LH.} In \cite{verwer2017learning} optimal regression 
trees are encoded as integer optimization problems where the regression task is enforced by means of suitable constraints. In \cite{zantedeschi2020learning} a sparse relaxation for regression 
trees is presented, which is able to learn splitting rules and tree pruning by means of argmin differentiation. 

\textcolor{black}{In \cite{zhang2022optimal}, a dynamic programming approach is proposed for constructing optimal sparse univariate regression trees. The idea is to restrict the exploration of the set of the feasible solutions using a novel lower bound based on an optimal solution of an ad hoc one-dimensional instance of the k-means clustering problem applied to the output values of the considered dataset.} 
\textcolor{black}{To deal with continuous input features, 
	the continuous domain is divided into equally sized intervals by means of binning techniques.}

\textcolor{black}{To the best of our knowledge, the first soft decision trees 
were introduced in \citep{suarez1999globally} and were referred to as ``fuzzy decision trees".
The authors proposed two distinct objective functions to tackle classification and regression tasks. For any input vector, the overall soft regression tree prediction is determined by combining the outputs 
of all leaf nodes.
Such soft decision trees are trained using a gradient-based algorithm similar to backpropagation for feedforward neural networks. 
%
In \cite{irsoy2012soft}, the authors consider the same soft splits where sigmoids are applied to linear combinations of the input features and present an algorithm to grow the tree, one node at a time, where the node parameters are adjusted via gradient descent.
Two variants of the approach are described to deal with classification and regression tasks.}

Recently, in \citep{blanquero2022sparse} the authors consider 
\textcolor{black}{soft regression trees, referred to as randomized regression trees,} where, at each branch node, a random variable is used to establish along which branch the input vector must be routed. For any given input vector, the tree prediction is a weighted combination of all the leaf nodes outputs, \textcolor{black}{where the weights are the probabilities that the input vector falls into the corresponding leaf nodes.} The proposed unconstrained nonlinear optimization formulation to train them 
is tackled with a state-of-the-art open-source NLO solver. \textcolor{black}{Sparsity and fairness can also be taken into account.}

\textcolor{black}{In \cite{bertsimas2021near}, ORT-LH and MARS \citep{friedman1991multivariate} are combined to obtain the so-called Near-optimal Nonlinear Regression Trees with Hyperplanes, where the predictions in the leaf nodes are based on polynomial functions instead of linear ones.}
\textcolor{black}{Such trees are trained by tackling an unconstrained nonlinear optimization problem with a differentiable regularized objective function using gradient-based methods.}



\textcolor{black}{To conclude this section, it is worth pointing out the main differences between deterministic regression trees and soft regression trees.}

\textcolor{black}{Due to the hard splits, deterministic 
trees satisfy the \textit{conditional computation} property \citep{bengio2015conditional}, namely, only a small part of the tree is involved in routing any specific 
\textcolor{black}{
input vector. 
In particular, in such trees every input vector is routed towards a single leaf node.} Selective activation brings notable computational benefits: it enables faster inference and reduces the parameter usage for each input vector. 
Moreover, it acts as a regularizer that enhances the statistical properties of the model \citep{breimanclassification,hastie2009elements,bengio2015conditional}. However, the MILO formulations that have been proposed so far to train deterministic regression trees are very challenging computationally even for medium-size datasets.}
\textcolor{black}{In traditional soft regression trees, the conditional computation property is not satisfied.}
Indeed, each input vector falls into all the leaf nodes and the corresponding prediction is defined as the weighted sum of the linear regressions associated to all the leaf nodes, where the weight for each leaf is the probability that the input vector falls into it.  


\textcolor{black}{Since soft splits involve sigmoid functions, the formulations for training soft regression trees are smooth but nonconvex and can be tackled with NLO 
algorithms such as gradient-based methods.}
\textcolor{black}{In such soft regression tree formulations, the number of variables only depends on the number of features $p$ and the depth of the tree $D$, while in the MILO formulations for deterministic regression trees it also depends on the number of data points $N$ which can be much larger.}
\textcolor{black}{Although NLO methods do not guarantee global optimality for nonconvex problems, they tend to scale better when minimizing the soft regression tree error function than their MILO counterparts for deterministic regression trees. On the one hand, the NLO formulations involve a substantially smaller number of variables than MILO ones and, on the other hand, they are amenable to decomposition with respect to the variables and the terms associated to both the model parameters and the data points.}

\section{
Soft regression trees with single leaf node 
predictions
}\label{sec:mrrt}

\textcolor{black}{In this section, we present a new variant of soft multivariate regression trees
where each input vector falls into a single leaf node, which is obtained by starting from the root node and by following 
at each internal node the branch with higher probability, and the corresponding prediction is defined as the linear regression of that single leaf node.}
\textcolor{black}{As we shall see, 
such soft trees satisfy the “conditional computation” property, which leads to both computational and statistical advantages.} 

In the remainder of this section, we first describe the new model variant and the associated training formulation, and then prove a universal approximation theorem for such soft trees.

\subsection{The soft regression tree model variant and the training formulation}\label{subsec:model-formulation}

In regression tasks, we consider a training set $I=\left \{(\mathbf{x}_{i} ,y_{i})\right \}_{1\leq i\leq N}$ consisting of $N$ data points, where $\mathbf{x}_{i} \in \mathbb{R}^{p}$ is the $p$-dimensional vector of  the input features and $y_{i} \in \mathbb{R}$ the associated response value\footnote{Categorical and continuous features can be dealt with via an appropriate scaling over the $[0,1]$ interval.}.

As in \citep{suarez1999globally} and \citep{blanquero2022sparse}, our soft regression trees are maximal binary multivariate trees of a given fixed depth $D \geq 2$. Each branch node has two children and all the leaf nodes have the same depth $D$. 
Let $\tau_{B}$ and $\tau_{L}$ denote, respectively, the set of branch nodes and the set of leaf nodes. 

At each branch node $t \in \tau_{B}$, a Bernoulli random variable is used to decide which branch to take. The probability of taking the left branch is determined by the value of a cumulative distribution function (CDF) evaluated over a linear combination of the input features.
For each input feature $j=1,\ldots,p$ and each branch node $t \in \tau_{B}$, let the variable $\w_{jt} \in \mathbb{R}$ denote the coefficient of the $j$-th input feature in the soft oblique splitting rule at branch node $t$ and let the variable $\w_{0t} \in \mathbb{R}$ denote the intercept of the linear combination. 
For each input vector $\mathbf{x}_{i}$, with $i=1,\ldots,N$, and each branch node $t \in \tau_{B}$, the corresponding parameter of the Bernoulli distribution is defined as follows:
$$ p_{it}= F(\bm{\w}_{ t}^T\mathbf{x}_i) =F(\w_{0t} + \frac{1}{p}\sum_{j=1}^{p}\w_{jt}x_{ij}), $$
where $\bm{\w}_{t}$ denotes the $(p+1)$-dimensional vector containing the $p$ coefficients and the intercept of branch node $t$, and the CDF is the logistic function
\begin{equation}\label{eq:cdf}
    F(u)=\frac{1}{1+exp\left ( -\mu u  \right ) }
\end{equation}
with the positive parameter $\mu$.
	
\begin{figure}
\centering
\begin{minipage}{.75\textwidth}
  \centering
  \vspace{-20pt}
  \includegraphics[width=\linewidth]{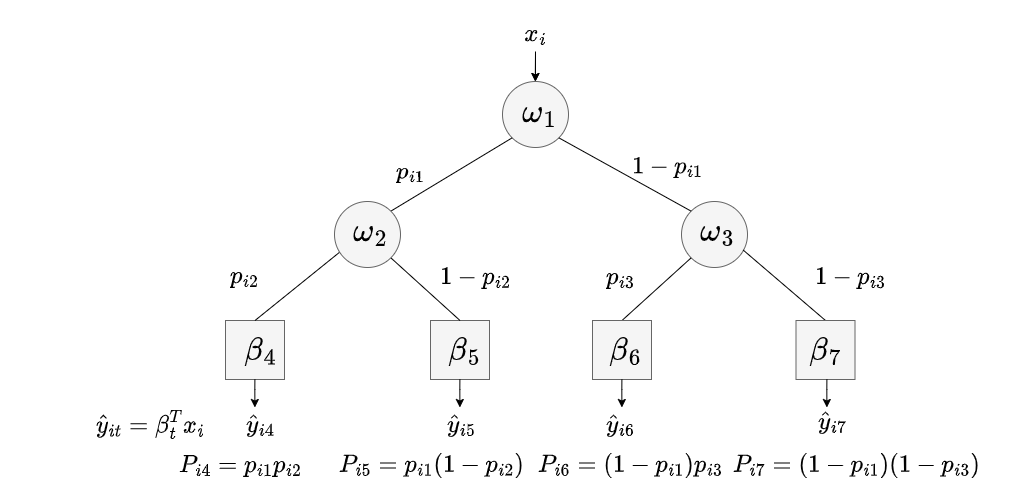}\label{fig:test1}
  
  (a)
  
\end{minipage}%
\begin{minipage}{.25\textwidth}
  \centering
  \includegraphics[width=.9\linewidth]{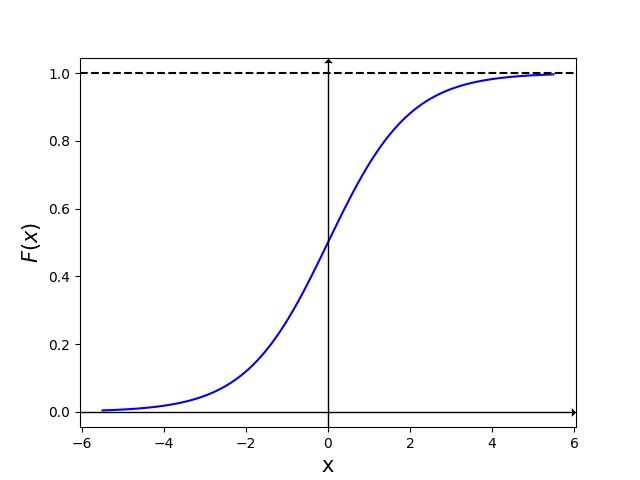}\label{fig:test2}
  
  (b)
  
\end{minipage}
\caption{(a) Depicts an example of soft regression tree of depth $D=2$. (b) Represents the logistic CDF which corresponds to the probability of taking the left branch.}\label{fig:prova}
\end{figure}

As shown in Figure \ref{fig:prova} (a), $p_{it}$ is the probability of taking the left branch and $1-p_{it}$ that of taking the right one.

Since the logistic CDF induces a soft splitting rule at branch nodes, 
every input vector falls with nonzero probability into every leaf node $t \in \tau_{L}$.
For any leaf node $t \in \tau_{L}$, let $A_{L}(t)$ denote the set of its ancestor nodes whose left branch belongs to the path from the root to node $t$, and $A_{R}(t)$ the set of its ancestor for the right branches. Adopting this notation, the probability that any input vector $\mathbf{x}_{i}$, with $i=1\ldots,N$, falls into any leaf node $t \in \tau_{L}$ is:
$$ P_{it} 
=\prod_{t_{l}\in A_{L(t)}}p_{it_{l}}\;\prod_{t_{r}\in A_{R(t)}}(1-p_{it_{r}}).$$


\textcolor{black}{For any input vector $\mathbf{x}_{i}$, with $i=1,\ldots,N$, and for any leaf node $t \in \tau_{L}$, the corresponding output is denoted by $\hat{y}_i$ and is defined as the following linear regression of the features
$$\hat y_{it} = \bm{\beta}_{t}^{T}\mathbf{x}_{i} = \beta_{0t}+\beta_{1t}x_{i1}+\cdots+\beta_{pt}x_{ip},$$
\noindent where the variable $\beta_{0t}$ represents the intercept and the variables $\beta_{jt}$, for $j=1,\ldots,p$, the coefficients. From now on, the}
 $(p + 1) \times |\tau_{L}|$ matrix  $\bm{\beta} = (\beta_{jt})_{j \in \{0,1,\ldots,p\}, t \in \tau_{L}}$ contains all the parameters of the leaf nodes and the $(p + 1) \times |\tau_{B}|$  matrix  $\bm{\omega} = (\omega_{jt})_{j \in \{0,1,\ldots,p\}, t \in \tau_{B}}$ all the parameters of the branch nodes.


\red{Assuming that values have been assigned to the variables} $\bm{\w}$ and $\bm{\beta}$, for any input vector $\mathbf{x} \in \mathbb{R}^{p}$ the prediction of our Soft Regression Tree (SRT) variant is deterministically defined as follows:
\begin{equation}\label{eq:prediction_model}
\hat{y} = \sum_{t\in \tau _{L}}\left(\prod_{\ell\in A_{L(t)}}
\mathds{1}_{0.5}(p_{\textbf{x}\ell}(\bm{\w}_{\ell})) \;\prod_{r\in A_{R(t)}}\mathds{1}_{0.5}(1-p_{\textbf{x}r}(\bm{\w}_r))\right)\hat y_{t}(\bm{\beta}_t),
\end{equation}

\noindent where $p_{\textbf{x}t}$ is the probability that $\textbf{x}$ is routed along the left branch of node $t$, and $\mathds{1}_{\alpha}(u) = 1$ if $u \geq \alpha$ while $\mathds{1}_{\alpha}(u) = 0$ if $u < \alpha$. Notice that exactly one parenthesis in \eqref{eq:prediction_model} is nonzero. 


\textcolor{black}{In words, for any input vector $\mathbf{x}$ we consider the specific root-to-leaf node path obtained by deterministically routing $\mathbf{x}$ at each branch node along the branch with the highest probability, which we referred to as its Highest Branch Probability (HBP) path. Then, the prediction for $\mathbf{x}$ is computed as the linear regression associated to the leaf node of the HBP path. In the sequel, we say that $\mathbf{x}$ deterministically falls into all the nodes of the HBP path.}

\begin{figure}[H]
    \centering
    \includegraphics[width = 0.6\textwidth]{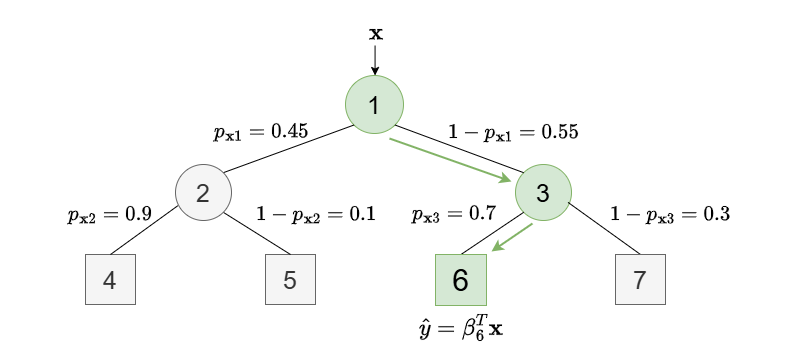}
    \caption{Example of SRT with single leaf node predictions. For input vector $\mathbf{x}$, the arrows indicate the branches belonging to its HBP path, while the prediction is equal to  $\mathbf{\beta}_6^T \mathbf{x}$.}
    \label{fig:percorso}
\end{figure}

Figure \ref{fig:percorso} illustrates the way a tree of depth $D=2$ provides a prediction for any given input vector $\mathbf{x}$. The corresponding HBP path is highlighted in green and the prediction is given by the linear regression $\hat{y} = \bm{\beta}_6^T\mathbf{x}$ associated to leaf node 6. 
\textcolor{black}{Notice that for any input vector $\bf x$ the leaf node of the corresponding HBP path is not necessarily the one into which $\bf x$ falls with the highest probability.}


\textcolor{black}{It is worth emphasizing that the deterministic way the prediction is defined in SRTs guarantees the conditional computation property since each prediction only depends on a small number of nodes and the corresponding parameters. From the computational point of view, routing every input vector along a single HBP path leads to substantial speed-ups in both prediction phase and training phase.}


\textcolor{black}{
In order to train SRTs, we consider the probabilistic perspective.} For any assignment of values to the variables $\bm{\w}$ and $\bm{\beta}$, the soft regression tree provides, for any input vector $\mathbf{x} \in \mathbb{R}^{p}$, $\vert \tau_{L} \vert$ potential linear predictions (one linear regression for every leaf node) 
with the corresponding probabilities $P_{\textbf{x}t}$ 
that $\mathbf{x}$ falls into the associated leaf node $t$. 
\textcolor{black}{Then we train SRTs by solving the following unconstrained nonlinear optimization problem:}
\begin{equation}
\label{eq:formulation1b}
\begin{split}
\min_{\bm{\w},\bm{\beta}} E(\bm{\w},\bm{\beta})= \frac{1}{N}\sum_{i \in I}\left(\sum_{t\in \tau _{L}}\left(\prod_{\ell\in A_{L(t)}}p_{i\ell}(\bm{\w}_{\ell})\;\prod_{r\in A_{R(t)}}(1-p_{ir}(\bm{\w}_r))\right)(\hat y_{it}(\bm{\beta}_t)-y_{i})^{2}\right)\\
= \frac{1}{N}\sum_{i \in I}\left(\sum_{t\in \tau _{L}}P_{it}(\bm{\w})(\hat y_{it}(\bm{\beta}_t)-y_{i})^{2}\right). \end{split}
\end{equation}
The objective function in (\ref{eq:formulation1b}), which expresses the prediction accuracy, amounts to a weighted mean squared error between the predictions and the associated responses over all the data points in the training set. 
The difference with respect to classic MSE is that for each data point the squared error of the linear regression associated to every leaf node is weighted by the probability that the data point falls into that leaf node.
As we shall see below, the training formulation \eqref{eq:formulation1b} is amenable to decomposition.

In \cite{blanquero2022sparse}, the authors propose soft regression trees where, at each leaf node, the constant outputs used in \citep{suarez1999globally} is replaced by a linear regression. 
For any input vector $\mathbf{x}$, the tree prediction is defined as a weighted sum of all the leaf node outputs, where the weights are the probabilities that $\mathbf{x}$ falls into the corresponding leaf nodes.  \textcolor{black}{The training problem amounts to the following unconstrained nonlinear optimization problem:}

\begin{equation}
\label{eq:formulation}
\min_{\bm{\w},\bm{\beta}} \frac{1}{N}\sum_{i \in I}\left(\sum_{t\in \tau _{L}}P_{it}(\bm{\w})\hat y_{it}(\bm{\beta}_t)-y_{i}\right)^{2}, 
\end{equation}
where the prediction associated to any input vector $\mathbf{x}_{i}$ is given by $\sum_{t\in \tau _{L}}P_{it}(\bm{\w})\hat y_{it}(\bm{\beta}_{t})$ and is deterministic since the probabilities $P_{it}$ are considered as weights. \textcolor{black}{It is worth pointing out that, unlike in \eqref{eq:formulation1b}, the output of such soft regression trees involves, for every $\mathbf{x}_i$, all the leaf nodes outputs $\hat{y}_{it}$.} Finally, note that traditional soft regression trees can be viewed as mixtures of experts, whose experts correspond to the linear regressions associated to the leaf nodes.

\subsection{Universal approximation property}\label{subsec:approximability}

In this subsection we present a universal approximation theorem for the new soft regression tree model variant 
introduced above. Our result, which is similar in spirit to the ones in \citep{cybenko1989approximation} for the linear superposition of sigmoidal functions 
and in \citep{nguyen2016universal} for mixture of experts models, provides approximation guarantees when applying such soft regression trees to datasets arising from nonlinear 
problems. 

Consider any SRT of depth $D$, with $D \geq 1$, where the input vector $\mathbf{x} \in \mathbb{X} \subset \mathbb{R}^p$ and the output $y \in \mathbb{R}$. 
For any choice of values for the branch and leaf nodes parameters $\bm{\w}$ and $\bm{\beta}$, such a SRT defines a real-valued function over $\mathbb{X}$.
For any input vector $\mathbf{x} \in \mathbb{X}$, the prediction is given by
\begin{equation}
	\label{eq:prediction}
    \sum_{t\in \tau_L}\left(\prod_{\ell\in A_{L(t)}}
	\mathds{1}_{0.5}\left(\frac{1}{1+exp\left ( -\mu (\bm{\w}_{\ell}^T\mathbf{x})  \right )}\right) \;\prod_{r\in A_{R(t)}}\mathds{1}_{0.5}\left(1 - \frac{1}{1+exp\left ( -\mu (\bm{\w}_{r}^T\mathbf{x}) \right )}\right)\right) 
	\bm{\beta}^T_t\mathbf{x}, 
\end{equation}
where the subsets $A_{L}(t)$ and $A_{R}(t)$ of the ancestors of the leaf node $t$ are as above. Notice that the only nonzero parenthesis in the summation is the one associated to the leaf node of the HBP path for $\mathbf{x}$.

For each input vector $\mathbf{x}$ and leaf node $t \in \tau_L$, the corresponding term of the summation in \eqref{eq:prediction} is denoted by
\begin{equation}
	\label{eq:pi_approx}
	\pi_{\mathbf{x}t}(\bm{\w}_{A(t)})
	= \prod_{\ell\in A_{L(t)}}
	\mathds{1}_{0.5}\left(\frac{1}{1+exp\left ( -\mu (\bm{\w}_{\ell}^T\mathbf{x})  \right )}\right) \;\prod_{r\in A_{R(t)}}\mathds{1}_{0.5}\left(1 - \frac{1}{1+exp\left ( -\mu (\bm{\w}_{r}^T\mathbf{x})  \right )}\right), 
\end{equation}
where $A(t) = A_{L}(t) \cup A_{R}(t)$ is the set all the ancestors of  $t$ and $\bm{\w}_{A(t)}$ is the parameter vector including all the $\w$ parameters associated to the branch nodes which are ancestors of $t$.

Let us define the class of all the functions that can be implemented by any SRT of depth $D \geq 1$, that is, of the form \eqref{eq:prediction}, as
$$\mathbb{M} = \{\sum_{t = 1}^{2^D} \pi_{\mathbf{x}t}(\bm{\w}_{A(t)}) \bm{\beta}^T_t\mathbf{x}\ |\ D \in \mathbb{N}, \bm{\chi} 
\in \mathbb{R}^{(p+1)(2^{D}-1)+(p+1)2^{D}}\},$$
where $\bm{\chi}$ is the vector containing all the model parameters in $\bm{\w}$ and $\bm{\beta}$. 
We also consider the subclass of $\mathbb{M}$ where all the leaf node parameters are set to zero except the intercepts $\beta_{0t}$:
$$\mathbb{H} = \{ \sum_{t = 1}^{2^D} \pi_{\mathbf{x}t}(\bm{\w}_{A(t)}) \beta_{0t}\ |\ D \in \mathbb{N}, \bm{\chi}' 
\in \mathbb{R}^{(p+1)(2^D-1)+2^{D}}\},$$
where $\bm{\chi}'$ is the vector containing all the model parameters 
in $\bm{\w}$ and $\bm{\beta}$.

In \ref{sec:appendixA}, 
we show that any continuous function defined over a compact set $\mathbb{X}$, i.e., any function in $C(\mathbb{X})$, can be approximated to any degree of accuracy $\varepsilon > 0$ by a function implemented by a SRT of an appropriate depth $D \geq 1$.


In particular, we prove the following universal approximation result:

\begin{theorem}
	\label{th:dense}
	Assuming that $\mathbb{X} \subset \mathbb{R}^p$ is an arbitrary compact set, the class $\mathbb{H}$ is dense in $\mathbb{C}(\mathbb{X})$. 
	In other words, for any $\varepsilon > 0$ and any $g \in \mathbb{C}(\mathbb{X})$, there exists a function $f \in \mathbb{H}$ such that 
	$$\sup_{\mathbf{x}\in \mathbb{X}}|f(\mathbf{x}) - g(\mathbf{x})| < \varepsilon.$$
	Since $\mathbb{H} \subset \mathbb{M}$, also the class $\mathbb{M}$ is dense in $\mathbb{C}(\mathbb{X})$. 
\end{theorem}

Notice that we make no assumptions on the domain $\mathbb{X}$ (other than compactness) and on the differentiability of the target functions $g$. The proof of  Theorem \ref{th:dense} is based on the Stone-Weierstrass theorem as in \cite{cotter1990stone-weierstrass}.

\section{A general decomposition scheme}\label{sec:dec_reg}
The soft regression tree model variant presented in Section \ref{sec:mrrt} is trained by minimizing the challenging nonconvex error function in \eqref{eq:formulation1b} with respect to the variables $\bm{\w}$ and $\bm{\beta}$. This is a challenging computational problem not only because of the nonconvexity 
but also because the computational load rapidly grows when the depth of the tree $D$, and the size of the dataset, namely $p$ and $N$, increase. 

For other ML models similar drawbacks have been tackled by developing decomposition algorithms where the original problem is splitted into a sequence of subproblems, in which at each iteration one optimizes over a different subset of variables (whose indices correspond to the so-called working set) while keeping fixed the other variables to their current values. \textcolor{black}{For instance, decomposition algorithms have been devised for Support Vector Machines (e.g., \cite{chang2011libsvm,manno2016convergent,manno2018parallel}), multilayer perceptrons \citep{grippo2015decomposition} and soft classification trees \citep{amaldi2023multivariate}.}

Decomposition algorithms are particularly effective whenever the subproblems have some favorable mathematical structure, as it is the case for formulation \eqref{eq:formulation1b}. Indeed, minimizing the error function in \eqref{eq:formulation1b} with respect to only the leaf node variables $\bm{\beta}$ (keeping fixed the branch node variables $\bm{\w}$) amounts to a convex problem, namely the Linear Least Squares Problem (LLSP). Preserving the separation between ``convex variables" ($\bm{\beta}$) and ``nonconvex variables" ($\bm{\w}$) in the decomposition subproblems proved to be effective in escaping from poor-quality solutions and in accelerating convergence for other ML models (see e.g., \cite{grippo2015decomposition} for multilayer perceptrons and \cite{buzzi2001convergent} for Radial Basis Function neural networks). 
 
In this section we present a
NOde-based DEComposition General Scheme, referred to as NODEC-GS, for solving the SRT training formulation \eqref{eq:formulation1b}, which exploits the intrinsic tree structure. NODEC-GS is based on a node-based working set selection procedure preserving the separation between branch node variables and leaf node variables, and encompasses different practical versions.  After a detailed description of the general scheme, we present asymptotic convergence guarantees under suitable conditions.

\subsection{Node-based decomposition scheme}\label{subsec:scheme}
We consider the following $\ell_2$-regularized version of the training formulation \eqref{eq:formulation1b}
\begin{equation}\label{eq:obj}
\min_{\bm{\w},\bm{\beta}} E(\bm{\w,\beta})= \frac{1}{N}\sum_{i \in I}\left(\sum_{t\in \tau _{L}}P_{it}(\bm{\w})(\hat y_{it}(\bm{\beta}_t)-y_{i})^{2}\right) + \frac{\lambda_{\bm{\w}}}{2} \|\bm{\w} \|^2+ \frac{\lambda_{\bm{\beta}}}{2} \|\bm{\beta}\|^2,
\end{equation}
where $\lambda_{\bm{\w}},\lambda_{\bm{\beta}}\geq 0$ are the regularization hyperparameters. Recall that the variable vectors $\bm{\w}_t \in \mathbb{R}^{p+1}$ for $t \in \tau_B$ and $\bm{\beta}_t \in \mathbb{R}^{p+1}$ for $t \in \tau_L$ can be rearranged into a matricial form as $\bm{\w} =(\bm{\w}_1 \,\dots\,\bm{\w}_{|\tau_B|}) \in \mathbb{R}^{(p+1) \times |\tau_B|}$ and $\bm{\beta} =(\bm{\beta}_{1}\,\dots\,\bm{\beta}_{|\tau_L|}) \in \mathbb{R}^{(p+1)\times |\tau_L|}$. Accordingly, the components of the whole gradient of the error function $\nabla E(\bm{\w,\beta})$ associated to the variables $\bm{\w}$ and $\bm{\beta}$, denoted by $\nabla_{\bm{\w}} E(\bm{\w,\beta})$ and $\nabla_{\bm{\beta}} E(\bm{\w,\beta})$, can be rearranged into matricial form as $\nabla_{\bm{\w}} E(\bm{\w,\beta})=(\nabla_{\bm{\w}_1} E(\bm{\w,\beta})\,\dots\,\nabla_{\bm{\w}_{|\tau_B|}} E(\bm{\w,\beta})) \in \mathbb{R}^{(p+1)\times |\tau_B|}$ and $\nabla_{\bm{\beta}} E(\bm{\w,\beta})=(\nabla_{\bm{\beta}_1} E(\bm{\w,\beta})\,\dots\,\nabla_{\bm{\beta}_{|\tau_L|}} E(\bm{\w,\beta})) \in \mathbb{R}^{(p+1)\times |\tau_L|}$, respectively. More generally, given a subset of branch (leaf) nodes $J \subseteq \tau_B$ ($J \subseteq \tau_L$) and its complement $\bar{J}$, the variable vectors $\bm{\w}_t$ for $t \in \tau_B$ ($\bm{\beta}_t$ for $t \in \tau_L$) can be rearranged as $(\bm{\w}_J,\bm{\w}_{\bar J})$  ($(\bm{\beta}_J,\bm{\beta}_{\bar J})$), and the gradient components associated to $J$ are denoted as $\nabla_J E(\bm{\w,\beta})$. Variables $\bm{\w}$, $\bm{\beta}$, gradient $\nabla E(\bm{\w,\beta})$, and all their subcomponents can be also considered in a flattened vectorial form when needed.
\par 
At each iteration $k$ of the proposed general decomposition scheme, the working set consists of the collection of indices of a subset of nodes. For any subset of nodes, the working set is divided into a branch node working set $W^k_B\subseteq \tau_B$ and a leaf node working set $W^k_L \subseteq \tau_L$. Whenever a node index $t$ is included in $W^k_B\subseteq \tau_B$ ($W^k_L \subseteq \tau_L$), all the associated variables $\bm{\w}_t$ ($\bm{\beta}_t$) are simultaneously considered in the corresponding subproblem.
NODEC-GS, which is reported below, first optimizes \eqref{eq:obj} with respect to the branch node variables whose node indices are in $W^k_B$ ({\bf BN Step}), and then with respect to the leaf node variables whose node indices are in $W^k_L$ ({\bf LN Step}).
\medskip
{ 
\par\medskip
\hrule
\smallskip
 \hspace{-12pt}{\bf NODEC-GS} Decomposition general scheme 
\smallskip
\hrule
\par\medskip\noindent
{Initialization:} set $k=0$, determine an initial solution\par
\centerline{$\bm{\w}^0 =({\bm{\w}^0}_1\,\dots\,{\bm{\w}^0}_{|\tau_B|})\in \mathbb{R}^{(p+1) \times |\tau_B|},\quad \bm{\beta}^0 =({\bm{\beta}^0}_{1}\,\dots\,{\bm{\beta}^0}_{|\tau_L|})\in \mathbb{R}^{(p+1)\times |\tau_L|} $}
\par\smallskip\noindent
{\textbf{While}} {\it termination test} is not satisfied \textbf{do}
\par\smallskip\noindent
\textcolor{white}{ciaoo} $\bullet$ choose $W^k_B\subseteq\tau_B, \; W^k_L\subseteq\tau_L$
\vspace{-6pt}
\par\smallskip\noindent
\begin{description}
\item[\textcolor{white}{}$\rhd$ {\bf BN Step}] {(Optimize \eqref{eq:obj} with respect to the Branch Node variables $\bm{\w}_{W^k_B}$)}
\par\smallskip\noindent
$\bullet$ If $ W_{B}^k\equiv \emptyset$ set $\bm{\w}^{k+1}=\bm{\w}^k$, otherwise starting from $(\bm{\w}^{k},\bm{\beta}^k)$ minimize $E(\bm{\w},\bm{\beta})$ in \eqref{eq:obj} with respect to all
$\bm{\w}_t$ for $t\in W^k_B$ determining $\bm{\w}^*_t$ for $t\in W^k_B$.\\
$\bullet$ Derive $(\bm\w^{k+1},\bm\beta^k)$ by setting  $\bm{\w}$ as $\bm{\w}_t^{k+1}=\bm{\w}^*_t$ for $t\in W^k_B$ and $\bm{\w}_t^{k+1}=\bm{\w}_t^k$ for $ t\notin W^k_B$.
\vspace{-25pt}

\par\smallskip\noindent
 \item[\textcolor{white}{}$\rhd$ {\bf LN Step}] {(Optimize \eqref{eq:obj} with respect
to the Leaf Node variables $\bm{\beta}_{W^k_{L}}$ - regression problem)}
\par\smallskip\noindent
$\bullet$ If $ W_{L}^k\equiv \emptyset$ set $\bm{\beta}^{k+1}=\bm{\beta}^k$, otherwise starting from $(\bm{\w}^{k+1},\bm{\beta}^k)$ minimize $E(\bm{\w},\bm{\beta})$ in \eqref{eq:obj} with respect to all
$\bm{\beta}_t$ for $t\in W^k_L$ determining $\bm{\beta}^*_t$ for $t\in W^k_L$. \\
$\bullet$ Derive $(\bm\w^{k+1},\bm\beta^{k+1})$ by setting $\bm{\beta}$ as $\bm{\beta}_t^{k+1}=\bm{\beta}^*_t$ for $t\in W^k_L$ and $\bm{\beta}_t^{k+1}=\bm{\beta}_t^k$ for $ t\notin W^k_L$.

\end{description}

\vspace{-2pt}
\par\smallskip  \hspace{10pt}$\bullet$ Set $k=k+1$.
 
 \medskip
\hrule
 \medskip
 
 }

\bigskip

NODEC-GS is very general and may encompass many different versions by specifying: the termination test (commonly a tolerance on the norm of $\nabla E(\bm{\w,\beta})$ or a maximum number of iterations), the working set selection rule, the way to update the branch node variables in the {\bf BN Step}, and the way to update the leaf node variables in the {\bf LN Step}.

A first requirement is to consider a node-based working set selection procedure where all variables associated to a certain node are optimized whenever such node index is inserted in the working set. In principle, one might consider working sets including only a subset of variables associated to a node of the tree, but these schemes will be not addressed here. Notice also that, although we assume a sequential ordering between {\bf BN Step} and {\bf LN Step} in the scheme, the possibility of setting $ W_{B}^k\equiv \emptyset$ or $ W_{L}^k\equiv \emptyset$ for any $k$, implies no actual ordering between the two minimization steps. 

The second important requirement is to maintain the separation between branch node variables and leaf node variables in order to exploit the favorable structure of the error function $E(\bm{\w,\beta})$. In the {\bf LN Step}, which is a convex LLSP with respect to variables $\bm{\beta_t}$ for $t \in  W_{L}^k$ (strictly convex when $\lambda_{\beta}>0$), the global minimizer can be determined exactly or approximately via efficient methods (see e.g. \cite{Bertsekas/99}). In the {\bf BN Step} the subproblem is nonconvex with respect to variables $\bm{\w}$ and we settle for a solution satisfying some given optimality conditions or we early stop at an approximate solution.
\textcolor{black}{Notice that the separation between branch node variables and leaf node ones is naturally facilitated by the error function in \eqref{eq:formulation1b} of our SRT formulation, where each leaf node contributes with an independent output and, hence, an independent squared error term, and the branch nodes affect the weights (probabilities $P_{it}$) of each error term.}


Two extreme NODEC-GS versions can be considered depending on the degree of  decomposition. The lowest decomposition degree is characterized by setting $W_{B}^k= \tau_{B}$ and $W_{L}^k= \tau_{L}$ for all $k$, resulting in an alternating minimization with respect to $\bm{\w}$ and $\bm{\beta}$ subsets of variables. The highest decomposition degree is obtained by optimizing over the variables associated to a single node, either $W_{B}^k=\{t\} \mbox{ with } t \in \tau_{B}\, \text{and}\, W_{L}^k= \emptyset$, or $ W_{B}^k= \emptyset \, \text{and}\,W_{L}^k=\{t\} \mbox{ with } t \in \tau_{L}$. Clearly, a wide range of intermediate NODEC-GS versions can be devised.



In Section \ref{subsec:implementation}, we present an intermediate version of NODEC-GS including a clustering-based initialization procedure and a Data points Reassignment heuristic, referred to as NODEC-DR. NODEC-DR is particularly suited to the error function in \eqref{eq:obj} and the tree structure. 



\subsection{Asymptotic convergence}\label{subsec:convergence}
In this section, we present the asymptotic convergence guarantees for NODEC-GS scheme, which is inspired by the ones proved in  \citep{grippo2015decomposition} for multi-layer perceptrons, and based on the theory developed in  \citep{grippof1999globally}. The details on the convergence analysis are reported in  \ref{appendix:proofs_dec1}.

To establish convergence of NODEC-GS towards stationary points, we first observe that if the hyperparameters $\lambda_{\bm\w}\,\text{and}\,\lambda_{\bm\beta}$ are strictly positive, the error function in \eqref{eq:obj} is not only continuous but also coercive and the problem of minimizing it is well defined. Indeed, all the level sets of $E(\bm\w,\bm\beta)$, defined for any $(\bm\w^0,\bm\beta^0)$ as ${\cal L}_0=\{(\bm\w,\bm\beta) \in \mathbb{R}^{(p+1) \times (|\tau_B|+|\tau_L|)} :E(\bm\w,\bm\beta)\leq E(\bm\w^0,\bm\beta^0)\}$, are compact and then the objective function in \eqref{eq:obj} admits a global minimizer.

Three conditions, which are relevant from both theoretical and practical point of views, must be satisfied to guarantee the convergence of an infinite sequence $\{(\bm\w^k,\bm\beta^k)\}$ generated by NODEC-GS.


The first condition requires that each branch node or leaf node is periodically inserted in the working set. This can be easily taken into account when devising the working set selection procedure.

\begin{condition}\label{con:1}
Assume that NODEC-GS generates an infinite sequence $\{(\bm\w^k,\bm\beta^k)\}$. Then there exists $R>0 \in \mathbb{N}$ such that, for every $j \in \tau_B$, for every $i \in \tau_L$, and for every $k\geq 0$, there exist iteration indices $s,s'$, with $k \leq s \leq k+R$ and $k \leq s' \leq k+R$, such that $j \in W_{B}^s$ and $i \in W_{L}^{s'}$. 
\end{condition}

The second condition is related to the {\bf BN Step} and requires that the partial update $(\bm\w^{k+1},\bm\beta^{k})$ is not worse than $(\bm\w^{k},\bm\beta^{k})$ in terms of error function value. Moreover, for an infinite subsequence in which the index $j$ of a given subset of variables $\bm\w_j$ is inserted in the branch node working set $W_{B}^k$, the difference between successive iterations must tend to zero.  Simultaneously, it is required that (at least) in the limit the first order optimality conditions are satisfied by the gradient components associated to the $\bm{\w}_j$ variables.\par 
\begin{condition}\label{con:2}
    For every positive integer $k$ we have:
    \begin{equation}\label{eq:descent-w}
        E(\bm\w^{k+1},\bm\beta^k)\leq E(\bm\w^{k},\bm\beta^k).
    \end{equation}
    Moreover, for every $j \in \tau_B$ and for every infinite subsequence indexed by $K$ such that $j \in W_{B}^k$ with $k\in K$, we have that:
    \begin{equation}\label{eq:proximal-w}
        \lim\limits_{k \to \infty, k \in K} \|\bm\w_j^{k+1}-\bm\w_j^k\|=0,
    \end{equation}
    \begin{equation}\label{eq:convergence-w}
        \lim\limits_{k \to \infty, k \in K} \nabla_{\bm\w_j}E(\bm\w^{k},\bm\beta^k)=0.
    \end{equation}
\end{condition}

The third condition is concerned with the {\bf LN Step} and is similar to Condition \ref{con:2}. For an infinite subsequence in which the index $j$ of a given subset of variables $\bm\beta_j$ is inserted in the leaf node working set $W_{L}^k$, the condition requires that, at least in the limit, a convergent algorithm is applied.

\begin{condition}\label{con:3}
    For every positive integer $k$ we have:
    \begin{equation}\label{eq:descent-b}
        E(\bm\w^{k+1},\bm\beta^{k+1})\leq E(\bm\w^{k+1},\bm\beta^k).
    \end{equation}
    Moreover, for every $j \in \tau_L$ and for every infinite subsequence indexed by $K$ such that $j \in W_{L}^k$ with $k\in K$, we have that:
    \begin{equation}\label{eq:convergence-b}
        \lim\limits_{k \to \infty, k \in K} \nabla_{\bm\beta_j}E(\bm\w^{k+1},\bm\beta^{k+1})=0.
    \end{equation}
\end{condition}

Notice that Conditions \textsc{\ref{con:2}} and  \textsc{\ref{con:3}} are similar, but the latter is less restrictive than the former. Indeed, due to the strict convexity of the {\bf LN Step} optimization problem (since $\lambda_{\bm\beta}>0$), the requirement that the distance between successive iterates tends to zero (as in \eqref{eq:proximal-w} of Condition \ref{con:2}) is automatically satisfied. The reader is referred to \ref{appendix:proofs_dec1} for details. 

Based on the above conditions, in \ref{subsec:ec-1-ndgs} we prove the following asymptotic convergence result together with two technical lemmas required in the proof.

\begin{proposition}\label{prop:prop1}
Suppose that NODEC-GS generates an infinite sequence
 $\{(\bm\w^k, \bm\beta^k)\}$ and that Conditions \ref{con:1},  \ref{con:2} and \ref{con:3}  are
 satisfied. Then
\begin{description}
 \item[(i)] $\{(\bm\w^k, \bm\beta^k)\}$ has limit points,  \hspace{0.3cm} {\bf(ii)}  $\{E(\bm\w^k, \bm\beta^k)\}$ converges to a limit, 
 \item[(iii)] $\lim\limits_{k\to\infty}\|\bm\w^{k+1}-\bm\w^k\|=0$, \hspace{0.8cm} {\bf (iv)}$\lim\limits_{k\to\infty}\|\bm\beta^{k+1}-\bm\beta^k\|=0$,
 \item[(v)] every limit point of $\{(\bm\w^k,
\bm\beta^k)\}$ is a stationary point of $E(\bm\w,
\bm\beta)$ in \eqref{eq:obj}.
\end{description}
\end{proposition}
\section{Practical version of the node-based decomposition algorithm}\label{subsec:implementation}

In this section, we present a practical version of NODEC-GS, 
referred to as NODEC-DR, which is asymptotically convergent towards stationary points.
\textcolor{black}{Since the SRT training process can be affected by the choice of the initial solution, NODEC-DR includes a clustering-based initialization procedure. Moreover, to favor a balanced routing of the data points among most of the root-to-leaf-node paths and hence to better exploit the representative power of the soft regression tree, we also devised an ad hoc heuristic for reassigning the data points across the tree}\footnote{The reader is referred to \citep{consolo2023sparse} for experiments indicating the incremental impact of the clustering-based initialization and the reassignment heuristic.}.
\textcolor{black}{After some important preliminary discussions on the {\bf BN Step} in Section \ref{subsec:preliminaries}, we describe NODEC-DR, the data points reassignment heuristic, and the initialization procedure in the next three subsections.  
}


\textcolor{black}{\subsection{Preliminary discussion on the \textbf{BN Step}}\label{subsec:preliminaries}
Before presenting the NODEC-DR algorithm, it is necessary to comment on a crucial aspect concerning the nonconvex \textbf{BN Step}. To prevent oscillating behaviour of the decomposition method and to ensure asymptotic convergence it is important to satisfy \textbf{Condition} \ref{con:2} (see Section \ref{subsec:convergence}) in the \textbf{BN Step} update. This can be easily achieved by using the partial steepest descent direction in tandem with the well-known Armijo step length update (see e.g., \cite{Bertsekas/99}). Given a current working set $W^k_B$, we define the next iterate as}

\begin{equation}
\label{eq:steepest}
  \bm{\w}_{W^k_B}^{k+1} = \bm{\w}_{W^k_B}^{k} + \alpha^k_{W^k_B} \bm{d}_{W^k_B}^k,
\end{equation}

\noindent where $\bm{d}_{W^k_B}^k = - \nabla_{\bm{\w}_{W^k_B}}E(\bm{\w}^k,\bm{\beta}^k)$, and the step length $\alpha^k_{W^k_B}$ satisfies the following Armijo condition:
\begin{equation}\label{eq:armijo}
    E(\bm{\w}_{W^k_B}^{k} + \alpha^k_{W^k_B} \bm{d}_{W^k_B}^k,\bm{\w}_{\overline W^k_B}^{k},\bm{\beta}^k) \leq E(\bm{\w}^k,\bm{\beta}^k) -\gamma \alpha^k_{W^k_B} \nabla_{\bm{\w}_{W^k_B}}E(\bm{\w}^k,\bm{\beta}^k)^T \bm{d}_{W^k_B}^k
\end{equation}
for a given control parameter $\gamma \in (0,1)$.

\textcolor{black}{As detailed in \ref{subsec:ec-1-armjio}, it is possible to show that this adaptation of the Armjio update to train SRTs, inherits some useful theoretical properties proved in \citep{grippo2015decomposition} for training multilayer perceptrons. Although, in the \textbf{BN Step}, $\bm{\w}_{W^k_B}^{k+1}$ can be determined via any unconstrained optimization method, the Armijo update acts as a reference update.} 

\textcolor{black}{Let us denote by $\bm{\w}^{ref}_{W^{k}_B}$ the Armijo update computed according to \eqref{eq:steepest} and \eqref{eq:armijo}, and by $\hat{\bm{\w}}_{W^k_B}$ the candidate update obtained by the adopted method. If we set $\bm{\w}_{W^k_B}^{k+1}=\hat{\bm{\w}}_{W^k_B}$, then convergence is ensured provided that the following two conditions are satisfied:  
\begin{equation}
\label{eq:descreasing1}
E(\hat{\bm{\w}}_{W^k_B}^k,\bm{\w}_{\bar W^k_B}^k,\bm{\beta}^k) \leq E(\bm{\w}^{ref}_{W^{k}_B},\bm{\w}_{\bar W^k_B}^{k},\bm{\beta}^k)
\end{equation}
\begin{equation}
\label{eq:decreasing2}
    E(\hat{\bm{\w}}_{W^k_B}^k,\bm{\w}_{\bar W^k_B}^k,\bm{\beta}^k) \leq E(\bm{\w}^k,\bm{\beta}^k) - \tau \|\hat{\bm{\w}}_{W^k_B}^k - \bm{\w}_{W^k_B}^k \|^2
\end{equation}
for a given $\tau > 0$. Condition \eqref{eq:decreasing2} is necessary to guarantee that the convergence of the values of the error function $E(\bm{\w}^k,\bm{\beta}^k)$  implies that $\| \bm{\w}_{W^k_B}^{k+1} - \bm{\w}_{W^k_B}^{k}\| \to 0$ when $k \rightarrow \infty$. For the Armijo update $\bm{\w}^{ref}_{W^k_B}$ this is automatically guaranteed by point $a)$ of statement $(ii)$ in \textcolor{black}{Proposition 2 which is reported and proved in \ref{subsec:ec-1-armjio}.}}

\subsection{NODEC-DR and its asymptotic convergence}\label{subsec:algorithm}
In this section we present the NODEC-DR algorithm whose pseudocode is reported below.


\textcolor{black}{After determining the initial solution $(\bm{\w}^{0},\bm{\beta}^{0})$ using the initialization procedure described in Section \ref{sec:Init_procedure}, NODEC-DR consists of a main external loop of macro iterations indexed by $it$ and repeated until an appropriate termination criterion is satisfied.} Each macro iteration $it$ consists of an internal loop of inner iterations indexed by $k$.

\begin{figure}[H]
    \centering
    \includegraphics[width=0.6\textwidth]{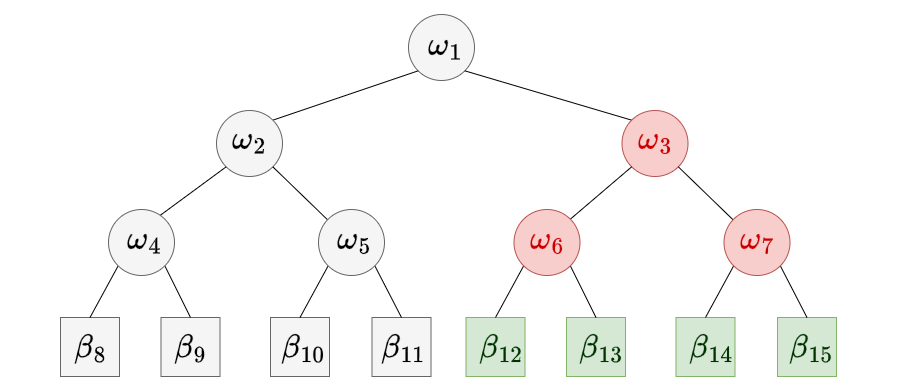}
    \caption{Example of NODEC-DR working set selection for an SRT of depth $D=3$. The branch node $t=3$ and the corresponding working sets $W_B=\{3,6,7\}$ (in red) and $W_L = \{12,13,14,15\}$ (in green) are selected. The associated variable vectors are indicated inside each node.}
    \label{fig:tree_dec1}
\end{figure}

At each inner iteration $k$ one branch node $t\in \tau_B$ is selected and the corresponding working sets $W^k_B \subseteq \tau_B$ and $W^k_L \subseteq \tau_L$ are constructed.
The index $t$ and all indices of its descendant branch nodes, denoted as ${\cal D}_B(t)$, are inserted in $W^k_B$, while all indices of decsendant leaf nodes, denoted as ${\cal D}_L(t)$, are inserted in $W^k_L$. Figure \ref{fig:tree_dec1} shows an example of SRT of depth $D=3$ in which the branch node $t=3$ is selected and the resulting working sets are $W^k_B=\{3,6,7\}$ in red and $W^k_L = \{12,13,14,15\}$ in green. When $t$ is the root node and $D > 1$, we set $W^k_B = \{t\}$ and $W_L = \emptyset$. Once the working sets $W^k_B$ and $W^k_L$ have been determined, the \textbf{BN Step} is performed before the \textbf{LN Step}.

\textcolor{black}{Regarding the $\textbf{BN Step}$, the subvector $\bm{\w}_{W^k_B}$ is actually updated only if $\| \nabla_{{W^k_B}}E(\bm{\w}^k,\bm{\beta}^k)\|$ is larger than a certain threshold which, for convergence purposes, tends to 0 when $k \rightarrow \infty$. At iteration $k$ the threshold value is $(\theta_{\w})^k$ with $0<\theta_{\w}<1$. Whenever $\| \nabla_{{W^k_B}}E(\bm{\w}^k,\bm{\beta}^k)\| > (\theta_{\w})^k$, the reference update $\bm{\w}_{W^k_B}^{ref}$ and the candidate update $\hat{\bm{\w}}_{W^k_B}$  are computed using, respectively, the Armijo method and the \textsc{UpdateBranchNode} procedure described in detail in Subection \ref{subsec:reass}.} 
\textcolor{black}{In particular, in \textsc{UpdateBranchNode} the subvector $\hat{\bm{\w}}_{W^k_B}$ can be determined by any standard unconstrained minimization algorithm or a specific heuristic.} 


\textcolor{black}{Due to the nonconvexity of the {\bf BN Step} problem, standard optimization algorithms may provide poor-quality solutions where most of the data points are routed along their HBP paths to a small subset of leaf nodes. To better exploit the expressiveness of the regression tree, the heuristic in the \textsc{UpdateBranchNode}  procedure aims at balancing the routing of data points among the root-to-leaf-node paths. Roughly speaking, the goal is to modify the routing at each branch node of a selected subset of data points so as to better balance the number of data points which are following its two branches. As described below, three positive thresholds $\varepsilon_1, \varepsilon_2 \,\text{and}\, \varepsilon_3$ are used to detect the extent of data points imbalance across the tree and to improve stability during the initial macro iterations of NODEC-DR.
While $\varepsilon_1$ and $\varepsilon_2$ account for the imbalance (high and low) level,  $\varepsilon_3$ determines the fraction of data points to be assigned to the other branch.
At the end of each inner iteration of NODEC-DR, the thresholds $\varepsilon_1, \varepsilon_2 \,\text{and}\, \varepsilon_3$  are decreased. The smaller their values the more data points imbalance is allowed.} 

\textcolor{black}{To ensure convergence, we select the subvector $\hat{\bm{\w}}_{W^k_B}$ for sufficiently large $k$ (i.e., $k > k_0$ where $k_0$ is a positive integer set by the user) only if it satisfies Conditions \eqref{eq:descreasing1} and \eqref{eq:decreasing2}, otherwise $\bm{\w}_{W^k_B}^{ref}$ is selected \textcolor{black}{(see Steps 17-20 of NODEC-DR)}. 
See Section \ref{subsec:reass} for the details of the \textsc{UpdateBranchNode} procedure.}

\renewcommand{\thealgorithm}{NODEC-DR} 
\begin{algorithm}[H]
\floatname{algorithm}{} 
\caption{- Decomposition algorithm}
\label{alg:Decomposition1}
$\textbf{Input:}\mbox{ }\text{depth} \,D; \text{dataset} \,I;\,\text{max}\,\text{number}\,\text{of}\,\text{iterations}\,M\_it; \varepsilon_{1}^0,\,\varepsilon_{2}^0,\,\varepsilon_{3}^0,\,\bm{\zeta},\,\theta_{\w},\, \theta_{\beta} \in \mbox{[0,1)}\,\mbox{with}$ 
$ \\ \mbox{\quad\quad\quad\, } \varepsilon^{0}_{1} >\varepsilon^{0}_{2};\,\upsilon \in [0,1);\, r \in \mathbb{N};\, k_0 \in \mathbb{N}, \tau>0$ 
$\\ \textbf{Output:}\, (\bm{\w^{*}},\,\bm{\beta^{*}})$
\vspace{-10pt}
  {\fontsize{9}{10}\selectfont
  \begin{algorithmic}[1]
  \Statex {} 
    \State $(\bm{\w}^{0},\bm{\beta}^{0}) \gets\textsc{InitializationProcedure}(I,D,r)$
    \Procedure{Decomposition}{$D,I,(\bm{\w^0},\bm{\beta^0}),\varepsilon_{1},\varepsilon_{2},\varepsilon_{3},\bm{\gamma},\theta_{\w},\theta_{\beta},\upsilon,\tau$}
      
      \State $\bullet$ $(\bm{\w}^{*},\bm{\beta}^{*}) \gets \bm{\w}^{0},\bm{\beta}^{0}$
      \State $\bullet$ $\varepsilon_{1},\,\varepsilon_{2},\,\varepsilon_{3}\gets \varepsilon_{1}^0,\,\varepsilon_{2}^0,\,\varepsilon_{3}^0$
      \State $\bullet$ $error_{best} \gets E(\bm{\w}^{*},\bm{\beta}^{*})$
      \State $\bullet$ $k \gets 0,\, it \gets 1,$
      \While{$ it < M\_it\,and\,$ not {\it termination criterion}}
       
        \For{$ t = 1,...,2^{D}-1 $}\Comment{loop over the branch nodes}
        
        \If{$t == 1\,\text{and}\, D > 1$}
        \State $W^k_B \gets \{t\}, \, W^k_L \gets \emptyset$
        \Else
        \State $W^k_B \gets \{t,{\cal D}_{B}(t)\}, \, W^k_L \gets {\cal D}_{L}(t)$
        \EndIf
        
        \Statex  {$\triangleright$ \, \textbf{BN Step}} \textit{(Optimization with respect to branch nodes parameters)}
        
        \If{$\| \nabla_{{W^k_B}}E(\bm{\w}^k,\bm{\beta}^k) \| \leq (\theta_{\w})^k$}
        \State $\bm{\w}^{k+1}\gets \bm{\w}^{k}$
        \Else
        \State $\bullet$ $\alpha_{W^k_B}^k \gets \textsc{armijoupdate}(a,\gamma,\delta,(\bm{\w}^{k},\bm{\beta}^k),W^k_B)$
        \State $\bullet$ $\bm{\w}_{W^k_B}^{ref} \gets \bm{\w}_{W^k_B} - \alpha_{W^k_B}^k \nabla_{W^k_B}E(\bm{\w}^k,\bm{\beta}^{k})$ \Comment{reference update found via partial steepest descent direction and Armijo step length update}
        \State $\bullet$ $\hat{\bm{\w}}_{W^k_B} \gets $\textsc{UpdateBranchNode}$(W^k_B,\bm{\w}^{k}_{W^k_B},\varepsilon_{1},\varepsilon_{2},\varepsilon_{3})$ 

        \If{($\hat{\bm{\w}}_{W^k_B}$ satisfies the conditions \eqref{eq:descreasing1}-\eqref{eq:decreasing2}) or ($k \leq k_0 $)} 
            \State $\bm{\w}_{W^k_B}^{k+1} \gets \hat{\bm{\w}}_{W^k_B}$ \Comment{select candidate update found using heuristic}
        \Else
            \State $\bm{\w}_{W^k_B}^{k+1} \gets \bm{\w}_{W^k_B}^{ref}$ \Comment{select reference update}
        \EndIf

        \EndIf

        \Statex  {$\triangleright$ \, \textbf{LN Step}} \textit{(Optimization with respect to leaf nodes parameters)}
        
        \If{$\| \nabla_{{W^k_L}}E(\bm{\w}^{k+1},\bm{\beta}^k) \| \leq (\theta_{\beta})^k$}
        \State $\bm{\beta}^{k+1}\gets \bm{\beta}^{k}$
        \Else
        \textcolor{black}{\If{$k \leq k_0 $} 
            \State compute $\bm{\beta}_{W^k_L}^{k+1} $ by any method
        \Else
            \State compute $\bm{\beta}_{W^k_L}^{k+1}$ such that \\ \qquad \qquad \qquad \;\;\;\, $E(\bm{\w}^{k+1},\bm{\beta}^{k+1})\leq E(\bm{\w}^{k+1},\bm{\beta}^{k})\; \text{and} \; \| \nabla_{W^k_L}E(\bm{\w}^{k+1},\bm{\beta}^{k+1}) \| \leq (\upsilon)^k$
        \EndIf
        }
        
        
        \EndIf
        \If{$E(\bm{\w}^{k+1},\bm{\beta}^{k+1})< error_{best}$}
        \State $\bullet$ $error_{best} \gets E(\bm{\w}^{k+1},\bm{\beta}^{k+1})$
        \State $\bullet$ ($\bm{\w}^{*},\bm{\beta}^{*}) \gets (\bm{\w}^{k+1},\bm{\beta}^{k+1})$
        \EndIf
        \State $k \gets k + 1$
        \EndFor
	  \State $\bullet$ $\varepsilon_{1},\,\varepsilon_{2},\,\varepsilon_{3} \gets \bm{\zeta}\,\varepsilon_{1},\,\bm{\zeta}\,\varepsilon_{2},\,\bm{\zeta}\,\varepsilon_{3}$
	  \State $\bullet$ $it \gets it + 1$
      \EndWhile\label{Decompositiondwhile}
      \State \textbf{return} $(\bm{\w}^{*},\bm{\beta}^{*})$
    \EndProcedure
  \end{algorithmic}
  }
\end{algorithm}


\textcolor{black}{As to the $\textbf{LN Step}$, the subvector $\bm{\beta}_{W^k_L}^{k}$ is actually updated only if $\| \nabla_{{W^k_L}}E(\bm{\w}^{k+1},\bm{\beta}^k) \|$ is greater than $(\theta_{\beta})^k$ and, as previously mentioned, the associated LLSP is solved via an iterative method for convex quadratic problems or by a direct method.}




\textcolor{black}{Notice that in both the \textbf{BN Step} and \textbf{LN Step}, whenever $k \leq k_0$, no specific restrictions are imposed on the updates $\bm{\w}_{W^k_B}^{k+1}$ and $ \bm{\beta}_{W^k_L}^{k+1}$. 
When solving the subproblems with respect to the variables $\bm{\w}_{W^k_B}^{k}$ and, respectively, $ \bm{\beta}_{W^k_L}^{k}$, we may take into account only a subset of data points, consider a proxy of the subproblem or both. This can speed up the first macro iterations. } 

\textcolor{black}{In NODEC-DR we adopt two strategies to enhance the efficiency \textcolor{black}{and reduce the computational time} of the \textbf{BN Step} and the \textbf{LN Step}. For any selected node $t$, in the proxy subproblems we consider: i) only the subtree rooted at $t$ so that the corresponding error function neglects all the variables $\bm{\w}$ and $\bm{\beta}$ (and probabilities $p_{it}$) outside the subtree, ii) only the data points whose HBP path contains $t$, which are likely to be more ``relevant" in the optimization over the variables whose indices belongs to $W_B^k$ and $W_L^k$.}


In \ref{subsec:prop3} we prove the following convergence result for NODEC-DR, by showing that it satisfies the convergence conditions for the general scheme NODEC-GS. 
\setcounter{proposition}{2}
\begin{proposition}
\label{prop:prop3}
Given an infinite sequence $\{(\bm{\w}^k,\bm{\beta}^k)\}$ generated by NODEC-DR, we have that:
\begin{description}
    \item[(i)] $\{(\bm{\w}^k,\bm{\beta}^k)\}$ has a limit point, \hspace{0.2cm} {\bf (ii)}  $\{E(\bm{\w}^k,\bm{\beta}^k)\}$ converges to a limit when $k \rightarrow \infty$,
    \item[(iii)]  $\lim\limits_{k \to \infty}\|\bm{\beta}^{k+1} -\bm{\beta}^k\|=0 $, \hspace{0.5cm} {\bf(iv)} For any $ t \in \tau_B$, $\lim\limits_{k \to \infty}\| \bm{\w}_{t}^{k+1} - \bm{\w}_{t}^{k} \|=0$,
    \item[(v)] Every limit point of $\{(\bm{\w}^k,\bm{\beta}^k)\}$ is a stationary point of $E(\bm\w,
\bm\beta)$ in \eqref{eq:obj}.
\end{description}
\end{proposition}

\textcolor{black}{We conclude with two remarks concerning the proxy subproblems and the decompostion strategy of NODEC-DR.}

\textcolor{black}{It is worth pointing out that focusing on the subtree in the proxy subproblems of the {\bf BN Step} and {\bf LN Step} amounts to setting to 1 all the probabilities $p_{it}$ associated to the branches on the corresponding HBP path from the root to node $t$ that do not belong to the subtree. Since the total probabilities $P_{it}$ 
can differ from the ones in the full tree, the differences between the error function of the proxy subproblem and the overall error function \eqref{eq:obj} lie not only 
in the number of terms, corresponding to the data point residuals, but also in the associated multiplicative factors $P_{it}$.}


\textcolor{black}{Finally, note that the above-mentioned subproblems optimization over a subset of data points differs from the mini-batch data point-based decomposition extensively used in training other type of ML models (see e.g. \cite{bottou2010large}). For any node $t$ the number of HBP paths (associated to data points) containing $t$ may substantially change over the iterations.}






\subsection{{\bf BN Step} optimization with data points reassignment heuristic}\label{subsec:reass}

\textcolor{black}{In the \textbf{BN Step} of NODEC-DR the new candidate update $\hat{\bm{\w}}_{W_B^k}$ is computed using the \textsc{UpdateBranchNode} procedure whose scheme is reported below. 
\textsc{UpdateBranchNode} includes a data points reassignment heuristic which deterministically reroutes data points across the tree by partially modifying the corresponding HBP paths. As previously mentioned, the goal is to avoid early convergence to poor degenerate solutions where a large number of data points are deterministically routed towards a few leaf nodes.
 }



\renewcommand{\thealgorithm}{Update Branch Node} 
\begin{algorithm}[H]
\floatname{algorithm}{} 
\caption{- Procedure with data points reassignment heuristic}
\label{alg:updatebranching}
$\textbf{Input:} \ \text{dataset} \ I; \ W^k_B\, \text{working} \, \text{set};\, \text{previous}\;\text{update}\, \bm{\w}^{k-1}_{W^k_B};\,\varepsilon_{1},\,\varepsilon_{2},\,\varepsilon_{3} \in (0,1),\, \varepsilon_1 > \varepsilon_2\\$ 
$\textbf{Output:}\,\text{optimal}\, \text{values}\; \hat{\bm{\w}}_{W^k_B}$
  {\fontsize{9}{10}\selectfont
  \begin{algorithmic}[1]
    
    \Procedure{UpdateBranchNode}{$I$,$W^k_B,\bm{\w}^{k-1}_{W^k_B},\varepsilon_{1},\varepsilon_{2},\varepsilon_{3}$}
        \State $t\gets$  $\min \{ {\hat t}\ | \;{\hat t}\in W^k_B\}$ \Comment{t is the ancestor of all the other nodes in $W^k_B$}
        \State $I_t\gets$  $\{ (\textbf{x}_i,y_i) \in I |$ \text{HBP path of} $(\textbf{x}_i,y_i)$ \text{contains} $t \}$ \Comment{data points of $I_t$ deterministically falling into $t$}
        
        \If{($\frac{N_{left_{t}}}{N_{t}} \leq \varepsilon_{1}\; \text{or}\; \frac{N_{left_{t}}}{N_{t}} \geq 1 - \varepsilon_{1}) \;\text{and}\;({\varepsilon_1}N \geq 1)$ }\Comment{imbalanced data points routing}
        \State $\hat{\bm{\w}}_{W^k_B \setminus t} \gets \bm{\w}^{k-1}_{W^k_B \setminus t}$
        \State \textbf{for} $(\textbf{x}_i,y_i) \in I_t$  \textbf{do} 
        \State $\;\quad$\textbf{if} $\mathbf{x}_i$ is routed towards the left child node of $t$ \textbf{then} 
        \State $\;\quad\quad$ set $c_i=1$ in \eqref{eq:reshuffle} \textbf{else} set $c_i=0$ in \eqref{eq:reshuffle}
            
            \If{$\frac{N_{left_{t}}}{N_{t}} \leq \varepsilon_{2}\;\text{or}\;\frac{N_{left_{t}}}{N_{t}} \geq 1 - { \varepsilon_{2}}$}\Comment{high imbalanced routing}
        \State   $\bullet$  set $d_{max} = argmax \{N_{left_{t}},N_{right_{t}}\}$\Comment{child node of $t$ with max number of $\mathbf{x}_i$ routed to it}
        \State $\bullet$ set $I_t^R$ as the $N_{d_{max}} \varepsilon_{3}$ data points with largest residuals routed towards the child node $d_{max}$ 
        \State $\bullet$ for every $(\textbf{x}_i,y_i) \in I_t^R$ set  $c_i = 1-c_i$ in \eqref{eq:reshuffle} \Comment{aims at routing $\textbf{x}_i$ along the other branch}
        \State
        $\bullet$ determine $\hat{\bm{\w}}_{t}$ by minimizing the WLR function \eqref{eq:reshuffle} restricted to $I_t$
        
            \Else \Comment{moderate imbalanced routing}
            \State $\bullet$ determine $\hat{\bm{\w}}_{t}$ by minimizing the WLR function \eqref{eq:reshuffle} restricted to $I_t$
            \EndIf
            \State $\hat{\bm{\w}}_{W^k_B} \gets \{\hat{\bm{\w}}_{t},\hat{\bm{\w}}_{W^k_B \setminus t} \}$ \Comment{$\hat{\bm{\w}}_{t}$ are the only updated variables}
        \Else
            \State determine $\hat{\bm{\w}}_{W^k_B}$ by solving \eqref{eq:obj} with respect to variables ${\bm{\w}}_{W^k_B}$  
            \Comment{minimize with any 
            algorithm} 
        \EndIf
        
        \State \textbf{return} $\hat{\bm{\w}}_{W^k_B}$
 \EndProcedure
  \end{algorithmic}
  }
\end{algorithm}

\textcolor{black}{The \textsc{UpdateBranchNode} procedure takes as input the dataset $I$, the current working set ${W_B^k}$, the previous update $\bm{\w}^{k-1}_{W^k_B}$ and the imbalance thesholds $\varepsilon_1, \varepsilon_2$ and $\varepsilon_3 \in (0,1)$.}

\textcolor{black}{
Let $t$ denotes the ancestor among all the nodes in ${W_B^k}$, that is, the node with the smallest index, and $I_t$ the restricted data set containing all the data points deterministically falling into 
node $t$.} 

\textcolor{black}{The candidate update $\hat{\bm{\w}}_{W_B^k}$ is obtained by considering the subtree rooted at $t$ (containing all the branch nodes with indices in $W_B^k$) and minimizing the error function \eqref{eq:obj} restricted to the variables $\bm{\w}_{l}$ with $l 
\in W_B^k$ corresponding to the subtree and to the partial dataset $I_t$, while keeping fixed all the others variables $\bm{\w}_{l}$ with $l \in \tau_B \setminus W_B^k$.}

\textcolor{black}{Considering the current inner iteration $k$ of NODEC-DR when the procedure is called, let $N_t$ denote the number of data points in $I_t$, and $N_{left_t}$ and $N_{right_t}$ the numbers of data points deterministically routed towards the left and, respectively, the right child of $t$.} The extent of the data points imbalance at branch node $t$ depends on the difference between $N_{left_t}$ and $N_{right_t}$.

If at branch node $t$, the routing of data points is sufficiently balanced, i.e., there exists $\varepsilon_{1} \in (0,1)$ close to $0.5$ such that $\varepsilon_{1} < \frac{N_{left_t}}{N_t} < 1 - \varepsilon_{1}$, the candidate update $\hat{\bm{\w}}_{W_B^k}$ is obtained by minimizing the error function \eqref{eq:obj} associated to the restricted subtree rooted at $t$ and the restricted dataset $I_t$, over the variables $\bm{\w}_l$ with $l \in W_B^k$. This can be achieved via any nonlinear unconstrained minimization algorithm, such as Quasi-Newton or conjugate gradient methods.  

If at branch node $t$ we have $\frac{N_{left_{t}}}{N_{t}} \leq \varepsilon_{1}$ or\, $\frac{N_{left_{t}}}{N_{t}} \geq 1 - \varepsilon_{1}$, the routing of data points is imbalanced and we consider a second threshold $\varepsilon_{2} < \varepsilon_{1}$ to distinguish between moderate and high imbalance. In both moderate and high imbalance cases, we aim at improving the routing balance of data points at branch node $t$ by optimizing a proxy of the original problem using any unconstrained optimization method. Specifically, the candidate update $\hat{\bm{\w}}_{W_B^k}$ is obtained by minimizing a Weighted Logistic Regression (WLR) function with respect to only the $\bm{\w}_t$ variables of the branch node $t$ and the restricted dataset $I_t$, keeping fixed all the $\bm{\w}_{l}$ with $l \in \tau_B \setminus t$. The WLR objective function is defined as follows:
\begin{equation}
\label{eq:reshuffle}
 -\frac{1}{N_t} \sum_{i \in I_t} w_ic_iln(p_{{\bf x}_i t}) + w_i(1 - c_i)ln(1-p_{{\bf x}_it}),  
\end{equation}

\noindent \textcolor{black}{where if input vector $\mathbf{x}_i$ is routed along the left branch at $t$ then $c_i = 1$ and $w_i=\frac{N_t}{2 N_{left_t}}$, and if ${\bf x}_i$ is routed along the right branch then $c_i = 0$ and $w_i = \frac{N_t}{2 N_{right_t}}$. Note that the weight $w_i$ is larger for $\mathbf{x}_i$ falling into the child node of $t$ towards which the smallest number of data points is routed.}


If the routing of data points at node $t$ is moderately imbalanced, i.e.,  $\frac{N_{left_{t}}}{N_{t}} \leq \varepsilon_{1}$ or\, $\frac{N_{left_{t}}}{N_{t}} \geq 1 - \varepsilon_{1}$ and $\varepsilon_{2} < \frac{N_{left_{t}}}{N_{t}} < 1 - \varepsilon_{2}$, the candidate update $\hat{\bm{\w}}_{W_B^k}$ is obtained by minimizing the WLR function \eqref{eq:reshuffle} associated to the restricted dataset $I_t$, over the variables $\bm{\w}_t$ and keeping fixed all the variables $\bm{\w}_l$ with $l \in \tau_B \setminus t$.

\textcolor{black}{If the routing of data points at node $t$ is highly imbalanced, i.e., $\frac{N_{left_{t}}}{N_{t}} \leq \varepsilon_{2}$ or\, $\frac{N_{left_{t}}}{N_{t}} \geq 1 - \varepsilon_{2}$, we minimize the WLR function \eqref{eq:reshuffle} with modified values of the parameters $c_i$ for some of the data points in $I_t$. The aim is to reroute the input vectors $\textbf{x}_i$ towards the other child node of $t$. Let $d_{max}$ denote the index of the child node of node $t$ where the largest number of data points of $I_t$ deterministically fall. Given $\varepsilon_3 \in (0,1)$, we define $I^R_t$ as the subset of $N_{d_{max}}\varepsilon_3$ data points with the largest residuals among those routed towards the child node $d_{max}$. Since larger residuals may indicate that the corresponding data points are not routed to the most appropriate child node, we try to reroute (reassign) each data point $(\textbf{x}_i,y_i)$ in $I^R_t$ to the other child node of node $t$. To do so, we set $c_i=1-c_i$ for $(\textbf{x}_i,y_i)$ in $I^R_t$ and minimize the WLR function \eqref{eq:reshuffle} using any unconstrained nonlinear optimization method\footnote{Clearly, the resulting solution of the proxy subproblem may not reroute all the data points in $I^R_t$ as indicated by the selected values of the parameter $c_i$.}. \textcolor{black}{Note that the data points to be reassigned to different child nodes are those corresponding to larger terms (residuals)} in the error function \eqref{eq:obj} associated to the restricted subtree rooted at $t$ and the restricted dataset $I_t$. Specifically, the error term corresponding to any given data point $(\mathbf{x}_i,y_i)$ amounts to:}

\begin{equation}
   \label{eq:partial_objective}
    \sum_{t\in {\cal D }_L(t)}P_{it}(\bm{\w})(\hat y_{it}(\bm{\beta}_t)-y_{i})^{2},
\end{equation}

\noindent \textcolor{black}{where ${\cal D }_L(t)$ denotes the subset of leaf nodes that are descendants of node $t$.}

\textcolor{black}{It is important to emphasize that the use of the WLR function \eqref{eq:reshuffle} provides a very effective stabilization of NODEC-DR during the first macro iterations, avoiding to get stuck in poor highly imbalanced solutions. Since minimizing \eqref{eq:reshuffle} is a proxy of the original subproblem and it is very unlikely that Conditions \eqref{eq:descreasing1}-\eqref{eq:decreasing2} are satisfied, such convergence conditions are enabled only after a certain number of iterations (for $k \geq k_0$).
It is easily verified that this automatically occurs in NODEC-DR, as after a finite number of macro iterations the algorithm considers only solutions satisfying Conditions \eqref{eq:descreasing1}-\eqref{eq:decreasing2}. Indeed, assuming that $\varepsilon_1$ is updated at the end of each macro iteration $it$ according to $\varepsilon^{it}_1 = \zeta\varepsilon^{it-1}_1$ with $\zeta \in (0,1)$, the maximum number of data points that can be considered imbalanced  after $it$ macro iterations is equal to $N \varepsilon^{it}_1$. Given an initial value $\varepsilon^0_1$, we simply need to impose $N \varepsilon^0_1\zeta^{it} \leq 1$ so that the condition on line 4 in \textsc{UpdateBranchNode} is no longer satisfied. Since $it \geq \frac{ln(N\varepsilon^0_1)}{-ln(\zeta)}$ and each macro iteration consists of $2^{D-1}$ inner iterations, after at most
$\bar{k} = {\Big \lceil} \frac{ln(N\varepsilon^0_1)}{-ln(\zeta)}{\Big \rceil} 2^{D-1}$ inner iterations the algorithm considers only solutions that satisfy Conditions \eqref{eq:descreasing1}-\eqref{eq:decreasing2}.}

\subsection{Initialization procedure}\label{sec:Init_procedure}
\textcolor{black}{Due to the nonconvexity of the NLO training formulation \eqref{eq:obj}, an appropriate choice of initial solutions can enhance the training phase and lead to SRTs with better testing accuracy.}

\textcolor{black}{The initialization procedure is based on a clustering method and silhouette score.}

\textcolor{black}{Since each internal node actually splits input vectors into two groups, we propose an initialization method based on a sequence of binary clustering problems, aiming at an initial solution $(\bm{\w}^{0},\bm{\beta}^{0})$ where data points are distributed into the leaf nodes so as to maximize an appropriate clustering quality metric. 
Silhouette scores are measures which express how close (similar) each input vectors is with respect to the other data points within its cluster compared to the ones of the neighboring clusters \citep{dudek2020silhouette}. In our implementation we consider as silhouette score the Davies-Bouldin index \citep{davies1979cluster} which expresses the average similarity measure of each cluster with its most similar cluster.} 

\textcolor{black}{
At each step, the procedure starts from the root node and recursively generates a hierarchical assignment of the input vectors of the training set $I$ to the branch nodes and leaf nodes. For any node $t$, let $C_t$ denote the subset of the input vectors $\mathbf{x}_i$ assigned to $t$, and $C_{\mbox{\footnotesize l-child(t)}}$ and $C_{\mbox{\footnotesize r-child(t)}}$ the subsets of $C_t$ assigned to the left and, respectively, right child of $t$. At the root node ($t=1$), we apply the K-means algorithm \citep{lloyd1982least} to $C_1$, which consists of all the input vectors in $I$, and obtain  $C_{\mbox{\footnotesize l-child(1)}}$ and $C_{\mbox{\footnotesize r-child(1)}}$, where l-child(1)=2 and r-child(1)=3. This binary partition is carried out at each branch node, proceeding recursively from the root to the branch nodes at depth $D-1$. This leads to a partition $A$ of all input vectors in $C_1$ into $2^D$ subsets $C_t$ associated to all the leaf nodes ($t \in \tau_L$).}


\renewcommand{\thealgorithm}{Initialization procedure} 
\begin{algorithm}[H]
\floatname{algorithm}{} 
\caption{}
\label{alg:init}
$\textbf{Input:}\, \text{dataset}\, I;\, \text{depth} \,D; \,\text{number}\,\text{of}\,\text{repetitions}\, r\\\textbf{Output:}\,\text{initial}\,\text{solution}\,(\bm{\w}^0,\bm{\beta}^0)$
  
  {\fontsize{10}{12}\selectfont
  \begin{algorithmic}[1]
    \Procedure{InitializationProcedure}{$I,D,r$}
    
    \State $\widehat{DB} \gets +\infty$ \Comment{initialization for the best Davies-Bouldin index}
    	\For{$ j = 1,...,r $}\Comment{generate r partitions of $I$ into $2^D$ (number of leaf nodes) subsets}
    	\State $A^j \gets [\emptyset]$ \Comment{at the end, $A^j$ is a list of subsets of data points assigned to each leaf node}
    	\State $C^j \gets [I]$ \Comment{at the end, $C^j$ is a list of subsets of data points for each node}
     
    		\For{$t = 1,...,2^{D}-1$}\Comment{loop over all branch nodes starting from the root node ($t=1$)}
            \State \textcolor{black}{$\textrm{apply K-means method to } C^j_t  \textrm{ to generate clusters } C^j_{\mbox{\footnotesize l-child(t)}} \textrm{ and } C^j_{\mbox{\footnotesize r-child(t)}}$}
      \State $\textrm{append } (C^j_{\mbox{\footnotesize l-child(t)}},C^j_{\mbox{\footnotesize r-child(t)}}) \textrm{ to } C^j$
    
    		\If{$\textrm{children of node } t \textrm{ are leaf nodes}$}
    		\State $\textrm{append } (C^j_{\mbox{\footnotesize l-child(t)}}, C^j_{\mbox{\footnotesize r-child(t)}}) \textrm{ to } A^j$
    		\EndIf
    		\EndFor
    		
    	\State $DB \gets \mbox{Davies-Bouldin}(A^j)$ \Comment{compute DB index for the leaf node clusters}
    	\If{$DB < \widehat{DB}$}
    	      \State $\widehat{DB} \gets DB$
           \State ${\hat{C}} \gets C^j$
    	      \State ${\hat{A}} \gets A^j$ \Comment{${\hat{A}}$ is the best assignment of data points found so far}

    	\EndIf
    	
    	\EndFor
    \For{$t = 1,...,2^{D}-1$}\Comment{loop over the branch nodes to derive the $\bm{\w}^0$ parameters}
    		 \State determine $\bm{\w}^0_{t}$ by applying Logistic Regression to $\hat{C}[\mbox{\footnotesize l-child(t)}]$ and $\hat{C}[\mbox{\footnotesize r-child(t)}]$ 
    		\EndFor
     \For{$t = 2^{D},...,2^{D+1}-1$}\Comment{loop over the leaf nodes to derive the $\bm{\beta}^0$ parameters}
    		 \State determine $\bm{\beta}^0_{t}$ by applying Linear Regression to $\hat{C}[t]$ 
    		\EndFor
    \State \textbf{return} $(\bm{\w}^0,\bm{\beta}^0)$ \Comment{returns the best initial solution according to DB index}
    \EndProcedure
  \end{algorithmic}
  }
\end{algorithm}

\textcolor{black}{By repeating $r$ times the above step, multiple candidate partitions $A^j$ of input vectors into leaf nodes, with $j = 1,\dots,r$, are obtained. Among all partitions $A^j$ the one with the smallest Davies-Bouldin silhouette score, say $\hat{A}$, together with its corresponding hierarchical assignment of input vectors $\hat{C}$, is selected. Then, for each branch node $t$ the variables $\bm{\w}^0_t$ are determined by applying a binary Logistic Regression in which the input vectors associated to the two classes are those in $\hat{C}[\mbox{\small l-child(t)}]$ and $\hat{C}[\mbox{\small r-child(t)}]$. Analogously, for each leaf node $t$ the related $\bm{\beta}^0_t$ variables are determined by applying a Linear Regression to the data points associated to the input vectors in $\hat{C}[t]$.} 

\textcolor{black}{Notice that alternative clustering methods and silhouette scores can be considered.}

\section{Experimental results}\label{sec:experiments_regression}

In this section we present and discuss comparative results obtained by training our SRT model with the NODEC-DR algorithm, and two state-of-the-art regression tree approaches. From now on the methodology consisting of training a SRT model with the NODEC-DR algorithm is simply referred to as SRT.

The section is organized as follows. Section \ref{sec:dataset_exp} describes the experimental settings and the datasets. In Section \ref{sec:results_exp} we report and discuss the results in terms of testing $R^2$ and computational time obtained by: two SRT variants, \textcolor{black}{the discrete optimization method ORT-L} proposed in \citep{dunn2018optimal,bertsimas2019machine} for training deterministic regression trees, and the nonlinear optimization formulation \textcolor{black}{ORRT} in \citep{blanquero2022sparse} for training soft \textcolor{black}{multivariate regression trees. In \ref{appendix:HME_forest}, we also compare our SRT variants with the tree-based ensemble model Random Forest (RF).} 

\subsection{Datasets and experimental setting}
\label{sec:dataset_exp} 
\textcolor{black}{In the numerical experiments} we consider 15 well-known datasets from the UCI Machine Learning Repository \citep{asuncion2007uci} and the KEEL repository \citep{alcala2011keel}, which have been extensively used in the literature (e.g. in \cite{dunn2018optimal}, \cite{bertsimas2019machine} and in \cite{blanquero2022sparse}). Table \ref{tab:dataset_reg} reports the name and the size \textcolor{black}{of the 15 datasets, namely, the number of features $p$ and the number of data points $N$.}

To assess the testing accuracy we use the determination coefficient $R^2$, namely, the testing $R^2=1 - \frac{SS_{res}^{test}}{SS_{tot}^{test}}$
where $SS_{res}^{test}$ is the sum of squares of residuals and $SS_{tot}^{test}$ is the total sum of squares.
For all datasets, all input features have been normalized in the $\left[0,1\right]$ interval while the output response has been standardized.

The testing $R^2$ is evaluated by means of $k$-fold cross-validation, with $k= 4$. 
For each fold the model is trained using $20$ different random seeds to generate different initial solutions with the initialization procedure, so that
the final testing $R^2$ and computational time are averaged on the $4$ folds and $20$ initial solutions ($80$ runs).

The experiments are carried out on a PC with Intel(R) Core(TM) i7-9700K CPU @ 3.60GHz with 16 GB of RAM. SRT has been implemented in Python 3.6.10. In particular, the tree structure has been built in Cython, a C extension for Python. 
As to NODEC-DR, the \texttt{scipy.sparse} 1.5.2 Python package is adopted to solve the linear least square problem of the {\bf LN Step}. Concerning the {\bf BN Step}, the SLSQP solver in \texttt{scipy.optimize} 1.5.2 in tandem with \texttt{numba} 0.51.2 has been used for the general case, while in the imbalanced case the Weighted Logistic Regression is performed using the \texttt{LogisticRegression} function of the \texttt{sklearn} 0.23.1 package with the default settings. The \texttt{Kmeans} function of \texttt{sklearn} is used in the initialization procedure.

\begin{table}[h]
	\centering
	\renewcommand{\arraystretch}{0.7}
	\resizebox{0.65\columnwidth}{!}{
		\begin{tabular}{lllllll}
			\hline
			\multicolumn{1}{c}{Dataset} &  & \multicolumn{1}{c}{Abbreviation} &  & \multicolumn{1}{c}{$N$} &  & \multicolumn{1}{c}{$p$} \\ \hline
			Abalone                     &  & Abalone                          &  & 4177                  &  & 8                     \\
			Ailerons                    &  & Ailerons                         &  & 7154                  &  & 40                    \\
			Airfoil                     &  & Airfoil                          &  & 1503                  &  & 5                     \\
			Auto-mpg                    &  & Auto                          &  & 392                   &  & 7                     \\
			Compactiv                     &  & Compact                          &  & 8192                  &  & 21                    \\
			Computer hardware           &  & Computer                         &  & 209                   &  & 37                    \\
			Cpu small                   &  & Cpu small                        &  & 8192                  &  & 12                    \\
			Delta ailerons              &  & Delta                    &  & 7129                  &  & 5                     \\
			Elevators                   &  & Elevators                        &  & 16599                 &  & 18                    \\
			Friedman artificial         &  & Friedman                         &  & 40768                 &  & 10                    \\
			Housing                     &  & Housing                          &  & 506                   &  & 13                    \\
			Kin8nm                      &  & Kin8nm                           &  & 8192                  &  & 8                     \\
			Lpga2009                        &  & Lpga2009                             &  & 146                  &  & 11                    \\
			Puma                        &  & Puma                             &  & 8192                  &  & 32                    \\
			Yacht hydrodynamics         &  & Yacht                            &  & 308                   &  & 6                     \\ \hline
		\end{tabular}
	}
	\caption{\textcolor{black}{The name and size of the 15 datasets}.} \label{tab:dataset_reg}
\end{table}

\textcolor{black}{In the CDF \eqref{eq:cdf} of the SRT model, we set $\mu=1$. Since we aim at good accuracy in relatively short training times, in the experiments we adopt an early stopped version of NODEC-DR with ${ M\_it}=10$ and $ k_0> M\_it$ (i.e. the asymptotic convergence conditions are disabled). In the reassignment heuristic,
	we set:  $\varepsilon_1=0.1 $, $ \varepsilon_2=0.3 $, $ \varepsilon_3=0.4 $ and $\zeta=0.8$.}


\subsection{Comparison with soft and deterministic regression trees approaches}
\label{sec:results_exp}

We consider two versions of SRT: with and without $\ell_2$ regularization term included into the error function \eqref{eq:obj}, which are referred to as SRT $\ell_2$ and SRT, respectively. For SRT $\ell_2$, the regularization hyperparameters $\lambda_{\bm{\w}}$ and $\lambda_{\bm{\beta}}$ are set as $\lambda_{\bm{w}}=\frac{2}{p \times |\tau_{B}|}$ and $\lambda_{\bm{\beta}}=\frac{2}{p \times |\tau_{L}|}$.  \textcolor{black}{As previously mentioned, these two versions of SRT are compared with ORT-L for deterministic regression trees and ORRT for traditional soft regression trees.}

\textcolor{black}{
	In ORT-L, a MILO formulation for training deterministic univariate regression trees involving linear predictions with Lasso sparsity terms at each leaf node, is tackled via a local search heuristic.} 
\textcolor{black}{
	The ORT-L version used in this work is the \texttt{OptimalTreeRegressor} function of the Interpretable AI package version 2.0.2 available at \url{https://docs.interpretable.ai/stable/}.
	The results reported below are obtained by disabling the gridsearch on both the complexity parameter ($cp=0$) and the $l_1$ regularization hyperparameter ($\lambda= 1e^{-3}$) in order to reduce computational time as much as possible. In previous computational experiments on all the 15 datasets reported in \ref{sec:ortl_cp}, we applied a gridsearch on both the complexity parameter $cp$ (internally handled by the software) and the regularization (choosing among three different values for $\lambda$). The results with the gridsearch are definitely comparable in terms of testing $R^2$ to the ones obtained by fixing the hyperparameters but the computational times are significantly higher.}
In \citep{dunn2018optimal,bertsimas2019machine} the authors also proposed a more general and computationally demanding version of ORT-L involving hyperplane splits, 
\textcolor{black}{referred to as ORT-LH. However, they point out that ORT-L and ORT-LH provide comparable $R^2$. In \ref{appendix:ortl_vs_ortlh}, we report experimental results on three datasets confirming that } 
the substantial increase in computational time of ORT-LH is not compensated by a significant 
\textcolor{black}{improvement in terms of $R^2$. }

\textcolor{black}{
	In ORRT, multivariate randomized regression trees are trained by solving a NLO formulation with an appropriate off-the-shelf solver. As in \cite{blanquero2022sparse}, we use the SLSQP method in \texttt{scipy.optimize} 1.5.2 package.} We adopt \texttt{numba} 0.51.2 to reduce training times.


\textcolor{black}{We carried out numerical experiments with regression trees of depth $D=2$ and $D=3$. For the sake of space, the results obtained for regression trees of depth $D=3$ are reported in Table \ref{tab:depth3}, while those for regression trees of depth $D=2$ are available in \ref{appendix:results-trees-depth-2}.}

For each method, the tables indicate the average testing $R^2$, its standard deviation $\sigma$ (divided by a $1e^{-2}$ factor for visualization reasons) and the average training time in seconds. \textcolor{black}{For ORRT, 
the numbers appearing in the $R^2<0$ column correspond to the number of times (out of the 80 runs)} that the training phase provides a suboptimal solution with a negative testing $R^2$ (i.e. worse than using the average as prediction). \textcolor{black}{The arithmetic and the geometric averages over all the datasets are reported at the bottom of the tables.}

As far as the two SRT versions with and without $\ell_2$ are concerned, 
\textcolor{black}{Table \ref{tab:depth3} and the results reported in \ref{appendix:results-trees-depth-2} for SRTs of depth $D=2$ show} that the $\ell_2$-regularized version leads to a slight improvement \textcolor{black}{in terms of average testing $R^2$}
(approximately $1\%$ for both depths) and a smaller average standard deviation $\sigma$, while requiring equivalent computational time. This suggests that the $\ell_2$ regularization term enhances the model robustness. \textcolor{black}{For datasets such as Housing and Lpga2009, SRT $\ell_2$ yields an increase in testing $R^2$ of at least $7.5\%$. For datasets such as Airfoil, Computer and Yatch, SRT leads to slightly higher testing $R^2$ than SRT $\ell_2$.}

\textcolor{black}{Concerning the comparison between SRT $\ell_2$ and ORRT for soft regression trees with $D=3$, SRT $\ell_2$ outperforms ORRT in terms of testing accuracy, improving the average testing $R^2$ by $29\%$. 
	Furthermore, ORRT turns out to be 
	sensitive to the choice of the initial solutions and less robust than SRT $\ell_2$, as shown by both the higher average standard deviation of the testing $R^2$ (more than three times higher) and by the number of runs leading to negative testing $R^2$. 
	Across all the datasets, ORRT yields a negative testing $R^2$ at least 3 times per dataset, and on average more than $14$ times. For the Ailerons dataset, 44 out of the 80 runs lead to a negative $R^2$. 
	As to the training times, SRT $\ell_2$ is on average substantially faster than ORRT, with a seven times smaller arithmetic average of the computational times. Note that ORRT requires a slightly smaller computational time for the three datasets, Airfoil, Kin8nm and Yacht, but the average testing $R^2$ is substantially lower.}

\textcolor{black}{For all the datasets except Lpga2009, SRT $\ell_2$ leads to a higher average testing $R^2$ compared to ORRT, and for 13 out of the 15 datasets the percentage improvement is larger than $6\%$. For example, for the Kin8nm and Yacht datasets the increase in average testing $R^2$ between SRT $\ell_2$ and ORRT is by more than $50\%$, while the computational times are similar. For the Puma dataset, the average testing $R^2$ is 4 times larger and the computational time substantially shorter. For the Compact dataset, SRT $\ell_2$ leads to an average testing $R^2$ which is $36\%$ higher within a twelve times shorter computational time. Note that for the Lpga2009 dataset, ORRT yields a slightly better testing accuracy (by $0.3\%$) compared to SRT $\ell_2$ within a slightly shorter computational time.}

\begin{table}[H]
	\centering
	\resizebox{\textwidth}{!}{\begin{tabular}{lccccccccccc}
			\cline{4-12}
			& \multicolumn{1}{l}{} & \multicolumn{1}{l}{}    & \multicolumn{9}{c}{D=3}                                                                                                                                                                                       \\ \cline{4-12} 
			& \multicolumn{1}{l}{} & \multicolumn{1}{l}{}    & \multicolumn{2}{c}{\textbf{SRT}}                             & \multicolumn{2}{c}{\textbf{SRT} \bm{$\ell_2$}}                    & \multicolumn{2}{c}{\textbf{ORT-L}}                             & \multicolumn{3}{c}{\textbf{ORRT}}                  \\ \hline
			Dataset                        & N                    & p                       & $R^2\, (\sigma \,1e^{-2})$ & Time                       & $R^2\, (\sigma\, 1e^{-2})$ & Time                       & $R^2 \,(\sigma\, 1e^{-2})$ & Time                         & $R^2\, (\sigma\, 1e^{-2})$ & Time   & $R^2<0$ \\ \hline
			\multicolumn{1}{l|}{Abalone}   & 4177                 & \multicolumn{1}{c|}{8}  & 0.558 (2.67)           & \multicolumn{1}{c|}{24.5}  & 0.564 (2.16)           & \multicolumn{1}{c|}{24}    & 0.545 (1.9)            & \multicolumn{1}{c|}{147.9}   & 0.516 (6.67)           & 107.6  & 20      \\
			\multicolumn{1}{l|}{Ailerons}  & 7154                 & \multicolumn{1}{c|}{40} & 0.826 (0.9)            & \multicolumn{1}{c|}{50.6}  & 0.835 (0.43)           & \multicolumn{1}{c|}{50.5}  & 0.825 (1.2)            & \multicolumn{1}{c|}{310.4}   & 0.784 (11.65)          & 2491.8 & 44      \\
			\multicolumn{1}{l|}{Airfoil}   & 1503                 & \multicolumn{1}{c|}{5}  & 0.816 (3.9)            & \multicolumn{1}{c|}{11.2}  & 0.807 (2.14)           & \multicolumn{1}{c|}{11.3}  & 0.842 (1.3)            & \multicolumn{1}{c|}{4.7}     & 0.513 (1.8)            & 9.04   & 20      \\
			\multicolumn{1}{l|}{Auto mpg}  & 392                  & \multicolumn{1}{c|}{7}  & 0.842 (4.12)           & \multicolumn{1}{c|}{6.1}   & 0.873 (2.17)           & \multicolumn{1}{c|}{6.1}   & 0.819 (6.47)           & \multicolumn{1}{c|}{10}      & 0.815 (2.45)           & 6.8    & 7       \\
			\multicolumn{1}{l|}{Compact}   & 8192                 & \multicolumn{1}{c|}{21} & 0.979 (0.38)           & \multicolumn{1}{c|}{72.1}  & 0.98 (0.27)            & \multicolumn{1}{c|}{72.7}  & 0.98 (0.48)            & \multicolumn{1}{c|}{1103.5}  & 0.721 (3.4)            & 905.6  & 5       \\
			\multicolumn{1}{l|}{Computer}  & 209                  & \multicolumn{1}{c|}{37} & 0.973 (1.8)            & \multicolumn{1}{c|}{4}     & 0.955 (3.37)           & \multicolumn{1}{c|}{4}     & 0.889 (16)             & \multicolumn{1}{c|}{3.2}     & 0.887 (13.8)           & 4.3    & 14      \\
			\multicolumn{1}{l|}{Cpu small} & 8192                 & \multicolumn{1}{c|}{12} & 0.969 (0.26)           & \multicolumn{1}{c|}{57.1}  & 0.97 (0.25)            & \multicolumn{1}{c|}{58.7}  & 0.971 (0.31)           & \multicolumn{1}{c|}{427.4}   & 0.705 (7.24)           & 323.2  & 3       \\
			\multicolumn{1}{l|}{Delta}     & 7129                 & \multicolumn{1}{c|}{5}  & 0.704 (1.11)           & \multicolumn{1}{c|}{28.4}  & 0.706 (1)              & \multicolumn{1}{c|}{29.5}  & 0.709 (0.77)           & \multicolumn{1}{c|}{17.3}    & 0.672 (2.7)            & 52.7   & 4       \\
			\multicolumn{1}{l|}{Elevators} & 16599                & \multicolumn{1}{c|}{18} & 0.886 (0.52)           & \multicolumn{1}{c|}{97.6}  & 0.884 (0.57)           & \multicolumn{1}{c|}{94.7}  & 0.812 (0.58)           & \multicolumn{1}{c|}{515.7}   & 0.783 (10.3)           & 1476.7 & 21      \\
			\multicolumn{1}{l|}{Friedman}  & 40768                & \multicolumn{1}{c|}{10} & 0.937 (0.53)           & \multicolumn{1}{c|}{310.4} & 0.938 (0.28)           & \multicolumn{1}{c|}{311.8} & 0.935 (0.06)           & \multicolumn{1}{c|}{1356.5}  & 0.721 (1.23)           & 455.7  & 13      \\
			\multicolumn{1}{l|}{Housing}   & 506                  & \multicolumn{1}{c|}{13} & 0.809 (8.3)            & \multicolumn{1}{c|}{7.2}   & 0.872 (4.48)           & \multicolumn{1}{c|}{7.35}  & 0.785 (8.3)            & \multicolumn{1}{c|}{78.4}    & 0.715 (6.1)            & 19.1   & 10      \\
			\multicolumn{1}{l|}{Kin8nm}    & 8192                 & \multicolumn{1}{c|}{8}  & 0.787 (1.9)            & \multicolumn{1}{c|}{50.1}  & 0.786 (1.9)            & \multicolumn{1}{c|}{50.7}  & 0.645 (1.1)            & \multicolumn{1}{c|}{946.4}   & 0.412 (1)              & 45.9   & 9       \\
			\multicolumn{1}{l|}{Lpga2009}  & 146                  & \multicolumn{1}{c|}{11} & 0.8 (11.3)             & \multicolumn{1}{c|}{4.9}   & 0.877 (2.5)            & \multicolumn{1}{c|}{4.9}   & 0.844 (4)              & \multicolumn{1}{c|}{16.5}    & 0.88 (4.2)             & 4.58   & 8       \\
			\multicolumn{1}{l|}{Puma}      & 8192                 & \multicolumn{1}{c|}{32} & 0.884 (0.91)           & \multicolumn{1}{c|}{135.9} & 0.883 (0.88)           & \multicolumn{1}{c|}{136.5} & 0.907 (0.39)           & \multicolumn{1}{c|}{28481.8} & 0.215 (3.2)            & 233.1  & 13      \\
			\multicolumn{1}{l|}{Yacht}     & 308                  & \multicolumn{1}{c|}{6}  & 0.99 (0.49)            & \multicolumn{1}{c|}{5.1}   & 0.983 (0.48)           & \multicolumn{1}{c|}{5.4}   & 0.992 (0.11)           & \multicolumn{1}{c|}{2.8}     & 0.645 (5.6)            & 3.6    & 21      \\ \hline
			\multicolumn{3}{l|}{Arithmetic avg}                                             & \textbf{0.851} (2.6)            & \multicolumn{1}{c|}{\textbf{57.7}}  & \textbf{0.861} (1.52)           & \multicolumn{1}{c|}{\textbf{57.9}}  & \textbf{0.833} (2.87)           & \multicolumn{1}{c|}{\textbf{2228.2}}  & \textbf{0.666} (5.42)           & \textbf{409.3}  & 14.13   \\ \hline
			\multicolumn{3}{l|}{Geometric avg}                                              & \textbf{0.842} (1.4)            & \multicolumn{1}{c|}{\textbf{25.9}}  & \textbf{0.853} (1.02)           & \multicolumn{1}{c|}{\textbf{26}}    & \textbf{0.823} (1.04)           & \multicolumn{1}{c|}{\textbf{109.5}}   & \textbf{0.632} (4.14)           & \textbf{66.9}   & 11.2    \\ \hline
		\end{tabular}
	}
	\caption{Comparison between SRT, SRT $\ell_2$, ORT-L and ORRT for regression trees of depth $D=3$. \textcolor{black}{The rightmost column corresponds to the number of times (out of the 80 runs) that ORRT provides a suboptimal solution with a negative testing $R^2$. For SRT, SRT $\ell_2$ and ORT-L, such a number is equal to $0$.}}\label{tab:depth3}
\end{table}

\textcolor{black}{Concerning the comparison between SRT $\ell_2$ and ORT-L, the last two rows of Table \ref{tab:depth3} indicate that SRT $\ell_2$ achieves a $3\%$ higher average testing $R^2$ with a lower average standard deviation, and hence turns out to be less sensitive with respect to the initial solutions. Moreover, the average computational time required by SRT $\ell_2$ is two orders of magnitude lower than that required by ORT-L. 
}

\textcolor{black}{According to Table \ref{tab:depth3} we can distinguish three cases. 
For a first group of datasets, consisting of Auto-mpg, Computer, Elevators, Housing, and Kin8nm, SRT $\ell_2$ substantially outperforms ORT-L in terms of average testing $R^2$ (by at least $6\%$) and in terms of average computational time (by at least $50\%$), except for the Computer dataset which can be dealt with in just a few seconds by both approaches. 
For a second group of datasets, consisting of Abalone, Ailerons, Compact, Cpu small, Friedman and Lpga2009, SRT $\ell_2$ leads to slightly higher or equal average testing $R^2$ (up to $3.9\%$ higher) and to considerable average speedup factors, ranging from 3 to 10. 
For a third group of datasets, consisting of Airfoil, Delta, Puma, and Yacht, ORT-L yields a slightly higher average testing $R^2$ than SRT $\ell_2$, with $0.3\%$ to $3.5\%$ improvements. For the Airfoil, Delta and Yatch datasets, ORT-L turns out to be faster by a factor between 1.5 and 2.5, while for the dataset Puma SRT $\ell_2$ is two orders of magnitude faster.
}


Overall, as shown in the last two rows of \textcolor{black}{Table \ref{tab:depth3} and of the table in 
\ref{appendix:results-trees-depth-2}
containing the results for regression trees of depth $D=2$, SRT $\ell_2$ leads to more accurate and robust regression trees than ORRT in shorter training times and on average to slightly higher testing $R^2$ in substantially shorter training times compared to ORT-L.}

\begin{table}[!htb]
	\centering
	\resizebox{\columnwidth}{!}{
		\begin{tabular}{lcccc|cccc}
			\hline
			& \multicolumn{4}{c|}{D=2}                                                           & \multicolumn{4}{c}{D=3}                                                           \\ \cline{2-9} 
			Dataset   & \multicolumn{2}{c}{$R^2$ improvement $\%$} & \multicolumn{2}{c|}{time saving $\%$} & \multicolumn{2}{c}{$R^2$ improvement $\%$} & \multicolumn{2}{c}{time saving $\%$} \\ \cline{2-9} 
			& \textbf{SRT $\ell_2$}             & \textbf{ORT-L}             & \textbf{SRT $\ell_2$}           & \textbf{ORT-L}          & \textbf{SRT $\ell_2$}             & \textbf{ORT-L}             & \textbf{SRT $\ell_2$}          & \textbf{ORT-L}          \\ \hline
			Abalone   & -                      & 0.7              & 87.6                & -              & 3.5                   & -                 & 83.8                & -              \\
			Ailerons  & 1.4                   & -                 & 92.8                & -              & 1.2                   & -                 & 83.7               & -              \\
			Airfoil   & 0.5                  & -                 & -                    & 59.9          & -                      & 4.3              & -                   & 58.7          \\
			Auto-mpg  & 3.8                   & -                 & 58.4                & -              & 6.59                   & -                 & 38.6               & -              \\
			Compact   &     -                   & 0.1               & 94.4                & -              & -                      & -                 & 93.4               & -              \\
			Computer  & 9.9                   & -                 & -                    & 4.5           & 7.4                   & -                 & -                   & 20.6           \\
			Cpu small & 0                      & 0                 & 85.4                & -              & -                      & 0.1               & 86.3               & -              \\
			Delta     & 0.6                   & -                 & -                    & 27.1          & -                      & 0.4              & -                   & 41.1          \\
			Elevators & 7.3                   & -                 & 91.8                 & -              & 8.9                   & -                 & 81.6              & -              \\
			Friedman  & 0.9                    & -                 & 54.3                 & -              & 0.3                   & -                 & 77.0               & -              \\
			Housing   & 8.2                   & -                 & 93.3                & -              & 11.1                  & -                 & 90.6               & -              \\
			Kin8nm    & 27.7                  & -                 & 95.7                & -              & 21.9                  & -                 & 94.6               & -              \\
			Lpga2009  & 3.6                    & -                 & 81.6                & -              & 3.9                   & -                 & 70.0               & -              \\
			Puma      & -                      & 1.0              & 99.6                & -              & -                      & 2.7              & 99.5               & -              \\
			Yacht     & -                      & 1.5               & -                    & 61          & -                      & 0.9              & -                   & 48.1          \\ \hline
		\end{tabular}
	}
	\caption{Comparison between SRT $\ell_2$ and ORT-L for regression trees with depths $D = 2,3$ in terms of the percentage improvement in testing $R^2$ and computational time, with respect to the other methods.}\label{tab:perc_gap}
\end{table}

\textcolor{black}{Table \ref{tab:perc_gap} reports the results obtained for regression trees of depths $D=2$ and $D=3$ in terms of percentage improvement in testing accuracy and computational time between SRT $\ell_2$ and ORT-L. Overall SRT $\ell_2$ outperforms ORT-L. For regression trees of depth $D=2$, most of the percentage $R^2$ improvements are in favor of SRT $\ell_2$ with a maximum of $27.7\%$. Note that in the four cases where ORT-L leads to higher average testing $R^2$ the percentage improvement does not exceed $1.5\%$. For 11 datasets out of 15, SRT $\ell_2$ saves between $54.3\%$ and $99.6\%$ of the computational time (for $9$ datasets more than $81\%$). As to ORT-L, the only four datasets with positive computational time savings (from $4.5\%$ to $61\%$) require less than 30 seconds. Similarly, for regression trees of depth $D=3$, SRT $\ell_2$ leads to a higher testing $R^2$ for 10 out of 15 datasets with a maximum improvement of approximately $21.9\%$ and substantial computational time savings for 11 datasets (up to $99.5\%$). Notice that ORT-L yields better testing $R^2$ with improvements of at most $4.3\%$.}

\textcolor{black}{To conclude, it is worth to point out that, unlike in \citep{blanquero2022sparse} where the reported results correspond to the best solutions obtained over a prescribed number of random initial solutions, in this work we compute the average performance measures of all the methods over all the initial solutions and the k-folds. This provides a more comprehensive and reliable assessment of the expected performance of the model.} \textcolor{black}{In \ref{appendix:orrt_multi}, we illustrate, on four datasets, the differences between reporting ORRT average solutions or ORRT best solutions.
Notice that SRT average solutions considerably outperforms also the ORRT best solutions.}

\section{Concluding remarks}\label{sec:conclusions} 
\textcolor{black}{We proposed a new soft regression tree variant where, for every input vector, the tree prediction is the output of a single leaf node, which is obtained by following from the root the branches of higher probability. 
Such SRTs satisfy the conditional computation property and benefit from the related advantages.
}

We investigated the universal \textcolor{black}{approximation power} of SRTs, by showing that the \textcolor{black}{class of functions expressed by such models} is dense in the class of all continuous functions over arbitrary compact domains. \textcolor{black}{We devised a general convergent node-based decomposition training scheme, and we implemented a specific version, NODEC-DR, which includes a clustering-based initialization procedure and a heuristic for reassigning the data points along the tree.}

The numerical experiments on 15 well-known datasets show that our SRT model variant trained with NODEC-DR outperforms ORRT  \citep{blanquero2022sparse} both in terms of testing accuracy and training times,
and achieves a remarkable speed up \textcolor{black}{and a slightly better average testing accuracy} with respect to the MILO-based local search for deterministic regression trees in \citep{dunn2018optimal,bertsimas2019machine}. 
\textcolor{black}{Results reported in \ref{appendix:HME_forest} indicate that SRT turns out to be slightly more accurate than Random Forest, which is less interpretable.}


\textcolor{black}{As to future work, we could investigate alternative working set selection strategies involving smaller subsets of variables at each iteration in order to tackle larger datasets. It would also be interesting to extend the SRT model to handle more complex data types such as functional data, network data, and spatial data.}
\section*{Acknowledgment}

The authors would like to thank Dr. Jack Dunn for kindly providing us with the license for the Interpretable AI package (\url{https://docs.interpretable.ai/stable/}).

\bibliographystyle{elsarticle-harv} 
\bibliography{mybibliography.bib}

\appendix

\section{Proof of the universal approximation property}
\label{sec:appendixA}

We prove Theorem \ref{th:dense} by showing that $\mathbb{H}$ is dense in $\mathbb{C}(\mathbb{X})$. Let us recall that
$$\mathbb{H} = \{ \sum_{t = 1}^{2^D} \pi_{\mathbf{x}t}(\bm{\w}_{A(t)}) \beta_{0t}\ |\ D \in \mathbb{N}, \bm{\chi}' 
\in \mathbb{R}^{(p+1)(2^{D}-1)+2^{D}}\}$$
\textcolor{black}{where $\bm{\chi}'$ is the vector containing all the model parameters in $\bm{\w}$ and $\bm{\beta}$, and $\pi_{\mathbf{x}t}(\bm{\w}_{A(t)})$ is defined in expression \eqref{eq:pi_approx}, namely: 
$$ \qquad \qquad \quad \prod_{\ell\in A_{L(t)}}
	\mathds{1}_{0.5}\left(\frac{1}{1+exp\left ( -\mu (\bm{\w}_{\ell}^T\mathbf{x})  \right )}\right) \;\prod_{r\in A_{R(t)}}\mathds{1}_{0.5}\left(1 - \frac{1}{1+exp\left ( -\mu (\bm{\w}_{r}^T\mathbf{x})  \right )}\right). \qquad    \, (6)
$$
}
\hspace{-4pt}To do so we exploit the Stone-Weierstrass theorem as stated in \citep{cotter1990stone-weierstrass}.

\begin{theorem}
\label{th:stone_weierstrass}
Let $\mathbb{X} \subset \mathbb{R}^p$ be a compact set, and let $\mathbb{F}$ be a set of continuous real-valued functions on $\mathbb{X}$ satisfying the following conditions:
\begin{enumerate}
    \item[i)] The constant function $f(\mathbf{x})=1$ is in $\mathbb{F}$.
    \item[ii)] For any two points $\mathbf{x}_1,\mathbf{x}_2 \in \mathbb{X}$ such that $\mathbf{x}_1 \neq \mathbf{x}_2,$ there exists a function $f \in \mathbb{F}$ such that $f(\mathbf{x}_1) \neq f(\mathbf{x}_2)$.
    \item[iii)] If $a \in \mathbb{R}$ and $f \in \mathbb{F}$, then $a\, f(\mathbf{x}) \in \mathbb{F}$.
    \item[iv)] If $f, g \in \mathbb{F}$, then $f\, g \in \mathbb{F}$.
    \item[v)] If $f, g \in \mathbb{F}$, then $f+g \in \mathbb{F}$.
\end{enumerate}
Then $\mathbb{F}$ is dense in $\mathbb{C}(\mathbb{X})$. 

\end{theorem}

\noindent \textbf{Proof}.

\noindent \textcolor{black}{For notational simplicity, from now on, in the logistic CDF we assume the parameter $\mu=1$.}

\noindent Conditions $i)$ to $iii)$ are straightforward.

\begin{claim}
	The constant function $f(\mathbf{x})=1$ is in $\mathbb{H}$.
\end{claim}
	Consider a SRT of depth $D=1$ which implements the function $f(\mathbf{x}) 
	=\mathds{1}_{0.5}(\frac{1}{1+exp\left ( -(\bm{\w}_{1}^T\mathbf{x})  \right )})\beta_{01} + \mathds{1}_{0.5}(1 - \frac{1}{1+exp\left ( -(\bm{\w}_{1}^T\mathbf{x})  \right )})\beta_{02}$ with $\beta_{01} = \beta_{02} = 1$. Then for all $\bm{\w}_1 \in \mathbb{R}^{p+1}$ we have $f(\mathbf{x})=1$.

\begin{claim}
    For any two points $\mathbf{x}_1,\mathbf{x}_2 \in \mathbb{X}$ such that $\mathbf{x}_1 \neq \mathbf{x}_2,$ there exists a function $f \in \mathbb{H}$ such that $f(\mathbf{x}_1) \neq f(\mathbf{x}_2)$.
\end{claim}
	Consider a SRT of depth $D=1$ which implements a function $f(\mathbf{x}) 
	=\mathds{1}_{0.5}(\frac{1}{1+exp\left ( -(\bm{\w}_{1}^T\mathbf{x})  \right )})\beta_{01} + \mathds{1}_{0.5}(1 - \frac{1}{1+exp\left ( -(\bm{\w}_{1}^T\mathbf{x})  \right )})\beta_{02}$.
	To guarantee that $f(\mathbf{x}_1) \neq f(\mathbf{x}_2)$ it is sufficient to seek values for the parameter vector $\bm{\w}_1$ such that $\mathbf{x}_1$ and $\mathbf{x}_2$ are separated by the hyperplane defined by 
	$$\bm{\w}_1^T \mathbf{x} = 0.$$ 
	Such values exist since any two distinct points can be separated by an hyperplane.

\begin{claim}
    If $a \in \mathbb{R}$ and $f \in \mathbb{H}$, then $a \, f \in \mathbb{H}$.
\end{claim}
	Consider the scalar $a \in \mathbb{R}$ and the SRT of depth $D$, with $D \geq 1$, implementing the function $f(\textbf{x})$. 
    Then
	$a\, f(\mathbf{x}) = a \sum_{t = 1}^{2^D} 
	\pi_{\mathbf{x}t}(\bm{\w}_{A(t)})\beta_{0t} = \sum_{t = 1}^{2^D} \pi_{\mathbf{x}t}(\bm{\w}_{A(t)})(a\beta_{0t}) = \sum_{t = 1}^{2^D} \pi_{\mathbf{x}t}(\bm{\w}_{A(t)})\tilde{\beta}_{0t}$.

\begin{claim}
If $f_1, f_2 \in \mathbb{H}$, then $f_1 \, f_2 \in \mathbb{H}$.
\end{claim}

\noindent \textcolor{black}{Consider the SRT $T_1$ of depth $D_1$ and the SRT $T_2$ of depth $D_2$ which implement the real-valued functions $f_1$ and, respectively, $f_2$ defined over $\mathbb{X}$. Without loss of generality, assume that $D_1 \geq D_2$. By indexing with $t_1$ the generic leaf node of $T_1$ and with $t_2$ the generic leaf node of $T_2$, we have:
\begin{equation}
\label{eq:f_1}
f_1(\mathbf{x}) =  \sum_{t_1 = 1}^{2^{D_1}}   \pi_{\mathbf{x}t_1}(\bm{\w}^1_{A(t_1)})\beta^1_{0t_1}
\end{equation}
and 
\begin{equation}
\label{eq:f_2}
f_2(\mathbf{x}) =  \sum_{t_2 = 1}^{2^{D_2}}  \pi_{\mathbf{x}t_2}(\bm{\w}^2_{A(t_2)})\beta^2_{0t_2}.
\end{equation}
}

\textcolor{black}{It is important to recall that, according to \eqref{eq:pi_approx}, for every input vector $\mathbf{x}$ exactly one $\pi_{\mathbf{x}t_1}(\bm{\w}^1_{A(t_1)})$ and one $\pi_{\mathbf{x}t_2}(\bm{\w}^2_{A(t_2)})$ are nonzero in \eqref{eq:f_1} and, respectively, in \eqref{eq:f_2}. Thus, $f_1(\mathbf{x})\, f_2(\mathbf{x})= \beta^1_{0{\hat t}_1(\mathbf{x})} \beta^2_{0{\hat t}_2(\mathbf{x})}$ where ${\hat t}_1(\mathbf{x})$ and ${\hat t}_2(\mathbf{x})$ are the indices of the corresponding leaf nodes of $T_1$ and $T_2$  such that $\pi_{\mathbf{x}{\hat t}_1(\mathbf{x})}(\bm{\w}^1_{A(t_1)})$ and $\pi_{\mathbf{x}{\hat t}_2(\mathbf{x})}(\bm{\w}^2_{A(t_2)})$ are nonzero.}

To verify that $f_1 \, f_2 \in \mathbb{H}$, we consider the tree $T_3$ of depth $D_1+D_2$ obtained by ``stacking" copies of the tree $T_2$ of depth $D_2$ under the tree $T_1$ of depth $D_1$, namely, by replacing each leaf node of $T_1$ with the root node of a copy of the tree $T_2$ of depth $D_2$.

\begin{figure}[htb!]
\centering
\includegraphics[width=0.7\textwidth]{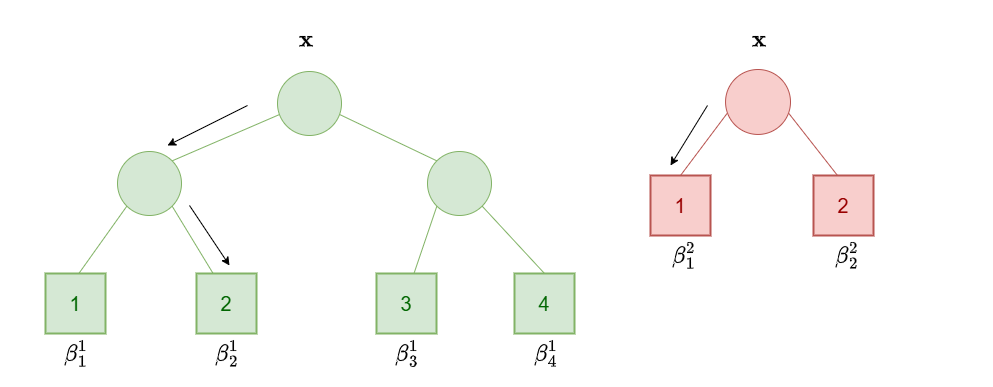}
\caption{Example of 2 SRTs of depth $D_1=2$ (green) and, respectively, $D_2 = 1$ (red),  with an input vector $\mathbf{x}$ routed along the two trees.}\label{fig:trees_nostack}
\end{figure}

\begin{figure}[htb!]
 \centering
 \includegraphics[width=0.6\textwidth]{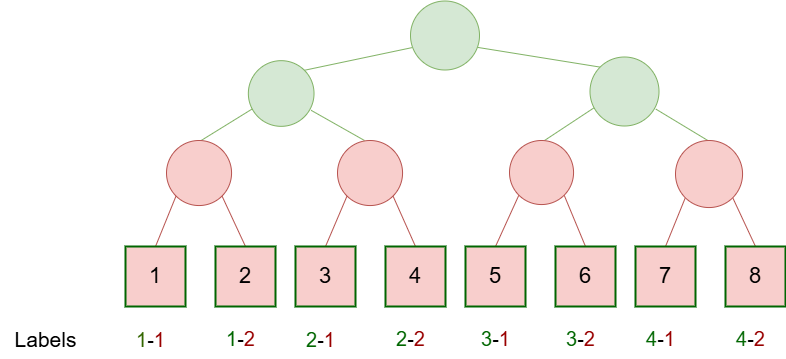}   \caption{The new SRT of depth $D = 3$ generated by stacking the green and the red SRTs. Each leaf node is uniquely identified by a pair of numbers in which the first term represents the index belonging to the leaf of the SRT of depth $D_1$ and the second term the index of the SRT of depth $D_2$.}\label{fig:tree_stack}
 \label{fig:three sin x}
\end{figure}

Figure \ref{fig:trees_nostack} illustrates an example involving two SRTs $T_1$ and $T_2$ of depth $D_1=2$ and, respectively, $D_2=1$. Figure \ref{fig:tree_stack} reports the resulting SRT $T_3$ of total depth $D_1+D_2=3$, where each leaf node of $T_3$, indexed with $t = 1, \dots, 2^{D_1 + D_2}$, is uniquely labeled by a pair of indices $(t_1, t_2)$ associated to the original SRTs $T_1$ and $T_2$.  \textcolor{black}{In particular, the first term $t_1$ of the pair indicates the leaf node index of $T_1$ to which the copy of $T_2$ has been stacked and is determined as $t_1 = { \lceil} { \lceil}\frac{t}{2^{D_2}}{ \rceil}\, \text{mod}\, (2^{D_1}+1){ \rceil}$, while the second term $t_2$ is the index of the corresponding leaf node of $T_2$ and is given by $[(t-1)\, \text{mod}\, 2^{D_2}]+1$.}
\textcolor{black}{Notice that the above labels $(t_1,t_2)$ of each leaf node $t$ in $T_3$ allow to easily determine the ancestors of $t$  as the ones of $t_1$ in $T_1$ and of $t_2$ in $T_2$.}

\textcolor{black}{Thus, for each leaf nodes $t= 1,\dots,2^{D_1+D_2}$ of SRT $T_3$ representing function $f_1 \, f_2$,  expression \eqref{eq:pi_approx} turns into:}

\begin{figure}[H]
    \centering
    \includegraphics[width=0.6\textwidth]{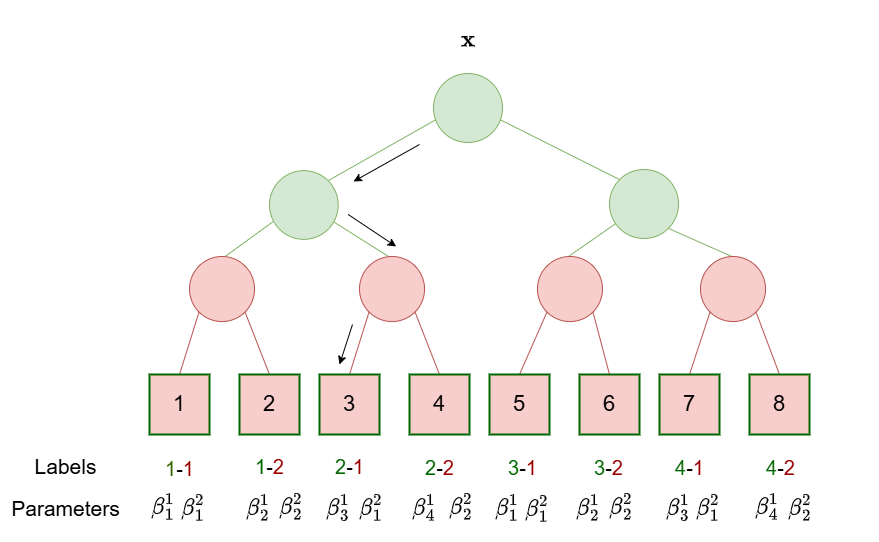}
    \caption{SRT of depth $D = 3$ generated by stacking the two SRTs with the new coefficients $\bm{\beta}$ for multiplication with an input vector $\mathbf{x}$ routed along the two trees.}
    \label{fig:albero_molti}
\end{figure}

\vspace{-30pt}
\textcolor{black}{
\begin{multline*}
    \tilde{\pi}_{\mathbf{x}t}(\tilde{\bm{\w}}_{A(t)}) = \pi_{\mathbf{x}t_1} (\bm{\w}_{A(t_1)}) \pi_{\mathbf{x}t_2} (\bm{\w}_{A(t_2)}) \\=\left(\prod_{{\ell^1}\in N_{L(t_1)}}
  \mathds{1}_{0.5}(\frac{1}{1+exp\left ( -(\bm{\w}_{\ell^1}^{1T}\mathbf{x})  \right )}) \;\prod_{r^1\in N_{R(t_1)}}\mathds{1}_{0.5}(1 - \frac{1}{1+exp\left ( -(\bm{\w}_{r^1}^{1T}\mathbf{x})  \right )}) \right)\,  \\ \left(\prod_{{\ell^2}\in N_{L(t_2)}}
  \mathds{1}_{0.5}(\frac{1}{1+exp\left ( -(\bm{\w}_{\ell^2}^{2T}\mathbf{x})  \right )}) \;\prod_{{r^2}\in N_{R(t_2)}}\mathds{1}_{0.5}(1 - \frac{1}{1+exp\left ( -(\bm{\w}_{r^2}^{2T}\mathbf{x})  \right )})\right)
\end{multline*}
    where $\tilde{\bm{\w}}_{A(t)} = (\bm{\w}^1_{A(t_1)},\bm{\w}^2_{A(t_2)}) = (\bm{\w}^1_{A({ \lceil} { \lceil}\frac{t}{2^{D_2}}{ \rceil}\, \text{mod}\, (2^{D_1}+1){ \rceil})},\bm{\w}^2_{A([(t-1)\, \text{mod}\, 2^{D_2}]+1)})$ and the corresponding
}
\hspace{-2pt}$\tilde{\beta}_{0t}$ for $\,t = 1,\dots, 2^{D_1+D_2}$ are as follows:
$$\tilde{\beta}_{0t} = \beta^1_{0{ \lceil} { \lceil}\frac{t}{2^{D_2}}{ \rceil}\, \text{mod}\, (2^{D_1}+1){ \rceil}} \, \beta^2_{[(t-1)\, \text{mod}\, 2^{D_2}]+1}\quad \text{for}\,\, t= 1,\dots,2^{D_1+D_2}.$$
See Figure \ref{fig:albero_molti} for an example with $T_1$ and $T_2$ of depth $D_1=2$ and, respectively, $D_2=1$.


\begin{claim}
If $f_1, f_2 \in \mathbb{H}$, then $f_1+f_2 \in \mathbb{H}$.
\end{claim}

\noindent \textcolor{black}{We proceed as in the proof of the previous \textbf{Claim 4}. Let $T_1$ and $T_2$ be the SRTs of depths $D_1$ and $D_2$, respectively, which implement the real-valued functions $f_1$ and $f_2$ over $\mathbb{X}$, with $D_1 \geq D_2$. By indexing again with $t_1$ the generic leaf node of $T_1$ and with $t_2$ the generic leaf node of $T_2$, we have:
$$f_1(\mathbf{x}) =  \sum_{t_1 = 1}^{2^{D_1}} \pi_{\mathbf{x}t_1}( \bm{\w}^1_{A(t_1)})\beta^1_{0t_1}$$ and $$f_2(\mathbf{x}) =  \sum_{t_2 = 1}^{2^{D_2}} \pi_{\mathbf{x}t_2}( \bm{\w}^2_{A(t_2)})\beta^2_{0t_2}.$$}

\textcolor{black}{Similarly to the previous claim, for every input vector $\mathbf{x}$, exactly one $\pi_{\mathbf{x}t_1}(\bm{\w}^1_{A(t_1)})$ are nonzero in \eqref{eq:f_1}, and  exactly one $\pi_{\mathbf{x}t_2}(\bm{\w}^2_{A(t_2)})$ are nonzero in \eqref{eq:f_2}. Thus, $f_1(\mathbf{x}) + f_2(\mathbf{x})= \beta^1_{0{\hat t}_1(\mathbf{x})} + \beta^2_{0{\hat t}_2(\mathbf{x})}$ where ${\hat t}_1(\mathbf{x})$ and ${\hat t}_2(\mathbf{x})$ are the indices of the leaf nodes of $T_1$ and $T_2$, respectively, such that $\pi_{\mathbf{x}{\hat t}_1(\mathbf{x})}(\bm{\w}^1_{A(t_1)})$ and $\pi_{\mathbf{x}{\hat t}_2(\mathbf{x})}(\bm{\w}^2_{A(t_2)})$ are nonzero.}

\textcolor{black}{Following the same reasoning of \textbf{Claim 4}, we can construct a SRT $T_3$ of depth $D_1 + D_2$ by ``stacking" copies of the tree $T_2$ (of depth $D_2$) under $T_1$ (of depth $D_1$), where each leaf node of $T_1$ is replaced by the root of a copy of $T_2$.}

For each leaf nodes $t= 1,\dots,2^{D_1+D_2}$ of SRT $T_3$ representing function $f_1 + f_2$,  expression \eqref{eq:pi_approx} turns into:


\textcolor{black}{
\begin{multline*}
    \tilde{\pi}_{\mathbf{x}t}(\tilde{\bm{\w}}_{A(t)}) = \pi_{\mathbf{x}t_1} (\bm{\w}_{A(t_1)}) \pi_{\mathbf{x}t_2} (\bm{\w}_{A(t_2)}) \\=\left(\prod_{{\ell^1}\in N_{L(t_1)}}
  \mathds{1}_{0.5}(\frac{1}{1+exp\left ( -(\bm{\w}_{\ell^1}^{1T}\mathbf{x})  \right )}) \;\prod_{r^1\in N_{R(t_1)}}\mathds{1}_{0.5}(1 - \frac{1}{1+exp\left ( -(\bm{\w}_{r^1}^{1T}\mathbf{x})  \right )}) \right)\,  \\ \left(\prod_{{\ell^2}\in N_{L(t_2)}}
  \mathds{1}_{0.5}(\frac{1}{1+exp\left ( -(\bm{\w}_{\ell^2}^{2T}\mathbf{x})  \right )}) \;\prod_{{r^2}\in N_{R(t_2)}}\mathds{1}_{0.5}(1 - \frac{1}{1+exp\left ( -(\bm{\w}_{r^2}^{2T}\mathbf{x})  \right )})\right)
\end{multline*}
    where $\tilde{\bm{\w}}_{A(t)} = (\bm{\w}^1_{A(t_1)},\bm{\w}^2_{A(t_2)}) = (\bm{\w}^1_{A({ \lceil} { \lceil}\frac{t}{2^{D_2}}{ \rceil}\, \text{mod}\, (2^{D_1}+1){ \rceil})},\bm{\w}^2_{A([(t-1)\, \text{mod}\, 2^{D_2}]+1)})$ and the corresponding
}
\hspace{-2pt}$\tilde{\beta}_{0t}$ for $\,t = 1,\dots, 2^{D_1+D_2}$ are as follows:
$$\tilde{\beta}_{0t} = \beta^1_{0{ \lceil} { \lceil}\frac{t}{2^{D_2}}{ \rceil}\, \text{mod}\, (2^{D_1}+1){ \rceil}} + \beta^2_{[(t-1)\, \text{mod}\, 2^{D_2}]+1}\quad \text{for}\,\, t= 1,\dots,2^{D_1+D_2}.$$

\begin{figure}[H]
    \centering
    \includegraphics[width=0.6\textwidth]{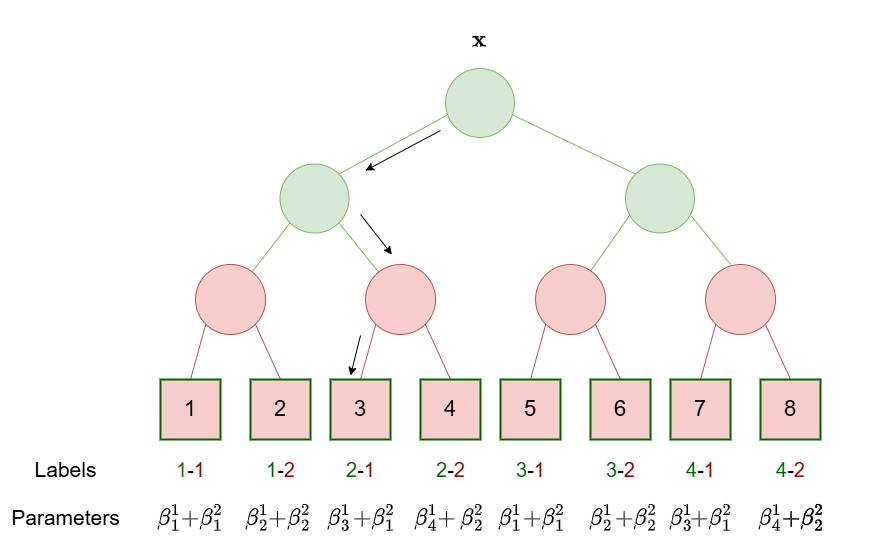}
    \caption{Tree of depth $D = 3$ generated by stacking the two trees with the new $\bm{\beta}$ coefficients for the sum with an input vector $\mathbf{x}$ routed along the two combined trees.}
    \label{fig:albero_somma}
\end{figure}

Figure \ref{fig:albero_somma} illustrates an example for the summation with two trees of depths $D_1=2$ and $D_2=1$.

Given Claims 1 to 5, the class of functions $\mathbb{H}$ satisfies the Conditions $i)$ to $v)$ of Theorem \ref{th:stone_weierstrass} and thus it is dense in $\mathbb{C}(\mathbb{X})$.

\qedblack
\section{Convergence analysis of the decomposition methods for SRTs}
\label{appendix:proofs_dec1}
\setcounter{condition}{0}
\setcounter{algorithm}{2}
\subsection{Asymptotic Convergence of NODEC-GS}\label{subsec:ec-1-ndgs}

Recall the assumption of  $\lambda_{\bm\w},\lambda_{\bm\beta}>0$, implying (by coercivity) that  \eqref{eq:obj} admits a global minimizer.\par 
For the sake of readability the three convergence conditions introduced in Section \ref{subsec:convergence}, are reported here.\par

\begin{condition}
Assume that NODEC-GS generates an infinite sequence $\{(\bm\w^k,\bm\beta^k)\}$. Then there exists $M>0 \in \mathbb{N}$ such that, for every $\bar t \in \tau_B$, for every $\hat t \in \tau_L$, and for every $k\geq 0$, there exist iteration indices $s,s'$, with $k \leq s \leq k+M$ and $k \leq s' \leq k+M$, such that $\bar t \in W_{B}^s$ and $\hat t \in W_{L}^{s'}$.
\end{condition}


\begin{condition}
For all $k$ it holds:
    \begin{equation}\label{eq:descent-w}
        E(\bm\w^{k+1},\bm\beta^k)\leq E(\bm\w^{k},\bm\beta^k).
    \end{equation}
    Moreover, for every $t \in \tau_B$ and for every infinite subsequence $K$ such that $t \in W_{B}^k$ with $k\in K$, we have that:
    \begin{equation}\label{eq:proximal-w1}
        \lim\limits_{k \to \infty, k \in K} \|\bm\w_t^{k+1}-\bm\w_t^k\|=0,
    \end{equation}
    \begin{equation}\label{eq:convergence-w}
        \lim\limits_{k \to \infty, k \in K} \nabla_{\bm\w_t}E(\bm\w^{k},\bm\beta^k)=0.
    \end{equation}
\end{condition}

\begin{condition}
For all $k$ it holds:
    \begin{equation}\label{eq:descent-b}
        E(\bm\w^{k+1},\bm\beta^{k+1})\leq E(\bm\w^{k+1},\bm\beta^k).
    \end{equation}
    Moreover, for every $t \in \tau_L$ and for every infinite subsequence $K$ such that $t \in W_{L}^k$ with $k\in K$, we have that:
    \begin{equation}\label{eq:convergence-b}
        \lim\limits_{k \to \infty, k \in K} \nabla_{\bm\beta_t}E(\bm\w^{k+1},\bm\beta^{k+1})=0.
    \end{equation}

\end{condition}

As highlighted in Section \ref{subsec:convergence}, Condition 3 for the {\bf LN Step} is similar to Condition 2 but it is less restrictive since, as proved in the following Lemma 1, the requirement of the distance between successive iterates tending to zero is automatically ensured by the strict convexity (provided $\lambda_{\bm\beta}>0$) of the involved minimization problem.  

\noindent \textsc{Lemma 1} {\it Let $\{(\bm\w^k,\bm\beta^k)\}$ be an infinite sequence of points in  the level set ${\cal L}_0$ of \eqref{eq:obj} and assume that for all $k$ it holds:
\begin{equation}\label{eq:mono}
    E(\bm\w^{k+1},\bm\beta^{k+1})\leq E(\bm\w^{k+1},\bm\beta^{k}) \leq E(\bm\w^{k},\bm\beta^{k}).
\end{equation}
Moreover, suppose that there exists a subsequence $\{(\bm\w^k,\bm\beta^k)\}_K$ and a non-empty index set $J \subseteq \tau_{L}$, such that for all $k \in K$ we have $W_{L}^k= J$, i.e.,
\begin{equation}\label{eq:I-update}
    \bm\beta^{k+1}=({\bm\beta_{J}^{k+1}}^T,{\bm\beta_{\bar J}^{k+1}}^T)^T \mbox{ with } {\bm\beta_{\bar J}^{k+1}}={\bm\beta_{\bar J}^{k}},
\end{equation}
and suppose also that 
\begin{equation}\label{eq:convergence-b2}
    \lim\limits_{k \to \infty, k \in K} \nabla_{J}E(\bm\w^{k+1},\bm\beta^{k+1})=0.
\end{equation}
Then 
\begin{equation}\label{eq:proximal-b}
    \lim\limits_{k \to \infty, k \in K} \|\bm\beta_J^{k+1}-\bm\beta_J^k\|=0.
\end{equation}
}


\noindent \textbf{Proof}.

\noindent  From \eqref{eq:mono} and the fact that $E(\bm\w,\bm\beta)$ admits a global minimizer (${\cal L}_0$ is compact), the sequence $\{E(\bm\w^k,\bm\beta^k)\}$ is monotonically non-increasing and bounded below, so it converges to a limit, and it holds
\begin{equation}\label{eq:seq2zero}
   \lim\limits_{k \to \infty, k \in K}  E(\bm\w^{k+1},\bm\beta^{k}) -  \{E(\bm\w^{k+1},\bm\beta^{k+1})\}=0.
\end{equation}
Considering that function $E(\cdot)$ is quadratic with respect to variables $\bm\beta$, from \eqref{eq:I-update} we can write for all $k \in K$
\begin{equation}\label{eq:taylor}
    E(\bm\w^{k+1},\bm\beta^k)=E(\bm\w^{k+1},\bm\beta^{k+1})+\nabla_{J}E(\bm\w^{k+1},\bm\beta^{k+1})^T(\bm\beta_{J}^{k}-\bm\beta_{J}^{k+1})+\frac{1}{2}(\bm\beta_{J}^{k}-\bm\beta_{J}^{k+1})^T H_J^k (\bm\beta_{J}^{k}-\bm\beta_{J}^{k+1}),
\end{equation}
where $H_J^k \in \mathbb{R}^{|J|(p+1) \times |J|(p+1)}$ is the partial Hessian matrix of $E(\bm\w^{k+1},\bm\beta^{k+1})$ relative to $\bm\beta_J$, which does not depend on $\bm\beta^{k+1}$. The strict convexity of $E(\cdot)$ implies that the smallest eigenvalue of $H_J^k$, namely $\sigma_{\mbox{min}}(H_{J}^k)$, satisfies $\sigma_{\mbox{min}}(H_{J}^k)>0$. Hence, from \eqref{eq:taylor} we have
\begin{equation}\label{eq:taylor-dis}
     E(\bm\w^{k+1},\bm\beta^k)-E(\bm\w^{k+1},\bm\beta^{k+1})\geq -\|\nabla_{J}E(\bm\w^{k+1},\bm\beta^{k+1})\| \|\bm\beta_{J}^{k}-\bm\beta_{J}^{k+1} \| +\frac{1}{2}\sigma_{\mbox{min}}(H_{J}^k) \| \bm\beta_{J}^{k}-\bm\beta_{J}^{k+1} \|^2.
\end{equation}
By recalling \eqref{eq:convergence-b2} and \eqref{eq:seq2zero}, taking the limits in \eqref{eq:taylor-dis}, we obtain \eqref{eq:proximal-b} and the proof is complete.

\qedblack

Before stating the convergence of NODEC-GS in Proposition 1, we recall a technical lemma for a general sequence of points $\{x^k\}$ in $\mathbb{R}^n$, whose proof is reported in \citep{grippo2015decomposition}.

\noindent \textsc{Lemma 2} {\it Let $\{x^k\}$, with $x^k\in \mathbb{R}^n$, be a sequence such that
\begin{equation}\lim\limits_{k\to\infty}\|x^{k+1}-x^k\|=0,\end{equation}
 and let $\bar x$ be  a limit point,
so that there exists a subsequence $\{x^k\}_K$
 converging to  $\bar x$. Then, for every fixed integer
 $M>0$, we have that every subsequence $\{x^{k+j}\}_K$,
for $j=1,\dots,M$ converges to $\bar x$.
}

\noindent \textsc{Proposition 1} {\it Suppose that NODEC-GS generates an infinite sequence
 $\{(\bm\w^k, \bm\beta^k)\}$ and that Conditions 1, 2 and 3  are
 satisfied. Then:
\begin{description}
 \item[(i)] the sequence $\{(\bm\w^k, \bm\beta^k)\}$ has limit points
 \smallskip
\item[(ii)] the sequence $\{E(\bm\w^k, \bm\beta^k)\}$ converges to a
limit \smallskip
 \item[(iii)] $\lim\limits_{k\to\infty}\|\bm\w^{k+1}-\bm\w^k\|=0$
\smallskip
 \item[(iv)] $\lim\limits_{k\to\infty}\|\bm\beta^{k+1}-\bm\beta^k\|=0$
\smallskip
 \item[(v)] every limit point of $\{(\bm\w^k,
\bm\beta^k)\}$ is a stationary point of $E(\bm\w,
\bm\beta)$ in \eqref{eq:obj}.
\end{description}

}

\noindent \textbf{Proof}.

\noindent From the instructions of NODEC-GS, it follows that
\begin{equation}\label{eq:mono1}
    E(\bm\w^{k+1},\bm\beta^{k+1})\leq E(\bm\w^{k+1},\bm\beta^{k}) \leq E(\bm\w^{k},\bm\beta^{k}).
\end{equation}
Therefore all points of the sequence $\{(\bm\w^k, \bm\beta^k)\}$ 
lie in the compact level set ${\cal L}_0$, and then it admits limit points demonstrating assertion (i). Moreover, sequence $\{E(\bm\w^k, \bm\beta^k)\}$, being monotonically non-increasing and bounded below ($E(\bm\w,
\bm\beta)$ admits a global minimizer), converges to a limit, so assertion (ii) must hold. Moreover, also $\{E(\bm\w^{k+1},\bm\beta^{k})\}$ converges to the same limit.\par
Now, to prove assertion (iii), let us assume, by contradiction that (iii) is false. Then, there exists a subsequence $\{(\bm\w^k, \bm\beta^k)\}_K$, and a number $\varepsilon>0$
such that, for all sufficiently large $k\in K$, say $k\ge \hat k$, it holds
\begin{equation}\label{eq:contradw}
\|\bm\w^{k+1}-\bm\w^k\|\ge\varepsilon,\;\hbox{for all $k\in
K$, $k\ge \hat k$},
\end{equation}
implying that, for all $k\in K$, $k\ge \hat k$, the index set $W_{B}^k$ considered in NODEC-GS must be non-empty. Then, since the number of components in $\bm\w$ are finite, we can find a further subsequence $\{(\lambda^k, w^k)\}_{K_1}$, with $K_1\subseteq K$, and an index
 set $J \subseteq \tau_{B}$, with $W_{B}^k= J$, such that from (\ref{eq:contradw}), we
 get:
 \begin{equation}\label{contradw1}
 \|\bm\w_{J}^{k+1}-\bm\w_{J}^k\|\ge \varepsilon,\;\hbox{for all
 $k\in K_1$, $k\ge \hat k$}.
 \end{equation}
Since $J$ is non-empty, by the instructions at the {\bf BN Step} and recalling Condition 2, we have that $$\lim\limits_{k \to \infty, k \in K} \|\bm\w_t^{k+1}-\bm\w_t^k\|=0, \mbox{ with } t \in J,$$ which contradicts \eqref{eq:contradw}. So, assertion (iii) must hold.\par
The proof of assertion (iv) proceeds analogously. Assume that there exist a subsequence $\{(\bm\w^k, \bm\beta^k)\}_K$, and a number $\varepsilon>0$
such that, for all sufficiently large $k\in K$, say $k\ge \hat k$, it holds
\begin{equation}\label{eq:contradb}
\|\bm\beta^{k+1}-\bm\beta^k\|\ge\varepsilon,\;\hbox{for all $k\in
K$, $k\ge \hat k$},
\end{equation}
implying that, for all $k\in K$, $k\ge \hat k$,  the index set $W_{L}^k$ considered in NODEC-GS must be non-empty. Moreover, as the number of components of $\bm\beta$ is finite, we can find a further subsequence $\{(\bm\w^k, \bm\beta^k)\}_{K_1}$, with $K_1\subseteq K$, and an index set $J \subseteq \tau_{L}$, such that the set $W_{L}^k$ considered in NODEC-GS is such that $W_{L}^k= J$, and hence from
 \eqref{eq:contradb}, we get:
  \begin{equation}\label{contradb1}
 \|\bm\beta_{J}^{k+1}-\bm\beta_{J}^k\|\ge \varepsilon,\;\hbox{for all
 $k\in K_1$, $k\ge \hat k$}.
 \end{equation}
 Since $J$ is non-empty, by the instructions at the {\bf LN Step} and recalling Condition 3, we have that
 $$\lim_{k\in K_1,k\to\infty} \nabla_{\bm\beta_J}
 E(\bm\w^{k+1},\bm\beta^{k+1})=0,$$
 and therefore, from Lemma 1, we get a contradiction to \eqref{eq:contradb} and this
 implies that assertion (iv) must hold.\par 
 Finally let us prove assertion (v). Reasoning by contradiction, let us assume that there exists a
subsequence $\{(\bm\w^k, \bm\beta^k)\}_{K}$ converging to a point
$(\bar{\bm\w},\bar{\bm\beta})$ and such that $\nabla E(\bar{\bm\w},\bar{\bm\beta})\neq 0$. 
Suppose first there exist an index, say $t \in \tau_{B}$, and a number $\theta>0$, such that
 \begin{equation}\label{eq:contradstat1}
 \|\nabla_{\bm\w_{t}}E(\bar{\bm\w},\bar{\bm\beta})\|\ge \theta.
 \end{equation}
 By Condition 1, there exists another subsequence denoted as $\{(\bm\w^{s_k},\bm\beta^{s_k})\}_K$, such that $t \in W_{B}^{s_k}$ and that $k\leq s_k\leq k+M$ for $k\in K$ and $M$ a given positive integer. From assertions (iii)-(iv) and Lemma 2 we have that $\{\bm\w^{s_k},\bm\beta^{s_k}\}$ converges to $(\bar{\bm\w},\bar{\bm\beta})$. On the other hand, since $t \in W_{B}^{s_k}$, by Condition 2 and by continuity we have that $\lim\limits_{k\to\infty,k \in K} \nabla_{\bm\w_{t}}E(\bm\w^{s_k},\bm\beta^{s_k})=\nabla_{\bm\w_{t}}E(\bar{\bm\w},\bar{\bm\beta})=0$, which contradicts \eqref{eq:contradstat1}.\par 
 Suppose now that there exist an index, say $t \in \tau_{L}$ and a number $\theta>0$, such that
 \begin{equation}\label{eq:contradstat2}
 \|\nabla_{\bm\beta_{t}}E(\bar{\bm\w},\bar{\bm\beta})\|\ge \theta.
 \end{equation} 
 Again, by Condition 1, there exists another subsequence denoted as $\{(\bm\w^{s_k},\bm\beta^{s_k})\}_K$, such that $t \in W_{L}^{s_k}$ and that $k\leq s_k\leq k+M$ for $k\in K$ and $M$ a given positive integer. From assertions (iii)-(iv) and Lemma 2 we have that both $\{\bm\w^{s_k},\bm\beta^{s_k}\}$ and $\{\bm\w^{s_k +1},\bm\beta^{s_k +1}\}$ converge to $(\bar{\bm\w},\bar{\bm\beta})$.
 On the other hand, since $t \in W_{L}^{s_k}$, by Condition 3 we have that $\lim\limits_{k\to\infty,k \in K} \nabla_{\bm\beta_{t}}E(\bm\w^{s_k+1},\bm\beta^{s_k+1})=\nabla_{\bm\beta_{t}}E(\bar{\bm\w},\bar{\bm\beta})=0$, which contradicts \eqref{eq:contradstat2}.\par
 This concludes the proof.
 \qedblack

\subsection{Theoretical properties of the Armijo method applied to the SRT framework}\label{subsec:ec-1-armjio}
\medskip

Let us consider the adaptation of the Armijo line search procedure to the {\bf BN Step} of NODEC-GS, whose scheme is reported below.
\smallskip

\renewcommand{\thealgorithm}{} 
\begin{algorithm}[H]
\floatname{algorithm}{Armijo update} 
\caption{}
\label{alg:Armijo}
$\bm{Input}: a\,>\,0;\, \gamma\, \in\, (0,1);\, \delta\,\in\,(0,1);\,(\bm{\w},\bm{\beta});\,J\\$ 
$\bm{Output}:\alpha_J.$
{\fontsize{10}{12}\selectfont
\begin{algorithmic}[1]
\Procedure{ArmijoUpdate}{$a,\gamma,\delta,(\bm{\w},\bm{\beta}),J$}
\State $\alpha \gets a$
\While{$ E(\bm{\w}_{J} + \alpha \bm{d}_{J},\bm{\w}_{\bar J},\bm{\beta}) < E(\bm{\w},\bm{\beta}) -\gamma \alpha \|\nabla_{J}E(\bm{\w},\bm{\beta}) \|^2 $}
\State $\alpha \gets \delta \alpha$

\EndWhile
\State $\alpha_J \gets \alpha$
\State \textbf{return} $\alpha_J$
\EndProcedure
\end{algorithmic}
}
\end{algorithm}

\noindent \textsc{Proposition 2}
{\it Let $J \subseteq \tau_B$ be a fixed set of indices and let  $\{(\bm{\w}^k,\bm{\beta}^k)\}_K$ be an infinite subsequence such that $$\nabla_{J} E(\bm{\w}^k,\bm{\beta}^k) \neq 0 $$ and assume that, for all $k \in K$, the step length $\alpha^k_{J}$ is calculated by Armijo method \ref{alg:Armijo} along the direction $\bm{d}_{J}^k = - \nabla_J E(\bm{\w}^k,\bm{\beta}^k)$.

Then, the algorithm terminates yielding a stepsize $\alpha^k_{J} > 0$ such that the following hold.
\begin{enumerate}
    \item[(i)] For all $k \in K$, we have
    \begin{equation}
        E(\bm{\w}_{J}^{k} + \alpha^k_{J} \bm{d}_{J}^k,\bm{\w}_{\bar J}^k,\bm{\beta}^k) < E(\bm{\w}^k,\bm{\beta}^k)
    \end{equation}
    \item[(ii)] If the following limit is valid:
        \begin{equation}
            \lim\limits_{k \to \infty, k \in K} E(\bm{\w}^k,\bm{\beta}^k) - E(\bm{\w}_{J}^{k} + \alpha^k_{J} \bm{d}_{J}^k,\bm{\w}_{\bar J}^k,\bm{\beta}^k)=0
        \end{equation}
        then 
            \begin{enumerate}
            \item[a)] $ \lim\limits_{k \to \infty, k \in K} \alpha^k_{J} \|\bm{d}^k_{J} \|=0$.
            \item[b)] If $\{(\bm{\w}^k,\bm{\beta}^k)\}_{K_1},\, K_1 \subseteq K$ is a convergent subsequence, we have 
            \begin{equation}
                \lim\limits_{k \to \infty, k \in K_1} \nabla_{J}E(\bm{\w}^k,\bm{\beta}^k)=0.
            \end{equation}
            \end{enumerate}
\end{enumerate}
}

\noindent \textbf{Proof}.

\noindent  It is well known (see e.g., \cite{Bertsekas/99}) that, since the objective function in \eqref{eq:obj} is continuously differentiable and $0 <\gamma <1$, for fixed $J$ and $k$ the Armjio procedure terminates, yielding a step size $\alpha^k_J > 0$ such that 

\begin{equation}
    E(\bm{\w}_{J}^{k} + \alpha_J^k \bm{d}_{J}^k,\bm{\w}_{\bar J}^k,\bm{\beta}^k) \leq E(\bm{\w}^k,\bm{\beta}^k) -\gamma \alpha_J^k \nabla_{J}E(\bm{\w}^k,\bm{\beta}^k)^T \bm{d}_J^k  < E(\bm{\w}^k,\bm{\beta}^k),
\end{equation}

\noindent implying $(i)$. Moreover we have 

\begin{equation}
\label{eq:d20}
    E(\bm{\w}^k,\bm{\beta}^k) - E(\bm{\w}_{J}^{k} + \alpha_J^k \bm{d}_{J}^k,\bm{\w}_{\bar J}^k,\bm{\beta}^k) \geq \gamma \alpha_J^k \|\bm{d}_J^k\|^2 \geq \frac{\gamma}{a} \|\alpha_J^k\bm{d}_J^k \|^2
\end{equation}
\noindent implying $(ii)-a)$. Now let assume that a subsequence $\{(\bm{\w}^k,\bm{\beta}^k)\}_{K_1},\, K_1 \subseteq K$ converges to a point $(\bm{\bar \w},\bm{\bar \beta})$. By continuity we have that 

\begin{equation}
             \label{eq:cont} \lim\limits_{k \to \infty, k \in K_1} \bm{d}_J^k = - \nabla_{J}E(\bm{\bar \w},\bm{\bar \beta}).
            \end{equation}
If we assume, by contradiction, that $(ii) - b)$ is false, then 

\begin{equation}
    \label{eq:contrad-a}
    \|\nabla_J E(\bm{\bar \w},\bm{\bar \beta}) \| > 0.
\end{equation}
Two cases are possible:
\begin{enumerate}
    \item[1)]  $\alpha_J^k = a$ for an infinite subsequence $k \in K_2 \subseteq K_1$,
    \item[2)] for all sufficiently large $k \in K_1$ we have $\alpha_J^k < \frac{\alpha_J^k}{\delta} \leq a. $
\end{enumerate}
Concerning 1), from  \eqref{eq:d20} we obtain $ \lim\limits_{k \to \infty, k \in K_2} \bm{d}_J^k = 0$, which contradicts \eqref{eq:contrad-a}. As to 2), since for all sufficiently large $k \in K_1$ the initial tentative step length is reduced, we have that 
\begin{equation}
E(\bm{\w}_{J}^{k} + \frac{\alpha_J^k}{\delta} \bm{d}_{J}^k,\bm{\w}_{\bar J}^k,\bm{\beta}^k) - E(\bm{\w}^k,\bm{\beta}^k) > \gamma \frac{\alpha^k_J}{\delta}\nabla_J E(\bm{\w}^k,\bm{\beta}^k)^T \bm{d}_J^k.
\end{equation}
By the Mean Theorem we have 
\begin{equation}
    E(\bm{\w}_{J}^{k} + \frac{\alpha_J^k}{\delta} \bm{d}_{J}^k,\bm{\w}_{\bar J}^k,\bm{\beta}^k)  = E(\bm{\w}^k,\bm{\beta}^k) + \frac{\alpha^k_J}{\delta}\nabla_J E(\bm{\w}_{J}^{k} + \xi_J^k\frac{\alpha_J^k}{\delta} \bm{d}_{J}^k,\bm{\w}_{\bar J}^k,\bm{\beta}^k)^T \bm{d}_J^k,
\end{equation}
for $\xi_J^k \in (0,1).$ Hence, we have that  
\begin{equation}\nabla_J E(\bm{\w}_{J}^{k} + \xi_J^k\frac{\alpha_J^k}{\delta} \bm{d}_{J}^k,\bm{\w}_{\bar J}^k,\bm{\beta}^k)^T \bm{d}_J^k >  \gamma \nabla_J E(\bm{\w}^k,\bm{\beta}^k)^T \bm{d}_J^k,
\end{equation}
which, considering that $\bm{d}_J^k = -\nabla_J E(\bm{\w}^k,\bm{\beta}^k)$, taking limits for $k \in K_1$, and recalling \eqref{eq:cont} $(ii) - a)$, yields 
\begin{equation}
\label{eq:normgrad}
    -\|\nabla_J E(\bm{\bar \w},\bm{\bar \beta}) \|^2 \geq -\gamma \|\nabla_J E(\bm{\bar \w},\bm{\bar \beta}) \|^2.
\end{equation}
Since $\gamma \in (0,1)$, \eqref{eq:normgrad} contradicts \eqref{eq:contrad-a} and the proof is complete. 
\qedblack

\setcounter{proposition}{5}

\subsection{Asymptotic convergence of NODEC-DR algorithm.}\label{subsec:prop3}
In this subsection we recall \textsc{Proposition 3} and provide the proof.

\medskip
\noindent \textsc{Proposition 3} {\it Given an infinite sequence $\{(\bm{\w}^k,\bm{\beta}^k)\}$ generated by the NODEC-DR algorithm. The following holds:
\begin{description}
    \item[(i)] $\{(\bm{\w}^k,\bm{\beta}^k)\}$ has a limit point
    \item[(ii)] The sequence $\{E(\bm{\w}^k,\bm{\beta}^k)\}$ converges to a limit when $k \rightarrow \infty$
    \item[(iii)] We have $\lim\limits_{k \to \infty}\|\bm{\beta}^{k+1} -\bm{\beta}^k\|=0 $
    \item[(iv)] For any $ t \in \tau_B$, we have $\lim\limits_{k \to \infty}\| \bm{\w}_{t}^{k+1} - \bm{\w}_{t}^{k} \|=0$
    \item[(v)] Every limit point of $\{(\bm{\w}^k,\bm{\beta}^k)\}$ is a stationary point of $E(\bm\w,
\bm\beta)$ in \eqref{eq:obj}.
\end{description}
}

\noindent \textbf{Proof}.

\noindent The goal is to prove that Conditions 1, 2, and 3 are satisfied in order to exploit the result of Proposition \ref{prop:prop1}. By assuming $M\_it=\infty$ and {\it termination criterion} $= \text{False}$ the algorithm generates an infinite sequence $\{(\bm{\w}^k,\bm{\beta}^k)\}$. Since we are interested in asymptotic convergence we can consider the infinite subsequence of  $\{(\bm{\w}^k,\bm{\beta}^k)\}$ for $k \geq k_0$ (i.e., when convergence conditions are enforced), that for simplicity we rename again as $\{(\bm{\w}^k,\bm{\beta}^k)\}$.

By considering that at each macro iteration $it$ the indices of each branch node and each leaf node are inserted at least once in $W_B$ and $W_L$ (see the for loop in NODEC-DR), Condition 1 is satisfied with $M = 2^D - 1$.


Concerning Condition 2, the branch node variables $\bm{\w}$ are updated in three possible ways:
\begin{itemize}
    \item[(i)] $\bm{\w}^{k+1} = \bm{\w}^{k}$ as $\| \nabla_{{W^k_B}}E(\bm{\w}^k,\bm{\beta}^k) \| \leq (\theta)^k,$
    \item[(ii)] $\bm{\w}^{k+1}$ is the result of  $\textsc{UpdateBranchNode}$ procedure, 
    \item[(iii)] $\bm{\w}^{k+1}$ is obtained by applying the  Armijo method along the direction $\bm{d}_{W_B^k}^k = -\nabla_{{W^k_B}}E(\bm{\w}^k,\bm{\beta}^k).$
\end{itemize}

In case (i), \eqref{eq:descent-w}-\eqref{eq:proximal-w1} are trivially satisfied and the same holds for \eqref{eq:convergence-w} as $\theta^k \to 0$ for $\theta \in (0,1)$. The update in (ii) is performed only if \eqref{eq:descreasing1} and hence \eqref{eq:descent-w} are satisfied, while \eqref{eq:descent-w} is automatically enforced by the Armijo step in case (iii) (recall Proposition 3-(i)). Now, the update in case (ii) has also to satisfy \eqref{eq:decreasing2}, while in the Armijo update (iii), recalling that $\alpha^k_{W^k_B} \leq a$ and the acceptance condition, we have

\begin{equation}
    E(\bm{\w}^{k+1},\bm{\beta}^k) \leq E(\bm{\w}^{k},\bm{\beta}^{k}) - \frac{\gamma}{a} \|\bm{\w}^{k+1}_{W_B^k} - \bm{\w}^{k}_{W_B^k} \|^2.
\end{equation}
Thus, in both cases (ii) and (iii), for every $W^k_B$ we have 
\begin{equation}
    \label{eq:descent_prop3}
    E(\bm{\w}^{k+1},\bm{\beta}^k) \leq E(\bm{\w}^{k},\bm{\beta}^{k}) - \hat{\xi} \|\bm{\w}^{k+1}_{W_B^k} - \bm{\w}^{k}_{W_B^k} \|^2.
\end{equation}

\noindent where $\hat{\xi}= \min \{\xi,\frac{\gamma}{a}\}$. By considering the instructions in the \textbf{LN Step} we have that 
\begin{equation}
\label{prop:decreasing_prop}
E(\bm{\w}^{k+1},\bm{\beta}^{k+1}) \leq E(\bm{\w}^{k+1},\bm{\beta}^{k}) \leq E(\bm{\w}^{k},\bm{\beta}^{k}),
\end{equation} 
which, by the compactness of the level set ${\cal L}_0$, the sequences of function values $\{E(\bm{w}^k,\bm{\beta}^k)\}$ and $ \{E(\bm{\w}^{k+1},\bm{\beta}^{k})\} $ converge to the same limit. Then, for an infinite subsequence $k \in K$ such that $t \in W_B^k$, we have that either $\bm{\w}_t^{k+1}= \bm{\w}_t^{k}$ or \eqref{eq:descent_prop3} holds, implying that \eqref{eq:proximal-w1} is satisfied.



To prove \eqref{eq:convergence-w}, and hence that Condition 2 holds, 
let assume by contradiction that there exists an infinite subset $K_1$ of the considered subsequence $K$, such that for an index $t \in W_B^k$, we have 
\begin{equation}
    \label{eq:contradiction_prop}
    \|\nabla_{\bm\w_t}E(\bm\w^{k},\bm\beta^k)\|\geq \eta > 0 \,\;\forall k \in K_1\
\end{equation}
By \eqref{prop:decreasing_prop} and the compactness of level set ${\cal L}_0$, it is possible to find another subsequence $\{(\bm{\w}^{k},\bm{\beta}^{k})\}_{K_2}$ with $K_2 \subseteq K_1$ that converges to a point $(\bm{\bar \w},\bm{\bar \beta})$ such that from \eqref{eq:contradiction_prop} we have

\begin{equation}
    \label{eq:gradient_greater}
    \|\nabla_{\bm\w_t}E(\bm{\bar \w},\bm{\bar \beta})\|\geq \eta.
\end{equation}
By \eqref{eq:contradiction_prop} and the instructions in the $\textbf{BN Step}$, we have that, for all large enough $k \in K_2$, the Armijo method is applied (at least to generate a reference point) and a step length $\alpha_{W_B^k}^k$ along the steepest direction $d_{W_B^k}^k$ is computed. As a consequence, for all large $k \in K_2$ we have 
\begin{equation}
    \label{eq:decreasing_contradiction}
    E(\bm{\w}^{k+1},\bm{\beta}^k) \leq E(\bm{\w}_{W_B^k}^{k} + \alpha_{W_B^k}^k \bm{d}_{W_B^k}^k,\bm{\w}_{\bar W_B^k}^k,\bm{\beta}^k) \leq E(\bm{\w}^k,\bm{\beta}^k)
\end{equation}
Moreover, since the sequence of function values is convergent, from \eqref{eq:decreasing_contradiction} we get
\begin{equation}
\lim\limits_{k \to \infty, k \in K_2} E(\bm{\w}_{W_B^k}^{k},\bm{\w}_{\bar W_B^k}^k,\bm{\beta}^k) - E(\bm{\w}_{W_B^k}^{k} + \alpha_{W_B^k}^k \bm{d}_{W_B^k}^k,\bm{\w}_{\bar W_B^k}^k,\bm{\beta}^k)=0.
\end{equation}
Then, from Proposition 2 we have
\begin{equation}
    \lim\limits_{k \to \infty, k \in K_2} \nabla_{W_B^k}E(\bm\w^{k},\bm\beta^k)=0,
\end{equation}
thus, since $t \in W_B^k$ and from the continuity of the gradient, we get $\lim\limits_{k \to \infty, k \in K_2} \nabla_{\bm\w_t}E(\bm\w^{k},\bm\beta^k) = \nabla_{\bm\w_t}E(\bm{\bar \w},\bm{\bar \beta}) = 0, $
which contradicts \eqref{eq:gradient_greater}.

Finally, by the instructions of the $\textbf{LN Step}$ and since $\upsilon \in (0,1)$ implies $(\upsilon)^k \to 0$, it is easy to see that Condition 3 is satisfied, and the proof is complete.
\qedblack

\section{Results for regressions trees of depth $2$ }\label{appendix:results-trees-depth-2}

\textcolor{black}{In this appendix we report the results of the numerical experiments obtained with regression trees of depth $D=2$.}


As in Table \ref{tab:depth3}, for each one of the three considered models and methods, namely, SRT $\ell_2$, ORT-L and ORRT, Table \ref{tab:depth2} indicates the average testing $R^2$, its standard deviation $\sigma$ (divided by a $1e^{-2}$ factor for visualization reasons) and the average training time in seconds. For ORRT, the numbers appearing in the $R^2<0$ column correspond to the number of times, out of the 80 runs, that the training phase found a suboptimal solution with a negative testing $R^2$ (i.e. worse than using the average as prediction). The arithmetic and the geometric averages are reported in the last two rows of the table.

The results in Table \ref{tab:depth2} are in line with those in Table \ref{tab:depth3}. Overall, SRT $\ell_2$ outperforms both ORRT and ORT-L in terms of average testing $R^2$, average $\sigma$, and average computational time.

\begin{table}[H]
\centering
\resizebox{\textwidth}{!}{\begin{tabular}{lccccccccccc}
\hline
                               & \multicolumn{1}{l}{}  & \multicolumn{1}{l}{}    & \multicolumn{9}{c}{D=2}                                                                                                                                                                                      \\ \cline{4-12} 
                               & \multicolumn{1}{l}{}  & \multicolumn{1}{l}{}    & \multicolumn{2}{c}{\textbf{SRT}}                             & \multicolumn{2}{c}{\textbf{SRT} \bm{$\ell_2$}}                    & \multicolumn{2}{c}{\textbf{ORT-L}}                             & \multicolumn{3}{c}{\textbf{ORRT}}                 \\ \cline{4-12} 
Dataset                        & \multicolumn{1}{l}{N} & \multicolumn{1}{l}{p}   & $R^2\, (\sigma\, 1e^{-2})$ & Time                       & $R^2\, (\sigma\, 1e^{-2})$ & Time                       & $R^2\, (\sigma\, 1e^{-2})$ & Time                         & $R^2\, (\sigma\, 1e^{-2})$ & Time  & $R^2<0$ \\ \hline
\multicolumn{1}{l|}{Abalone}   & 4177                  & \multicolumn{1}{c|}{8}  & 0.551 (2.18)           & \multicolumn{1}{c|}{8.6}   & 0.554 (2.24)           & \multicolumn{1}{c|}{8.7}   & 0.558 (1.7)            & \multicolumn{1}{c|}{70.6}    & 0.525 (3.32)           & 18.17 & 8       \\
\multicolumn{1}{l|}{Ailerons}  & 7154                  & \multicolumn{1}{c|}{40} & 0.833 (0.6)            & \multicolumn{1}{c|}{18.9}  & 0.836 (0.56)           & \multicolumn{1}{c|}{18.9}  & 0.824 (1.13)           & \multicolumn{1}{c|}{262.2}   & 0.771 (13.53)          & 427.7 & 45      \\
\multicolumn{1}{l|}{Airfoil}   & 1503                  & \multicolumn{1}{c|}{5}  & 0.762 (3.15)           & \multicolumn{1}{c|}{5.2}   & 0.748 (2.9)            & \multicolumn{1}{c|}{5.2}   & 0.744 (1.6)            & \multicolumn{1}{c|}{2.07}    & 0.516 (2.89)           & 1.81  & 4       \\
\multicolumn{1}{l|}{Auto mpg}  & 392                   & \multicolumn{1}{c|}{7}  & 0.853 (2)              & \multicolumn{1}{c|}{2.5}   & 0.87 (2.37)            & \multicolumn{1}{c|}{2.5}   & 0.838 (2.8)            & \multicolumn{1}{c|}{5.94}    & 0.811 (2.23)           & 0.89  & 17      \\
\multicolumn{1}{l|}{Compact}   & 8192                  & \multicolumn{1}{c|}{21} & 0.979 (0.25)           & \multicolumn{1}{c|}{35.5}  & 0.98 (0.168)           & \multicolumn{1}{c|}{37.2}  & 0.981 (0.34)           & \multicolumn{1}{c|}{668.4}   & 0.692 (13.1)           & 183.2 & 11      \\
\multicolumn{1}{l|}{Computer}  & 209                   & \multicolumn{1}{c|}{37} & 0.977 (1.86)           & \multicolumn{1}{c|}{1.9}   & 0.964 (2.49)           & \multicolumn{1}{c|}{1.9}   & 0.877 (13.4)           & \multicolumn{1}{c|}{1.9}     & 0.907 (5.5)            & 0.6   & 10      \\
\multicolumn{1}{l|}{Cpu small} & 8192                  & \multicolumn{1}{c|}{12} & 0.969 (0.32)           & \multicolumn{1}{c|}{33.8}  & 0.969 (0.23)           & \multicolumn{1}{c|}{31.3}  & 0.97 (0.24)            & \multicolumn{1}{c|}{215.1}   & 0.72 (4)               & 62.6  & 6       \\
\multicolumn{1}{l|}{Delta}     & 7129                  & \multicolumn{1}{c|}{5}  & 0.701 (0.64)           & \multicolumn{1}{c|}{10.9}  & 0.701 (0.63)           & \multicolumn{1}{c|}{10.4}  & 0.697 (0.7)            & \multicolumn{1}{c|}{7.577}   & 0.678 (0.7)            & 8.56  & 4       \\
\multicolumn{1}{l|}{Elevators} & 16599                 & \multicolumn{1}{c|}{18} & 0.872 (0.5)            & \multicolumn{1}{c|}{40.7}  & 0.871 (0.47)           & \multicolumn{1}{c|}{39.5}  & 0.812 (0.58)           & \multicolumn{1}{c|}{482.6}   & 0.813 (0.6)            & 220.4 & 26      \\
\multicolumn{1}{l|}{Friedman}  & 40768                 & \multicolumn{1}{c|}{10} & 0.895 (0.75)           & \multicolumn{1}{c|}{134.3} & 0.894 (0.85)           & \multicolumn{1}{c|}{134.5} & 0.886 (0.13)           & \multicolumn{1}{c|}{294.2}   & 0.723 (0.18)           & 56.28 & 6       \\
\multicolumn{1}{l|}{Housing}   & 506                   & \multicolumn{1}{c|}{13} & 0.85 (4.6)             & \multicolumn{1}{c|}{2.9}   & 0.846 (3.4)            & \multicolumn{1}{c|}{2.9}   & 0.782 (5.17)           & \multicolumn{1}{c|}{43.49}   & 0.699 (10.4)           & 3.8   & 7       \\
\multicolumn{1}{l|}{Kin8nm}    & 8192                  & \multicolumn{1}{c|}{8}  & 0.72 (1.9)             & \multicolumn{1}{c|}{19.6}  & 0.723 (2.28)           & \multicolumn{1}{c|}{19}    & 0.566 (1.8)            & \multicolumn{1}{c|}{439.8}   & 0.412 (1.2)            & 10.9  & 5       \\
\multicolumn{1}{l|}{Lpga2009}  & 146                   & \multicolumn{1}{c|}{11} & 0.765 (7.4)            & \multicolumn{1}{c|}{2.2}   & 0.891 (2.58)           & \multicolumn{1}{c|}{2.23}  & 0.86 (3.1)             & \multicolumn{1}{c|}{12.1}    & 0.864 (10.9)           & 0.89  & 4       \\
\multicolumn{1}{l|}{Puma}      & 8192                  & \multicolumn{1}{c|}{32} & 0.815 (0.96)           & \multicolumn{1}{c|}{45.3}  & 0.815 (0.9)            & \multicolumn{1}{c|}{44.95} & 0.823 (0.9)            & \multicolumn{1}{c|}{10592.2} & 0.216 (3.2)            & 53.54 & 7       \\
\multicolumn{1}{l|}{Yacht}     & 308                   & \multicolumn{1}{c|}{6}  & 0.987 (0.58)           & \multicolumn{1}{c|}{2.54}  & 0.976 (0.77)           & \multicolumn{1}{c|}{2.49}  & 0.991 (0.18)           & \multicolumn{1}{c|}{0.972}   & 0.643 (6.2)            & 0.636 & 11      \\ \hline
Arithmetic avg                 & \multicolumn{1}{l}{}  & \multicolumn{1}{l|}{}   & \textbf{0.835} (1.85)           & \multicolumn{1}{c|}{\textbf{24.3}}  & \textbf{0.843} (1.5)            & \multicolumn{1}{c|}{\textbf{24.1}}  & \textbf{0.814} (2.25)           & \multicolumn{1}{c|}{\textbf{873.3}}   & \textbf{0.666} (5.2)            & \textbf{70}    & 11.4    \\ \hline
Geometric avg                  & \multicolumn{1}{l}{}  & \multicolumn{1}{l|}{}   & \textbf{0.827} (1.2)            & \multicolumn{1}{c|}{\textbf{11.1}}  & \textbf{0.834} (1.07)           & \multicolumn{1}{c|}{\textbf{10.9}}  & \textbf{0.804} (1.11)           & \multicolumn{1}{c|}{\textbf{55.8}}    & \textbf{0.633} (3.03)           & \textbf{11.73} & 8.62    \\ \hline
\end{tabular}
}
\caption{Comparison between SRT, SRT $\ell_2$, ORT-L and ORRT for regression trees of depth $D=2$. \textcolor{black}{The numbers appearing in the rightmost column correspond to the number of times (out of the 80 runs) that ORRT provides a suboptimal solution with a negative testing $R^2$. For SRT, SRT $\ell_2$ and ORT-L, such a number is equal to $0$.}}\label{tab:depth2}
\end{table}

\section{ORT-L with grid search for hyperparameters tuning}\label{sec:ortl_cp}
We assess the impact of a grid search for both the complexity parameters $cp$ and the $l_1$ regularization hyperparameters on the performance of ORT-L with the MILO-based local search method in \citep{dunn2018optimal,bertsimas2019machine}.

The results reported in Section \ref{sec:experiments_regression} are obtained by fixing $cp=0$ and the $l_1$ hyperparameter $\lambda= 1e^{-3}$ in order to reduce computational time as much as possible. Here we present results on all the 15 datasets where we apply a grid search on both the complexity parameter $cp$ (internally handled by the software) and the regularization, choosing among three different values for $\lambda$ ($1e^{-3},5e^{-3},1e^{-2}$). 

The results on the impact of enabling and disabling the grid search for ORT-L with a depth of $D=2$ are presented in Table \ref{tab:depth2_cp}, while those with a depth $D=3$ are reported in Table \ref{tab:depth3_cp}. The ORT-L trained via the grid search is referred to as ORT-L-grid. The tables show the average testing $R^2$, its standard deviation $\sigma$ (divided by a $1e^{-2}$ factor for visualization reasons) and the average training time in seconds. For the sake of comparison we also report the results obtained by SRT $\ell_2$ (our $\ell_2$ regularized SRT model trained with NODEC-DR). The arithmetic and geometric averages shown in the last two rows of Table \ref{tab:depth3_cp} for the results at depth $D=3$ exclude the Puma dataset because ORT-L with grid search (ORT-L-grid) needed more than 500 minutes of computation time.

The results with the grid search were perfectly comparable to the ones obtained by fixing the hyperparameters in terms of testing $R^2$ but the computational times were significantly higher.

\begin{table}[H]
\centering
\resizebox{\columnwidth}{!}{
\begin{tabular}{lccccccccc}
\hline
               & \multicolumn{9}{c}{D=2}                                                                                          \\ \cline{2-10} 
               & \multicolumn{3}{c}{\textbf{SRT}\bm{$\ell_2$}}  & \multicolumn{3}{c}{\textbf{ORT-L}}   & \multicolumn{3}{c}{\textbf{ORT-L-grid}}         \\ \hline
Dataset        & $R^2$ & $\sigma(1e^{-2})$ & Time   & $R^2$ & $\sigma(1e^{-2})$ & Time     & $R^2$ & $\sigma(1e^{-2})$ & Time     \\ \hline
Abalone        & 0.554 & 2.14              & 8.74   & 0.558 & 1.7               & 70.6    & 0.55  & 2.2               & 166.4   \\
Ailerons       & 0.836 & 0.56              & 18.9  & 0.824 & 1.13              & 262.2   & 0.828 & 1                 & 1160.2   \\
Airfoil        & 0.748 & 2.9               & 5.16   & 0.744 & 1.6               & 2.07     & 0.743 & 1.76              & 7.96     \\
Auto mpg       & 0.87  & 2.37              & 2.47   & 0.838 & 1.21              & 5.94     & 0.838 & 2                 & 17.7    \\
Compact        & 0.98  & 0.168             & 37.2  & 0.981 & 0.34              & 668.4   & 0.979 & 0.55              & 1985.4  \\
Computer       & 0.964 & 2.49              & 1.9   & 0.877 & 13.4              & 1.9      & 0.908 & 13.37             & 6.03     \\
Cpu small      & 0.969 & 0.23              & 31.3  & 0.97  & 0.24              & 215.1   & 0.969 & 0.43              & 728.2   \\
Delta          & 0.701 & 0.63              & 10.4   & 0.697 & 0.7               & 7.6    & 0.698 & 0.68              & 27.9    \\
Elevators      & 0.871 & 0.47              & 39.5  & 0.812 & 0.58              & 482.6    & 0.812 & 0.58              & 1352.0  \\
Friedman       & 0.894 & 0.85              & 134.5 & 0.886 & 0.13              & 294.2    & 0.886 & 0.13              & 1081.6  \\
Housing        & 0.846 & 3.4               & 2.9   & 0.782 & 5.17              & 43.5    & 0.764 & 9.22              & 134.01   \\
Kin8nm         & 0.723 & 2.28              & 19.0  & 0.566 & 1.8               & 439.8   & 0.566 & 1.79              & 1394.2  \\
Lpga2009       & 0.891 & 2.58              & 2.2   & 0.86  & 3.1               & 12.1    & 0.879 & 3.8               & 42.8     \\
Puma           & 0.815 & 0.9               & 44.9  & 0.823 & 0.9               & 10592.2 & 0.823 & 1.23              & 33956.9 \\
Yacht          & 0.976 & 0.77              & 2.49   & 0.991 & 0.18              & 0.972    & 0.987 & 0.52              & 4.28     \\ \hline
Arithmetic avg & \textbf{0.842} & 1.51              & \textbf{24.1}  & \textbf{0.814} & 2.14              & \textbf{873.27}   & \textbf{0.815} & 2.62              & \textbf{2804.4}  \\ \hline
Geometric avg  & \textbf{0.837} & 1.07              & \textbf{10.9}  & \textbf{0.803} & 1                 & \textbf{55.83}    & \textbf{0.804} & 1.3               & \textbf{186.3}   \\ \hline
\end{tabular}
}
\caption{Comparison between SRT $\ell_2$, ORT-L (without grid search) and ORT-L-grid (with grid search) for regression trees of depth $D = 2$.}\label{tab:depth2_cp}
\end{table}

Overall, for both depths the results with the grid search were perfectly comparable to the ones obtained by fixing the hyperparameters in terms of testing $R^2$ but the computational times were significantly higher. In particular, the difference in the average testing $R^2$ obtained from the ORT-L model with and without grid search is $0.1\%$ for trees of depth $D=2$ and $0.6\%$ for depth $D=3$, respectively. The very slight improvement in accuracy obtained through grid search is offset by a significant rise in computational time. At both depths, the final geometric average rows illustrates that employing grid search leads to a computational time more than two times greater compared to the solution obtained without grid search. Indeed, at depth $D=2$ ORT-L reaches on average a testing $R^2$ of $80.3\%$ in less than 55.83 seconds; while ORT-L-grid achieves an accuracy of $80.4\%$ in 186.26 seconds. In the case of depth $D=3$, where the Puma dataset is not considered since ORT-L-grid reached the time limit, ORT-L obtains a testing $R^2$ of $81.8\%$ in 73.59 seconds; while ORT-L-grid achieves an accuracy of $82.4\%$ in more than 191.81 seconds. It is worth pointing out that at both depths the SRT $\ell_2$ outperforms ORT-L-grid in terms of accuracy ($83.7\%$ for depth $D=2$ and $85.1\%$ for $D=3$) and computational time (one order of magnitude).

\begin{table}[H]
\resizebox{\columnwidth}{!}{
\begin{tabular}{lccccccccc}
\hline
                & \multicolumn{9}{c}{D=3}                                                                                         \\ \cline{2-10} 
                & \multicolumn{3}{c}{\textbf{SRT} \bm{$\ell_2$}}  & \multicolumn{3}{c}{\textbf{ORT-L}}   & \multicolumn{3}{c}{\textbf{ORT-L-grid}}        \\ \hline
Dataset         & $R^2$ & $\sigma(1e^{-2})$ & Time   & $R^2$ & $\sigma(1e^{-2})$ & Time     & $R^2$ & $\sigma(1e^{-2})$ & Time    \\ \hline
Abalone         & 0.564 & 2.16              & 23.9  & 0.545 & 1.9               & 147.9   & 0.549 & 1.86              & 254.2  \\
Ailerons        & 0.835 & 0.43              & 50.5  & 0.825 & 1.2               & 310.5   & 0.828 & 0.98              & 1303.7 \\
Airfoil         & 0.807 & 2.14              & 11.3  & 0.842 & 1.3               & 4.7     & 0.85  & 1.83              & 18.0   \\
Auto mpg        & 0.873 & 2.17              & 6.1   & 0.819 & 6.47              & 10.0    & 0.84  & 2.82              & 29.5   \\
Compact         & 0.98  & 0.27              & 72.7  & 0.98  & 0.48              & 1103.5  & 0.978 & 0.8               & 3001.1 \\
Computer        & 0.955 & 3.37              & 4.0   & 0.889 & 16                & 3.2      & 0.89  & 17.87             & 8.7    \\
Cpu small       & 0.97  & 0.25              & 58.7  & 0.971 & 0.31              & 427.4   & 0.971 & 0.45              & 1375.1 \\
Delta           & 0.706 & 1                 & 29.5  & 0.709 & 0.77              & 17.4    & 0.701 & 1.06              & 55.8    \\
Elevators       & 0.884 & 0.57              & 94.7  & 0.812 & 0.58              & 515.8   & 0.812 & 0.58              & 1275.0 \\
Friedman        & 0.938 & 0.28              & 311.8 & 0.935 & 0.06              & 1356.6  & 0.935 & 0.06              & 4987.2 \\
Housing         & 0.872 & 4.48              & 7.3   & 0.785 & 8.3               & 78.4     & 0.81  & 7.3               & 25.7   \\
Kin8nm          & 0.786 & 1.9               & 50.7  & 0.645 & 1.1               & 946.5   & 0.643 & 1.3               & 2970.1 \\
Lpga2009        & 0.877 & 2.5               & 4.9   & 0.844 & 4                 & 16.5    & 0.876 & 39.4              & 61.2   \\
Puma            & 0.883 & 0.88              & 136.5 & 0.907 & 0.39              & 28481.8 &       &                   &         \\
Yacht           & 0.983 & 0.48              & 5.4   & 0.992 & 0.11              & 2.81     & 0.992 & 0.28              & 8.68    \\ \hline
Arithmetic avg* & \textbf{0.859} & 1.57              & \textbf{52.27}  & \textbf{0.828} & 3.04              & \textbf{352.94}   & \textbf{0.834} & 5.47              & \textbf{1098.15} \\ \hline
Geometric avg*  & \textbf{0.851} & 1.03              & \textbf{23.14}  & \textbf{0.818} & 1.12              & \textbf{73.59}    & \textbf{0.824} & 1.48              & \textbf{191.81}  \\ \hline
\end{tabular}
}
\caption{Comparison between SRT $\ell_2$, ORT-L (without grid search) and ORT-L-grid (with grid search) with depth $D = 3$. The last two rows reporting the averages are with * since we do not consider the Puma dataset for their computation.}\label{tab:depth3_cp}
\end{table}
\section{Comparison of ORT-L with ORT-LH}\label{appendix:ortl_vs_ortlh}

\textcolor{black}{In this appendix we compare the ORT-L and ORT-LH approaches proposed in \citep{dunn2018optimal,bertsimas2019machine} to train univariate and, respectively, multivariate deterministic regression trees on a selection of three datasets.}

\textcolor{black}{Table \ref{tab:ortlh_results} reports the average testing $R^2$, standard deviation $\sigma$, computational time in seconds and negative $R^2$ on the datasets Auto mpg, Computer and Housing, for regression trees of depth $D=2$ and $D=3$. The experimental results for trees of both depths are quite similar. As already observed in \citep{dunn2018optimal,bertsimas2019machine}, ORT-LH performs similarly to ORT-L in terms of testing $R^2$ and the substantial increase in computational time with respect to ORT-L (at least two orders of magnitude higher) is not compensated by a significant improvement in terms of $R^2$. Therefore in this article we consider ORT-L. 
}

\begin{table}[h]
\centering
\resizebox{\textwidth}{!}{
\begin{tabular}{lccccccccc|ccccccc}
\hline
         & \multicolumn{1}{l}{} & \multicolumn{1}{l}{}    & \multicolumn{7}{c|}{D = 2}                                                                 & \multicolumn{7}{c}{D = 3}                                                                 \\ \cline{4-17} 
         & \multicolumn{1}{l}{} & \multicolumn{1}{l}{}    & \multicolumn{3}{c}{\textbf{ORT-L}}                       & \multicolumn{4}{c|}{\textbf{ORT-LH}}              & \multicolumn{3}{c}{\textbf{ORT-L}}                       & \multicolumn{4}{c}{\textbf{ORT-LH}}              \\ \cline{4-17} 
Dataset  & N                    & p                       & $R^2$ & $\sigma$\tiny($1e^{-2}$) & Time                       & $R^2$ & $\sigma$\tiny($1e^{-2}$) & Time    & $R^2<0$ & $R^2$ & $\sigma$\tiny($1e^{-2}$) & Time                       & $R^2$ & $\sigma$\tiny($1e^{-2}$) & Time   & $R^2<0$ \\ \hline
Auto mpg & 392                  & \multicolumn{1}{c|}{7}  & 0.838 & 2.8        & \multicolumn{1}{c|}{5.9}  & 0.84  & 3.4        & 492.1  & 0         & 0.819 & 6.47       & \multicolumn{1}{c|}{10.0} & 0.83  & 3.7        & 936.7 & 0         \\
Computer & 209                  & \multicolumn{1}{c|}{37} & 0.877 & 13.4       & \multicolumn{1}{c|}{1.9}   & 0.859 & 14.6       & 124.8  & 3         & 0.889 & 16         & \multicolumn{1}{c|}{3.2}   & 0.842 & 14.4       & 189.9  & 1         \\
Housing  & 506                  & \multicolumn{1}{c|}{13} & 0.782 & 5.17       & \multicolumn{1}{c|}{43.5} & 0.807 & 6.67       & 2712.7 & 1         & 0.785 & 8.33       & \multicolumn{1}{c|}{78.4}  & 0.784 & 10.6       & 5327   & 2         \\ \hline
\end{tabular}}
\caption{Comparison of the MILO formulations ORT-L and ORT-LH  on 3 datasets in terms of testing $R^2$, standard deviation $\sigma$ (to be multiplied by $1e^{-2}$), computational time and anomalies.}
\label{tab:ortlh_results}
\end{table}

\section{Average solutions of SRT versus average and best solutions of ORRT} \label{appendix:orrt_multi}

\textcolor{black}{In this appendix we compare the best solutions provided by ORRT over a prescribed number of initial solutions, as considered in \citep{blanquero2022sparse}, with the average solutions found by ORRT, SRT and SRT $\ell_2$, where the averages are taken over all the initial solutions and folds.}

\textcolor{black}{In Table \ref{tab:multistart_orrt} we report the results obtained for four datasets (Airfoil, Cpu small, Puma and Yacht) with soft trees of depths $D=2$ and $D=3$ in terms of testing $R^2$ and standard deviation $\sigma$. 
In ORRT* we consider the average over the $k=4$ folds of the best solutions found over 20 initial solutions, 
while in ORRT, SRT and SRT $\ell_2$ we consider the average solutions over the 80 runs.}

\textcolor{black}{For soft trees of both depths, the multistart approach ORRT* yields better solutions in terms of testing $R^2$ with respect to ORRT but with still substantially lower testing accuracy and higher average $\sigma$ than the two SRT versions.}

\begin{table}[h]
\centering
\resizebox{0.9\columnwidth}{!}{
\begin{tabular}{lcccccccc|cccccccc}
\hline
                               & \multicolumn{8}{c|}{D=2}                                                                                                                 & \multicolumn{8}{c}{D=3}                                                                                                                  \\ \cline{2-17} 
                               & \multicolumn{2}{c}{\textbf{ORRT}}          & \multicolumn{2}{c}{\textbf{ORRT*}}         & \multicolumn{2}{c}{\textbf{SRT}}          & \multicolumn{2}{c|}{\textbf{SRT} \bm{$\ell_2$}} & \multicolumn{2}{c}{\textbf{ORRT}}          & \multicolumn{2}{c}{\textbf{ORRT*}}         & \multicolumn{2}{c}{\textbf{SRT}}           & \multicolumn{2}{c}{\textbf{SRT} \bm{$\ell_2$}} \\ \cline{2-17} 
Dataset                        & $R^2$ & $\sigma$\tiny($1e^{-2}$)                  & $R^2$ & $\sigma$\tiny($1e^{-2}$)                  & $R^2$ & $\sigma$\tiny($1e^{-2}$)                  & $R^2$       & $\sigma$\tiny($1e^{-2}$)       & $R^2$ & $\sigma$\tiny($1e^{-2}$)                  & $R^2$ & $\sigma$\tiny($1e^{-2}$)                  & $R^2$ & $\sigma$\tiny($1e^{-2}$)                   & $R^2$       & $\sigma$\tiny($1e^{-2}$)     \\ \hline
\multicolumn{1}{l|}{Airfoil}   & 0.516 & \multicolumn{1}{c|}{2.89} & 0.619 & \multicolumn{1}{c|}{0.76} & 0.767 & \multicolumn{1}{c|}{2.53} & 0.757       & 2.6            & 0.513 & \multicolumn{1}{c|}{1.8}  & 0.603 & \multicolumn{1}{c|}{1.95} & 0.855 & \multicolumn{1}{c|}{1.97}  & 0.819       & 0.8           \\
\multicolumn{1}{l|}{Cpu small} & 0.72  & \multicolumn{1}{c|}{4}    & 0.917 & \multicolumn{1}{c|}{5.6}  & 0.969 & \multicolumn{1}{c|}{0.32} & 0.969       & 0.23           & 0.705 & \multicolumn{1}{c|}{7.24} & 0.934 & \multicolumn{1}{c|}{1.54} & 0.97  & \multicolumn{1}{c|}{0.29}  & 0.97        & 0.27          \\
\multicolumn{1}{l|}{Puma}      & 0.216 & \multicolumn{1}{c|}{3.2}  & 0.285 & \multicolumn{1}{c|}{3.4}  & 0.818 & \multicolumn{1}{c|}{0.46} & 0.819       & 0.48           & 0.215 & \multicolumn{1}{c|}{3.2}  & 0.3   & \multicolumn{1}{c|}{4.5}  & 0.895 & \multicolumn{1}{c|}{0.68}  & 0.896       & 0.66          \\
\multicolumn{1}{l|}{Yacht}     & 0.643 & \multicolumn{1}{c|}{6.2}  & 0.861 & \multicolumn{1}{c|}{12.5} & 0.992 & \multicolumn{1}{c|}{0.13} & 0.979       & 0.68           & 0.645 & \multicolumn{1}{c|}{5.6}  & 0.861 & \multicolumn{1}{c|}{8.1}  & 0.994 & \multicolumn{1}{c|}{0.058} & 0.981       & 0.56          \\ \hline
\end{tabular}
}
\caption{Comparison of the evaluation criteria for ORRT on 4 datasets in terms of testing $R^2$ and standard deviation $\sigma$ (to be multiplied by $1e^{-2}$). The multistart evaluation criterion is referred to as ORRT*.}\label{tab:multistart_orrt}
\end{table}

\section{Comparison with Random Forest}\label{appendix:HME_forest}

In this appendix we compare SRT $\ell_2$ (our $\ell_2$ regularized SRT model trained with NODEC-DR) to the well-known highly-performing Random Forest (RF) ensemble method.


Ensemble methods are ML techniques that combine outputs from multiple models. They generally provide improved accuracies over individual greedy trees but this improvement incurs reduced interpretability and increased computational time. 
One of the most successful and widely used ensemble method is RF \citep{breiman2001random}. RF combines individual trees, each built independently using bootstrapped data points and random subsets of input features through a greedy approach. In regression tasks, the final prediction is obtained by averaging the individual tree predictions.

We adopt the same experimental settings described in Subsection \ref{sec:dataset_exp} with $k=4$ fold cross-validation and 20 different random seeds per fold. For the RF training we use the \texttt{randomForest} package \citep{rcolorbrewer2018package} in R 4.0.1 with the default parameter setting.

It is important to point out that in the case of SRT $\ell_2$ we have a single multivariate tree, whereas RF is a computationally very light ensemble method that generates on the order of hundreds of trees. 

Table \ref{tab:RF} reports the testing $R^2$ and the standard deviation $\sigma$. In terms of computational time RF turns out to be substantially less demanding than SRT $\ell_2$, but the latter globally optimizes the parameters of a single regression tree which is more amenable to interpretation. 

\begin{table}[!htb]
\centering
\begin{tabular}{lcccccc}
\hline
        & \multicolumn{2}{c}{\textbf{SRT} \bm{$\ell_2$} (D= 2) } & \multicolumn{2}{c}{\textbf{SRT} \bm{$\ell_2$} (D=3)} & \multicolumn{2}{c}{\textbf{RF}}  \\ \cline{2-7} 
Dataset & $R^2$        & $\sigma (1e^{-2})$        & $R^2$        & $\sigma (1e^{-2})$        & $R^2$ & $\sigma (1e^{-2})$ \\ \hline
Abalone   & 0.554 & 2.14  & 0.564 & 2.16 & 0.554 & 2.19 \\
Ailerons  & 0.836 & 0.56  & 0.835 & 0.43 & 0.824 & 0.7  \\
Airfoil   & 0.748 & 2.9   & 0.807 & 2.14 & 0.726 & 0.48 \\
Auto-mpg  & 0.87  & 2.37  & 0.873 & 2.17 & 0.876 & 1.97 \\
Compact   & 0.98  & 0.17 & 0.98  & 0.27 & 0.982 & 0.2  \\
Computer  & 0.964 & 2.49  & 0.955 & 3.37 & 0.897 & 5.54 \\
Cpu small & 0.969 & 0.23  & 0.97  & 0.25 & 0.976 & 0.2  \\
Delta     & 0.701 & 0.63  & 0.706 & 1    & 0.708  & 0.94 \\
Elevators & 0.871 & 0.47  & 0.884 & 0.57 & 0.812 & 0.73 \\
Friedman  & 0.894 & 0.85  & 0.938 & 0.28 & 0.924  & 0.13 \\
Housing   & 0.846 & 3.4   & 0.872 & 4.48 & 0.878  & 3.5  \\
Kin8nm    & 0.723 & 2.28  & 0.786 & 1.9  & 0.682 & 0.97 \\
Lpga2009  & 0.891 & 2.58  & 0.877 & 2.5  & 0.89   & 4.2  \\
Puma      & 0.815 & 0.9   & 0.883 & 0.88 & 0.871  & 0.5  \\
Yacht     & 0.976 & 0.77  & 0.983 & 0.48 & 0.925  & 2    \\ \hline
Arithmetic avg  & \textbf{0.842} & 1.5   & \textbf{0.861} & 1.52 & \textbf{0.835}  & 1.6  \\ \hline
Geometric avg  & \textbf{0.834} & 1.07  & \textbf{0.853} & 1.02 & \textbf{0.826}  & 0.94 \\ \hline
\end{tabular}
\caption{Comparison between SRT $\ell_2$ and RF on 15 datasets in terms of testing $R^2$, standard deviation $\sigma$ (to be multiplied by $1e^{-2}$).}\label{tab:RF}
\end{table}

As shown by the last rows reporting the average testing $R^2$, despite just considering a single tree, SRT $\ell_2$ achieves slightly better accuracy than the RF ensemble method. This is particularly interesting when looking at the average SRT $\ell_2$ results of depth $D=2$ where comparable accuracy corresponds to a remarkable level of interpretability. SRT $\ell_2$ exhibits better performance compared to RF in most of the datasets. In particular, SRT $\ell_2$ achieves better testing $R^2$ on 8 out of 15 datasets for depth $D=2$, and on 9 out of 15 datasets for depth $D=3$. For some datasets, such as Elevators and Kin8nm, SRT $\ell_2$ of depth $D=3$ is able to reach a level of testing $R^2$ up to $7.2\%$ and $10.4\%$ higher, respectively. On the same datasets the difference between SRT $\ell_2$ and HMEs for depth $D=2$ is of more than $5\%$ and $4\%$, respectively. 

There is a small group of datasets, consisting of Auto-mpg, Compact, Cpu small, Delta, and Housing where RF achieves a slightly higher level of testing $R^2$ than SRT $\ell_2$ of depth $D=3$, but it is never more than $0.6\%$. Indeed, for Auto-mpg, Compact and Delta datasets the gain in terms of testing $R^2$ obtained by RF is no more than $0.3\%$. Moreover, looking at the standard deviations the results for SRT $\ell_2$ have less variability, and thus are less sensitive to the initial solutions.

In general, SRT $\ell_2$ proves to be highly competitive and frequently outperforms RF models. Our single tree SRT $\ell_2$ model is able to achieve higher level of testing $R^2$ than the well-established ensemble method, while also offering a higher level of interpretability.
\end{document}